\RequirePackage{rotating}

\documentclass[manuscript,screen,authorversion,nonacm]{acmart}

\AtBeginDocument{%
  }

\setcopyright{none}
\copyrightyear{2024}
\acmYear{2024}
\acmDOI{XXXXXXX.XXXXXXX}

\acmJournal{CSUR}

\usepackage{paralist}
\usepackage[utf8]{inputenc} 
\usepackage[T1]{fontenc}    
\usepackage{hyperref}       
\usepackage{url}            
\usepackage{wrapfig,lipsum,booktabs}       
\usepackage{amsfonts}       
\usepackage{nicefrac}       
\usepackage{microtype}      
\usepackage{colortbl}
\usepackage{pifont}
\usepackage{rotating}
\usepackage{makecell}

\usepackage{bbm}

\usepackage{caption}
\usepackage{subcaption}
\usepackage{amsmath}
\usepackage{xspace}
\usepackage{multirow}

\usepackage{algorithm}
\usepackage{algpseudocode}

\usepackage[titletoc]{appendix}

\usepackage{pdflscape}

\newlength\savewidth

\newcolumntype{C}{>{\centering\let\newline\\\arraybackslash\hspace{0pt}}m{2cm}}

\newcommand{\cmark}{\ding{51}}%
\newcommand{\xmark}{\ding{55}}%
\newcommand{\club}{\ding{168}}

\definecolor{gray1}{HTML}{f2f2f2}
\definecolor{gray2}{HTML}{e5e5e5}
\definecolor{gray3}{HTML}{cccccc}
\definecolor{cream}{HTML}{fef9d2}
\definecolor{green1}{HTML}{e2fee2}
\definecolor{blue1}{HTML}{d7e0f9}

\usepackage[pscoord]{eso-pic}

\newcommand{\placetextbox}[3]{
  \setbox0=\hbox{#3}
  \AddToShipoutPictureFG*{
    \put(\LenToUnit{#1\paperwidth},\LenToUnit{#2\paperheight}){\vtop{{\null}\makebox[0pt][c]{#3}}}%
  }%
}%

\usepackage{amsmath,amsfonts,bm,eqnarray}
\usepackage{cancel}
\usepackage{amsthm}
\usepackage{tikz}
\usetikzlibrary{tikzmark}

\newtheorem{definition}{Definition}[section]

\newtheorem*{remark}{Remark}

\newtheorem{innercustomgeneric}{\customgenericname}
\providecommand{\customgenericname}{}
\newcommand{\newcustomtheorem}[2]{%
  \newenvironment{#1}[1]
  {%
   \renewcommand\customgenericname{#2}%
   \renewcommand\theinnercustomgeneric{##1}%
   \innercustomgeneric
  }
  {\endinnercustomgeneric}
}

\newcustomtheorem{customThm}{Theorem}
\newcustomtheorem{customLemma}{Lemma}
\newcustomtheorem{customCor}{Corollary}
\newcustomtheorem{customProposition}{Proposition}

\def\Tabref#1{Table~\ref{#1}}

\def\Defref#1{Definition~\ref{#1}}

\def\Figref#1{Fig.~\ref{#1}}

\def\appref#1{Appendix~\ref{#1}}
\def\Secref#1{Section~\ref{#1}}

\def\eqref#1{equation~\ref{#1}}

\def\1{\bm{1}}

\def\vtheta{{\bm{\theta}}}

\def\vv{{\bm{v}}}

\def\vx{{\bm{x}}}
\def\vy{{\bm{y}}}

\def\mA{{\bm{A}}}

\def\mI{{\bm{I}}}

\def\mP{{\bm{P}}}

\def\mSigma{{\bm{\Sigma}}}

\DeclareMathAlphabet{\mathsfit}{\encodingdefault}{\sfdefault}{m}{sl}
\SetMathAlphabet{\mathsfit}{bold}{\encodingdefault}{\sfdefault}{bx}{n}

\def\gA{{\mathcal{A}}}

\def\gD{{\mathcal{D}}}
\def\gE{{\mathcal{E}}}

\def\gH{{\mathcal{H}}}
\def\gI{{\mathcal{I}}}

\def\gL{{\mathcal{L}}}

\def\gT{{\mathcal{T}}}

\def\gX{{\mathcal{X}}}
\def\gY{{\mathcal{Y}}}

\newcommand{\E}{\mathbb{E}}

\renewcommand{\tilde}{\widetilde}
\renewcommand{\hat}{\widehat}
\renewcommand{\frac}{\tfrac}

\begin{document}

\title{Continual Learning of Large Language Models: A Comprehensive Survey}

\author{Haizhou Shi}
\authornote{Correspondence to: Haizhou Shi <haizhou.shi@rutgers.edu> and Hao Wang <hw488@cs.rutgers.edu>.}
\email{haizhou.shi@rutgers.edu}
\author{Zihao Xu}
\author{Hengyi Wang}
\author{Weiyi Qin}
\author{Wenyuan Wang}
\authornote{Work done as visiting students at Rutgers Machine Learning Lab.}

\author{Yibin Wang}
\authornotemark[2]

\affiliation{%
    \institution{Rutgers University}
    \country{USA}
}



\author{Zifeng Wang}
\author{Sayna Ebrahimi}
\affiliation{%
    \institution{Google Cloud AI Research}
    \city{Mountain View}
    \country{USA}
}

\author{Hao Wang}
\authornotemark[1]
\email{hw488@cs.rutgers.edu}
\affiliation{%
    \institution{Rutgers University}
    \country{USA}
}





\renewcommand{\shortauthors}{Shi et al.}
\authorsaddresses{} 

\begin{abstract}
    The challenge of effectively and efficiently adapting statically pre-trained Large Language Models~(LLMs) to ever-evolving data distributions remains predominant. When tailored for specific needs, pre-trained LLMs often suffer from significant performance degradation in previous knowledge domains -- a phenomenon known as \emph{``catastrophic forgetting''}. While extensively studied in the Continual Learning~(CL) community, this problem presents new challenges in the context of LLMs. In this survey, we provide a comprehensive overview and detailed discussion of the current research progress on LLMs within the context of CL. Besides the introduction of the preliminary knowledge, this survey is structured into four main sections: we first describe an overview of continually learning LLMs, consisting of two directions of continuity: \emph{vertical continuity (or vertical continual learning)}, i.e., continual adaptation from general to specific capabilities, and \emph{horizontal continuity  (or horizontal continual learning)}, i.e., continual adaptation across time and domains~(\Secref{sec:overview}). Following vertical continuity, we summarize three stages of learning LLMs in the context of modern CL: Continual Pre-Training~(CPT), Domain-Adaptive Pre-training~(DAP), and Continual Fine-Tuning~(CFT)~(\Secref{sec:stages}). We then provide an overview of evaluation protocols for continual learning with LLMs, along with currently available data sources~(\Secref{sec:eval-and-data}). Finally, we discuss intriguing questions related to continual learning for LLMs~(\Secref{sec:discussion}). This survey sheds light on the relatively understudied domain of continually pre-training, adapting, and fine-tuning large language models, suggesting the necessity for greater attention from the community.  Key areas requiring immediate focus include the development of practical and accessible evaluation benchmarks, along with methodologies specifically designed to counter forgetting and enable knowledge transfer within the evolving landscape of LLM learning paradigms. The full list of papers examined in this survey is available at \href{https://github.com/Wang-ML-Lab/llm-continual-learning-survey}{https://github.com/Wang-ML-Lab/llm-continual-learning-survey}.
\end{abstract}

\ccsdesc[500]{Computing methodologies~Lifelong machine learning}
\ccsdesc[500]{Computing methodologies~Natural language processing}
\ccsdesc[500]{Computing methodologies~Neural networks}


\keywords{Large Language Models, Continual Learning.}


\maketitle

\section{Introduction}
\label{sec:intro}
Recent advances in large language models~(LLMs) have demonstrated considerable potential for achieving artificial general intelligence (AGI)~\cite{radford2019language,brown2020language,achiam2022chatgpt,achiam2023gpt,chowdhery2023palm,anil2023palm,touvron2023llama,touvron2023llama2}.
Researchers have observed that complex abilities such as multi-step reasoning, few-shot in-context learning, and instruction following improve as the scale of parameter size increases~\cite{wei2022chain,wei2022emergent,yao2024tree,wei2021finetuned,min2022rethinking}.
The development of LLMs is impactful and revolutionary, prompting machine learning practitioners to reconsider traditional computational paradigms for once-challenging human-level tasks.
However, LLMs are typically trained on static, pre-collected datasets encompassing general domains, leading to gradual performance degradation over time~\cite{loureiro2022timelms,jang2022towards,jin2022lifelong,jang2022temporalwiki,amba2021dynamic,dhingra2022time} and across different content domains~\cite{gupta2023continual,jin2022lifelong,ke2022continual-train,sun2020ernie,cossu2022continual,gururangan2022demix,qin2023recyclable,chen2023lifelong,qin2022elle}.
Additionally, a single pre-trained large model cannot meet every user need and requires further fine-tuning~\cite{weyssow2023usage,winata2023overcoming,zheng2023learn,winata2023overcoming,biderman2023pythia,zheng2023learn,bai2023enhancing,ke2021achieve,wei2022circle,qin2021lfpt5,chen2024parameterizing}.
While one potential solution is re-collecting pre-training data and re-training models with additional specific needs, this approach is prohibitively expensive and impractical in real-world scenarios. 

To efficiently adapt LLMs to downstream tasks while minimizing performance degradation on previous knowledge domains, researchers employ the methodology of Continual Learning~(CL), also known as \emph{lifelong learning} or \emph{incremental learning}~\cite{pentina2016theoretical,chen2018lifelong,van2022three,wang2024comprehensive}. 
Inspired by the incremental learning pattern observed in human brains~\cite{mcclelland1995there,kandel2000principles,pallier2003brain,mccaffary2021towards}, CL trains machine learning models sequentially on a series of tasks with the expectation of maintaining performance across all tasks~\cite{kirkpatrick2017overcoming,li2017learning,ebrahimi2020adversarial,ebrahimi2019uncertainty}. Throughout training, models have limited or no access to previous data, posing a challenge in retaining past knowledge as optimization constraints from unseen previous data are absent during current-task learning~\cite{li2017learning,lomonaco2020rehearsalfree,shi2024unified}. This challenge, known as \emph{catastrophic forgetting}~\cite{mccloskey1989catastrophic}, has been a central focus in continual learning research since its inception.
Over the years, researchers have explored various techniques to mitigate forgetting. These include replay-based methods~\cite{chaudhry2019tiny,schwarz2018progress,shi2024unified}, parameter regularization~\cite{kirkpatrick2017overcoming,ritter2018online,aljundi2018memory}, and model architecture expansion~\cite{ramesh2021model,wang2022coscl}. Together, these techniques have significantly advanced the goal of achieving zero forgetting in continual learning across diverse tasks, model architectures, and learning paradigms.

In the context of training and adapting LLMs sequentially, the significance of CL is undergoing semantic shifts of its own as well. 
To highlight this ongoing shift, in this paper, we provide a comprehensive overview and detailed discussion of the current research progress on continual LLMs. 
For the general picture of continual LLMs, we for the first time divide it into two directions of continuity that need to be addressed by practitioners~(details in \Secref{sec:overview}):
\begin{itemize}
    \item \textbf{Vertical continuity (or vertical continual learning)}, which refers to the ongoing adaptation of LLMs as they transition from large-scale general domains to smaller-scale specific domains, involving shifts in learning objectives and entities of execution. 
    For example, healthcare institutions may develop LLMs tailored to the medical domain while retaining their general reasoning and question answering capabilities for users.
    \item \textbf{Horizontal continuity (or horizontal continual learning)}, which refers to continual adaptation across time and domains, often entails multiple training stages and increased vulnerability to forgetting. 
    For example, social media platforms continuously update LLMs to reflect recent trends, ensuring accurate targeting of downstream services like advertising and recommendations without compromised experience for existing users.
\end{itemize}
Importantly, separating vertical and horizontal CL transcends mere modification of existing paradigms, like domain-incremental learning, which aligns with horizontal continuity. This distinction offers a robust framework for analyzing complex CL paradigms in language models. For instance, Recyclable Tuning preserves both vertical and horizontal continuity simultaneously~\cite{qin2023recyclable}, and future designs might include zigzagging between horizontal and vertical CL.

In \Figref{fig:overview}, following \emph{vertical continuity}, we delineate three key stages of LLM learning within modern CL: Continual Pre-Training~(CPT), Domain-Adaptive Pre-training~(DAP), and Continual Fine-Tuning~(CFT)~(details in \Secref{sec:stages}). 
In CPT, existing research primarily investigates three types of distributional shifts: temporal, content-level, and language-level. Each presents distinct focuses and challenges.
In DAP, CL evaluation and techniques are frequently utilized. However, there is a noticeable lack of diversity in these techniques, considering the maturity of the conventional CL community.
In CFT, our focus is on the emerging field of learning LLMs, covering topics such as Continual Instruction Tuning (CIT), Continual Model Refinement (CMR), Continual Model Alignment (CMA), and Continual Multimodal LLMs (CMLLMs).
Next, we present a compilation of publicly available evaluation protocols and benchmarks~(details in \Secref{sec:eval-and-data}).
We conclude our survey with a discussion covering emergent properties of continual LLMs, changes in the roles of conventional CL types and memory constraints within the context of continual LLMs, and prospective research directions for this subject~(details in \Secref{sec:discussion}).

In summary, this survey provides a comprehensive review of existing continual learning studies for LLMs, which significantly distinguishes itself from existing literature on related topics~\cite{biesialska2020continual,ke2023continual,wang2024comprehensive,wu2024continual,yang2024recent}.
Our survey highlights the underexplored research area of continually developing LLMs, especially in the field of CPT and DAP. We emphasize the needs for increased attention from the community, including the development of practical, accessible, and widely acknowledged evaluation benchmarks. Additionally, methodologies need to be tailored to address forgetting in emerging LLM learning paradigms.
We hope this survey can provide a systematic and novel perspective of continual learning in the rapidly-changing field of LLMs and can help the continual learning community contribute to the challenging goals of developing LLMs in a more efficient, reliable, and sustainable manner~\cite{jang2022temporalwiki,su2023efficient,xie2023efficient,Cao2023InstructMol,attanasio2023worth}.


\section{Background and Related Work}
\label{sec:background}

\subsection{Large Language Models}
\label{sec:background-llm}
Primarily built on the transformer architecture, pre-trained language models~(PLMs) have established a universal hidden embedding space through extensive pre-training on large-scale unlabeled text corpora~\cite{devlin2018bert, liu2019roberta, raffel2020exploring}.
By scaling parameters to billions or even hundreds of billions and training on massive text datasets~\cite{kaplan2020scaling, hoffmann2022training}, PLMs not only demonstrate superior language understanding and generation capabilities but also manifest emergent abilities such as in-context learning, instruction following, and multi-step reasoning~\cite{wei2022chain,wei2022emergent,yao2024tree,wei2021finetuned,min2022rethinking}. 
These larger models are commonly referred to as Large Language Models~(LLMs). For more detailed introduction, please refer to \appref{app:preliminary-llm}.

\subsubsection{Pre-training of LLMs}
\label{sec:background-llm-pretraining}
There are two popular pre-training paradigms for LLMs. (1) \emph{Decoder-only models} typically employ auto-regressive language modeling (LM) tasks during pre-training, including the GPT family~\cite{radford2019language,brown2020language,achiam2022chatgpt,achiam2023gpt}, Gemini family~\cite{team2023gemini,reid2024gemini}, and the open-source Llama family~\cite{touvron2023llama,touvron2023llama2}.
Specifically, given a sequence of tokens $\vx = [x_1, x_2, \cdots, x_N ]$, LM predicts the next token $x_t$ autoregressively based on all preceding tokens $\vx_{<t} = [x_1, x_2, \cdots, x_{t-1}]$, and trains the entire network by minimizing the negative log-likelihood $-\sum^N_{t=1} \log P( x_t | \vx_{<t} )$,
where $P(x_1|\vx_{<1})\triangleq P(x_1)$ is the unconditional probability estimation of the first token.
(2) \emph{Encoder-only models}, e.g., BERT~\cite{devlin2018bert,liu2019roberta}, use masked language modeling (MLM) as a common pre-training objective. 
In MLM, for the input sequence $\vx$, a subset of input tokens $m(\vx)$ are masked and replaced with the special [MASK] token. The pre-training goal is to utilize the unmasked parts $\vx_{\backslash m(\vx)}$ to predict the masked portions $m(\vx)$. 
In summary, the overarching goal of MLM is to minimize the negative log-likelihood $-\sum_{\hat{x} \in m(\vx)}{\rm log} \, P( \hat{x}|\vx_{\backslash m(\vx)} )$.

\subsubsection{Adaptation of LLMs}
\label{sec:background-llm-adaptation}
LLMs are primarily trained to generate linguistically coherent text. However, this training may not align with human values, preferences, or practical needs. Furthermore, the pre-training data can be outdated, leading to knowledge cutoffs or inaccuracies. To address these issues, various computational paradigms such as Instruction Tuning~(IT)~\cite{zhang2024instruction}, Model Refinement~(MR)~\cite{de2021editing}, and Model Alignment~(MA)~\cite{ouyang2022rlhf,rafailov2024dpo} have been proposed. These approaches adapt LLMs to better meet diverse downstream tasks and user requirements.

Numerous studies show that \textbf{Instruction Tuning~(IT)} can notably improve LLMs' ability to follow textual instructions~\cite{zhang2024instruction,wei2021finetuned,jiang2024instructiontuned,sanh2022multitask,ouyang2022rlhf}, leveraging the pre-existing knowledge within LLMs to bridge the gap between general and task-specific performance~\cite{wei2022finetuned}. Recent works like WizardLM~\cite{xu2023wizardlm} and CodecLM~\cite{wang2024codeclm} further tailor synthetic data to steer LLMs' behavior through IT.  Additionally, IT enhances the interaction between humans and LLMs, providing a more natural interface and aligning LLM outputs more closely with human expectations and preferences~\cite{luo2023empirical}.
LLMs make mistakes, such as inaccurate translations or outdated information~\cite{de2021editing}. 
Directly fine-tuning the model to correct these mistakes may disrupt its performance on previously learned tasks. To overcome these challenges, \textbf{Model Refinement~(MR)} is proposed to rectify the model's errors while preserving its performance on other inputs, with only moderate computing resources~\cite{sinitsin2020editable,de2021editing,fast_edit,hase2021language,huang2023transformer,mitchell2022memory,hartvigsen2023aging}.
\textbf{Model Alignment~(MA)} ensures AI systems' actions and outputs align with human values, ethics, and preferences~\cite{ouyang2022rlhf,rafailov2024dpo}.
MA can be broadly categorized into two types: Reinforcement Learning-based (RL-based) and Supervised Learning-based (SL-based). RL-based approaches~\cite{ouyang2022rlhf,schulman2017proximal} are trained to make decisions reinforced by human feedback, using a reward system to guide them towards desirable outcomes. In contrast, SL-based approaches~\cite{hendrycks2023aligning, rafailov2024dpo,ji2024ai} directly train models on datasets of human preferences, aligning their output with demonstrated human values.

\subsection{Continual Learning}
\label{sec:background-cl}
Humans can accumulate knowledge and skills across tasks without significant performance decline on previous tasks \cite{mcclelland1995there,kandel2000principles,pallier2003brain,mccaffary2021towards}. In contrast, machine learning models, which are typically data-centric, often experience performance degradation on old tasks when trained on new ones, a phenomenon known as \emph{``catastrophic forgetting.''} The challenge of adapting models to a sequence of tasks without forgetting, especially when little to no past data can be preserved, is extensively studied in the continual learning community \cite{pentina2016theoretical,chen2018lifelong,van2022three,wang2024comprehensive}. For formal definitions, a detailed introduction to the three CL scenarios and techniques, please refer to \appref{app:preliminary-cl}.

\subsubsection{Types of Continual Learning}
\label{sec:background-cl-types}
To lay the groundwork for subsequent discussions (as illustrated in \Tabref{tab:cft} and \Secref{sec:discussion-xil}), we follow the conceptual framework proposed by \cite{van2022three,kim2022theoretical,wang2024comprehensive}. There are three primary types of continual learning scenarios: (i) Task-Incremental Learning~(TIL), where task indices are available to the model during inference~\cite{li2017learning,kirkpatrick2017overcoming}; (ii) Domain-Incremental Learning~(DIL), where the model learns a sequence of tasks with the same formulation but without task indices during inference~\cite{shi2024unified}; and (iii) Class-Incremental Learning~(CIL), where the model learns new classes of data during training~\cite{rebuffi2017icarl,kim2022theoretical}.

\subsubsection{Techniques of Continual Learning}
\label{sec:background-cl-techs}
Existing CL techniques can be roughly categorized into five groups~\cite{wang2024comprehensive}: (i)~replay-based, (ii)~regularization-based, (iii)~architecture-based, (iv)~optimization-based, and (v)~representation-based.
Here, we provide a concise yet comprehensive introduction to the first three categories of continual learning techniques, as they are extensively applied in continual LLMs.

\textbf{Replay-based methods} adopt the relaxed memory constraint by keeping a small buffer of observed data and retraining the model on it when learning new tasks. 
Although replay-based methods may theoretically lead to loose generalization bounds~\cite{shi2024unified}, they are valued for their simplicity, stability, and high performance, even with a small episodic memory~\cite{chaudhry2019tiny,riemer2018learning,buzzega2020dark,rebuffi2017icarl}.
\textbf{Regularization-based methods} adopt a regularization term $\lambda \left\| \vtheta - \vtheta_{t-1}\right\|_\mSigma$ that penalizes large deviation from the history model in the parameter space,
where $\|\vv\|_\mSigma = \vv^\top \mSigma \vv$ is the vector norm evaluated on a positive-semi-definite matrix $\mSigma$, and $\lambda$ is the regularization coefficient, a hyper-parameter introduced to balance the past knowledge retention and current knowledge learning. 
The matrix $\mSigma$ introduced is to measure the different level of importance of each parameters and their correlations in retaining the past knowledge. 
In practice, to reduce computational overhead, diagonal matrices are often designed to encode only the importance of each parameter~\cite{kirkpatrick2017overcoming,aljundi2018memory,rongali2021continual}.
\textbf{Architecture-based methods}, especially expanding the network architecture dynamically to assimilate new knowledge, is considered the most efficient form of CL~\cite{wang2022learning,wang2022dualprompt}. This method primarily tackles adaptation challenges and can achieve zero-forgetting when task IDs are available during inference or can be correctly inferred~\cite{gururangan2022demix,wistuba2023}. 
However, due to the difficulty of task ID inference, architecture expansion is predominantly utilized in TIL but is scarcely explored in DIL or CIL.
In conjunction with pre-trained backbone large models like ViT~\cite{dosovitskiy2020image}, CoLoR~\cite{wistuba2023} trains various low-rank adaptation (LoRA)~\cite{hu2021lora} modules for different tasks. It estimates and stores prototypes for each task and utilizes the natural clustering ability of the pre-trained model during testing to infer task IDs, selecting the corresponding LoRA component for prediction generation.
In the domain of continual LLMs, architecture expansion has resurged in popularity following the rise of parameter-efficient fine-tuning (PEFT)~\cite{shazeer2017outrageously,hu2021lora,dettmers2023qlora}, a topic we will delve into shortly~\cite{yang2024moral,wang2023orthogonal,li2024examining,jang2022towards,jin2022lifelong,paul2024ircoder,yan2023af,wu2024llama}.

\subsubsection{Evaluation Metrics of Continual Learning}
\label{sec:background-cl-eval}
There are four evaluation protocols primarily designed for continual learning. \textbf{Overall Performance~(OP)}~\cite{ke2021achieve,zhang2022continual,zhang2023copf} calculates the average performance up until the current training stage, measuring the overall ability of a model balancing the performance of each task. As noted in \cite{shi2024unified}, OP corresponds to the primary optimization objective of continual learning, and hence receives the most attention.
\textbf{Forgetting~(F)} represents the largest performance drop observed of each task throughout the training process, averaged over all training stages. 
It quantifies the negative impact of learning new tasks brought to previously acquired knowledge. Ideally, a robust continual learning framework should achieve \textbf{Backward Transfer~(BWT)}, where learning new tasks enhances performance on prior tasks. 
BWT is measured by negating the forgetting, and hence a negative forgetting indicates a an improvement in performance on earlier tasks. 
\textbf{Forward Transfer (FWT)} measures the generalization ability of the continual learning algorithms to unseen tasks. It is defined as the difference between the current model's performance evaluated on the future tasks and the randomly initialized model. 
Refer to \appref{app:eval} for more details. 

\section{Continual Learning Meets Large Language Models: An Overview}
\label{sec:overview}

Large language models (LLMs) are extensive in various dimensions, including the size of model parameters, pre-training datasets, computational resources, project teams, and development cycles~\cite{radford2019language,brown2020language,achiam2022chatgpt,achiam2023gpt,chowdhery2023palm,anil2023palm,touvron2023llama,touvron2023llama2}. The substantial scale of LLMs presents notable challenges for development teams, particularly in keeping them updated amidst rapid environmental changes~\cite{amba2021dynamic,jin2022lifelong,dhingra2022time,jang2022towards,jang2022temporalwiki}. To illustrate, in 2023, the average daily influx of new tweets exceeds 500 million\footnote{Source: \href{https://www.omnicoreagency.com/twitter-statistics}{https://www.omnicoreagency.com/twitter-statistics} }, and training on even a subset of this large volume of data is unaffordable. 
Recyclable Tuning~\cite{qin2023recyclable} is the first work to explicitly outline the supplier-consumer structure in the modern LLM production pipeline.
On the supplier side, the model is continually pre-trained over a sequence of large-scale unlabeled datasets. After every release of the pre-trained model, the consumer utilizes the stronger and more up-to-date upstream model for downstream tasks. 
Compared to the upstream supplier, downstream users often lack capacity of collecting and storing large-scale data, maintaining large-scale hardware systems, and training LLMs themselves. 
In this survey, we extend this framework and further present a comprehensive modern production pipeline encompassing various studies on continual LLM pre-training, adaptation, and deployment~(\Figref{fig:overview}). What sets our framework apart from existing studies~\cite{wu2024continual} is the \emph{incorporation of two directions of continuity: \textbf{Vertical Continuity} and \textbf{Horizontal Continuity}}.



\subsection{Vertical Continuity (Vertical Continual Learning)}
\begin{figure*}[t]
	\begin{center}
		\includegraphics[width=1\textwidth]{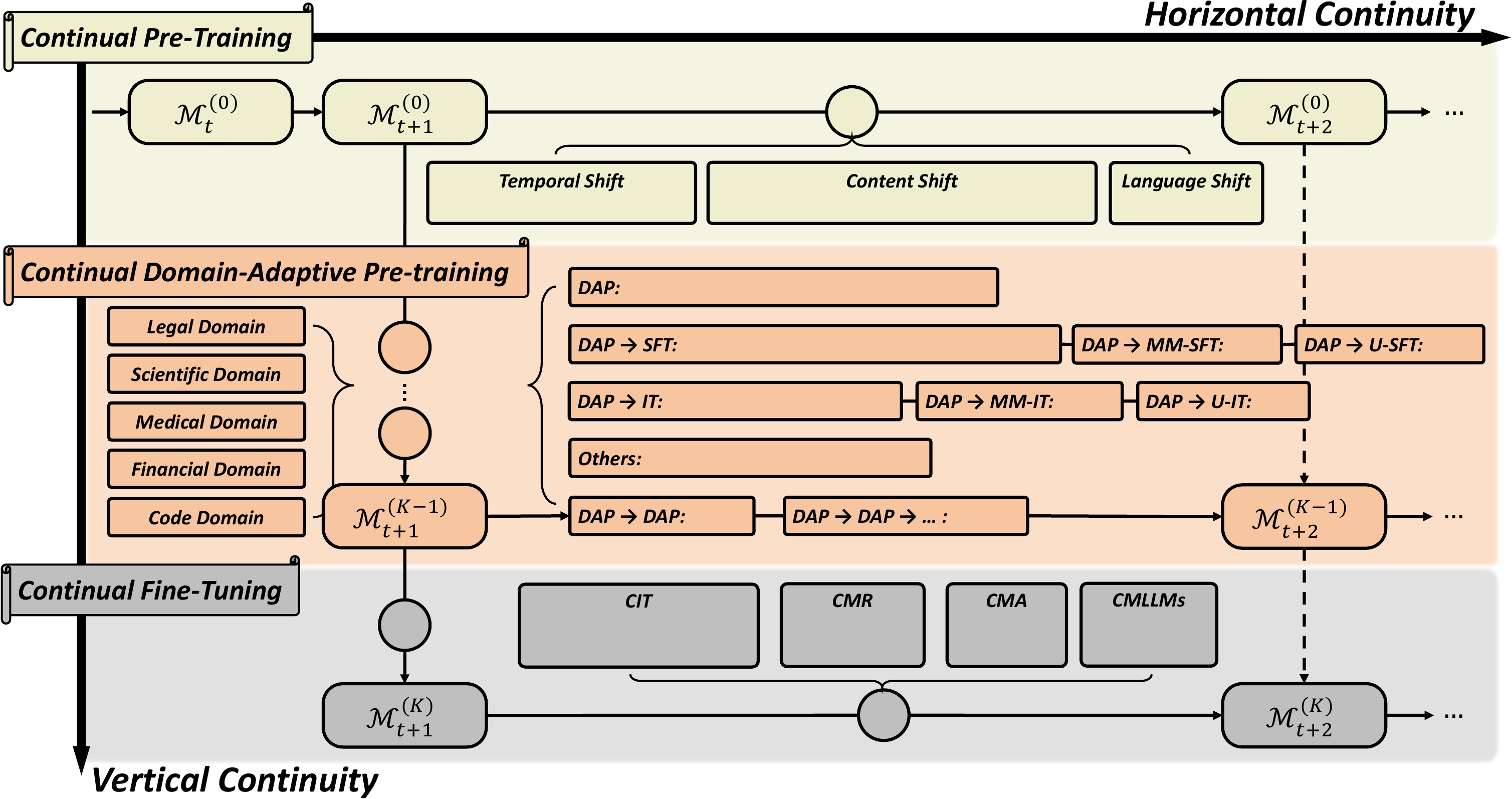}
	\end{center}
	\caption{
        A high-level overview of the modern pipeline for continually pre-training and fine-tuning LLMs, where two dimensions of continuity are described. 
        \textbf{Vertical Continuity~(or Vertical Continual Learning):} LLM training can be vertically divided into three stages: (i)~Continual Pre-Training~(CPT), (ii)~Domain-Adaptive Pre-training~(DAP), and (iii)~Continual Fine-Tuning~(CFT). 
        The main focus is the retention of the LLM's general knowledge~(prevention of vertical forgetting).
        \textbf{Horizontal Continuity~(or Horizontal Continual Learning):} After the LLMs are deployed, the models are continually updated when a new set of data becomes available. 
        The primary goal is to prevent horizontal forgetting in a long sequence of tasks.
    }
	\label{fig:overview}
    \vspace{-1em}
\end{figure*}

\textbf{Definition.}\quad 
Vertical continuity (or vertical continual learning) has long been studied, either implicitly or explicitly, in existing literature.
Vertical continuity is characterized by a hierarchical structure encompassing data inclusiveness, task scope, and computational resources. Specifically, the training task transitions gradually from general pre-training to downstream tasks, typically undertaken by distinct entities within the production pipeline~\cite{qin2023recyclable,gururangan2022demix,rongali2021continual,guo2023continuous,yan2023af,xie2023efficient}. 
\Figref{fig:overview} shows a typical pipeline for vertical continuity in LLMs, i.e., ``pre-training'' $\rightarrow$ ``domain-adaptive training'' $\rightarrow$ ``downstream fine-tuning''~\cite{luo2023biomedgpt,li2023cfgpt,deng2023learning,han2021econet,zhou2020pre,guo2023continuous,gururangan2020dont,colombo2024saullm7b,wu2023pmc,wu2024llama,yan2023af,rongali2021continual,ma2023ecomgptct,huang2023lawyer}: 
\begin{itemize}
\item \textbf{Pre-training.} During the \emph{pre-training} stage, a substantial amount of data from diverse domains is required to develop a general-purpose LLM. This phase demands a sizable research and development team dedicated to training and benchmarking the model, along with considerable computational resources. 
\item \textbf{Domain-Adaptive Pre-training.} Subsequently, downstream institutions may opt for \emph{domain-adaptive pre-training} to tailor the model for specific tasks using domain-specific data unavailable to the upstream supplier. 
\item \textbf{Finetuning.} Finally, the LLM undergoes \emph{fine-tuning} on annotated data for downstream tasks before deployment.
\end{itemize}

\placetextbox{0.44}{0.818}{\tiny{\cite{loureiro2022timelms,jang2022towards,jin2022lifelong,jang2022temporalwiki,amba2021dynamic,dhingra2022time,yildiz2024investigating}}} 
\placetextbox{0.596}{0.818}{\tiny{\cite{gupta2023continual,ke2022continual-train,jin2022lifelong,sun2020ernie,cossu2022continual,gururangan2022demix,qin2023recyclable,chen2023lifelong,qin2022elle,ibrahim2024simple}}} 
\placetextbox{0.73}{0.818}{\tiny{\cite{gogoulou2024continual,li2024examining,ibrahim2024simple}}} 

\placetextbox{0.55}{0.7893}{\tiny{\cite{Lu2023BBTFin,Nguyen2023AstroLLaMA,xie2023efficient,li2023starcoder,guo2024deepseekcoder,Xue2023WeaverBird,paul2024ircoder,cheng2024adapting,dou2024sailor,fujii2024continual,takahashi2024pretraining}}} 
\placetextbox{0.56}{0.7694}{\tiny{\cite{li2023cfgpt,deng2023learning,han2021econet,zhou2020pre,guo2023continuous,gururangan2020dont,yan2023af,rongali2021continual,ma2023ecomgptct}}} 
\placetextbox{0.76}{0.7694}{\tiny{\cite{luo2023biomedgpt}}} 
\placetextbox{0.852}{0.7694}{\tiny{\cite{wu2024llama}}} 

\placetextbox{0.535}{0.7486}{\tiny{\cite{Bi2023OCEANGPT,wu2023pmc,Yang2023PLLaMa,li2024blade,xie2024me,nakamura2024aurora}}} 
\placetextbox{0.68}{0.7486}{\tiny{\cite{Zheng2023MarineGPT}}} 
\placetextbox{0.77}{0.7486}{\tiny{\cite{colombo2024saullm7b}}} 

\placetextbox{0.535}{0.7284}{\tiny{\cite{song2024code,lin2023geogalactica,rozière2024code,Chen2023HuatuoGPTII,Zhang2023xuanyuan,huang2023lawyer,thulke2024climategpt,acikgoz2024hippocrates}}} 
\placetextbox{0.51}{0.7077}{\tiny{\cite{nijkamp2022codegen}}} 
\placetextbox{0.63}{0.7077}{\tiny{\cite{ke2022continual-pre}}} 

\placetextbox{0.476}{0.661}{\tiny{\cite{scialom2022fine,wang2023trace,mok2023large,zhang2023citb,huang2024mitigating}}} 
\placetextbox{0.476}{0.668}{\tiny{\cite{he2024dont,yin2022contintin,wang2023orthogonal,zhao2024sapt}}} 
\placetextbox{0.575}{0.661}{\tiny{\cite{lin2022continual,hartvigsen2023aging,hu2024wilke,wang2024wise}}} 
\placetextbox{0.575}{0.668}{\tiny{\cite{das2024larimar,yu2023melo,li2023continual}}} 
\placetextbox{0.647}{0.661}{\tiny{\cite{zhang2023copf,zhangcppo}}} 
\placetextbox{0.647}{0.668}{\tiny{\cite{lin2024mitigating}}} 
\placetextbox{0.713}{0.661}{\tiny{\cite{he2023continual,zheng2024antiforgetting,chen2024coin}}} 
\placetextbox{0.713}{0.668}{\tiny{\cite{zhu2024model,zhao2024reconstruct}}} 

Throughout the process, the unlabeled domain-specific dataset is smaller in scale than the upstream pre-training phase but larger than the final downstream task fine-tuning phase. This pattern extends to computational resources, team size, and other factors. It is important to note that vertical continuity can involve more than three stages~\cite{nijkamp2022codegen,lin2023geogalactica,rozière2024code,huang2023lawyer}. In real-world applications, during domain-adaptive pre-training, additional ``layers'' can be added to accommodate multiple entities, such as various departments with distinct objectives but operating within the same domain.

\textbf{Vertical Forgetting.}\quad
We term the performance degradation (in terms of general knowledge) 
due to 
vertical continual learning \emph{``vertical forgetting''}. As shown in \Figref{fig:continuity}, for vertical continual learning, the data distribution of upstream tasks partially covers the downstream, meaning the model might start off at a decent initialization for the subsequent stage of training. 
Two significant challenges must be addressed to prevent vertical forgetting: 
\begin{itemize}
    \item \textbf{Task Heterogeneity.} Stemming from the inherent disparity between the formulation of upstream tasks and downstream tasks, \emph{task heterogeneity} can lead to differences in model structures and training schemes, which has long been recognized as a major hurdle~\cite{rebuffi2017icarl,li2017learning,wu2019large,ni2021revisiting,kim2022theoretical}. To mitigate this issue, practitioners often employ methodologies such as freezing shared parameters during downstream phases or reformulating downstream tasks to match the structure of pre-training tasks~\cite{yang2024moral,wang2023orthogonal,li2024examining,paul2024ircoder,yan2023af,wu2024llama}.
    \item \textbf{Inaccessible Upstream Data.} This challenge arises primarily from varying levels of confidentiality across entities undertaking vertical continual learning. Data collected and curated under different protocols may not be accessible to some downstream entities. This scenario is even more challenging than the strict memory constraint presented in conventional CL~(\Defref{def:memory}),
    as algorithms for latter case rely on access to previous data at specific points for parameter importance measurement~\cite{kirkpatrick2017overcoming,aljundi2018memory} or for replay~\cite{riemer2018learning,chaudhry2019tiny,buzzega2020dark,shi2024unified}.
    To address the challenge of \emph{inaccessible upstream data}, existing methods either use public datasets or generate pseudo-examples to create proxy pre-training datasets~\cite{qin2021lfpt5}.
\end{itemize}

\subsection{Horizontal Continuity (Horizontal Continual Learning)}
\textbf{Definition.}\quad 
Horizontal continuity (or horizontal continual learning) refers to continual adaptation across time and domains, a topic extensively explored within the continual learning community. The primary rationale for preserving horizontal continuity lies in the dynamic nature of data distribution over time. To stay updated with these content shifts, an LLM must incrementally learn newly-emerged data. Otherwise, the cost of re-training will become prohibitively expensive and impractical~\cite{chaudhry2019efficient,amba2021dynamic,su2023efficient,xie2023efficient}.
Empirical evidence has consistently shown that despite their impressive capabilities, LLMs struggle to generalize effectively to future unseen data, particularly in the face of temporal or domain shifts~\cite{amba2021dynamic,jang2022towards,jang2022temporalwiki,dhingra2022time}. Additionally, they struggle to retain complete knowledge of past experiences when adapting to new temporal domains, although they do demonstrate a higher level of robustness against catastrophic forgetting~\cite{tao2022can,luo2023investigating,zheng2023learn,mehta2023empirical}.
The necessity of employing complex CL algorithms to address challenges in LLMs remains an open question. For instance, during large-scale continual pre-training, large institutions can typically afford the storage costs of retaining all historical data, rendering memory constraints meaningless. 
Several studies have demonstrated that with full access to historical data, simple sparse replay techniques can effectively mitigate forgetting~\cite{tao2022can,scialom2022fine,prabhu2023online,garg2024tic}. In contrast, numerous continual learning studies have showcased superior performance compared to naive solutions, suggesting the importance of continual learning techniques in LLM training~\cite{jang2022temporalwiki,jin2022lifelong,qin2022elle,chen2023lifelong}. 

\begin{figure*}[t]
	\begin{center}
		\includegraphics[width=0.9\textwidth]{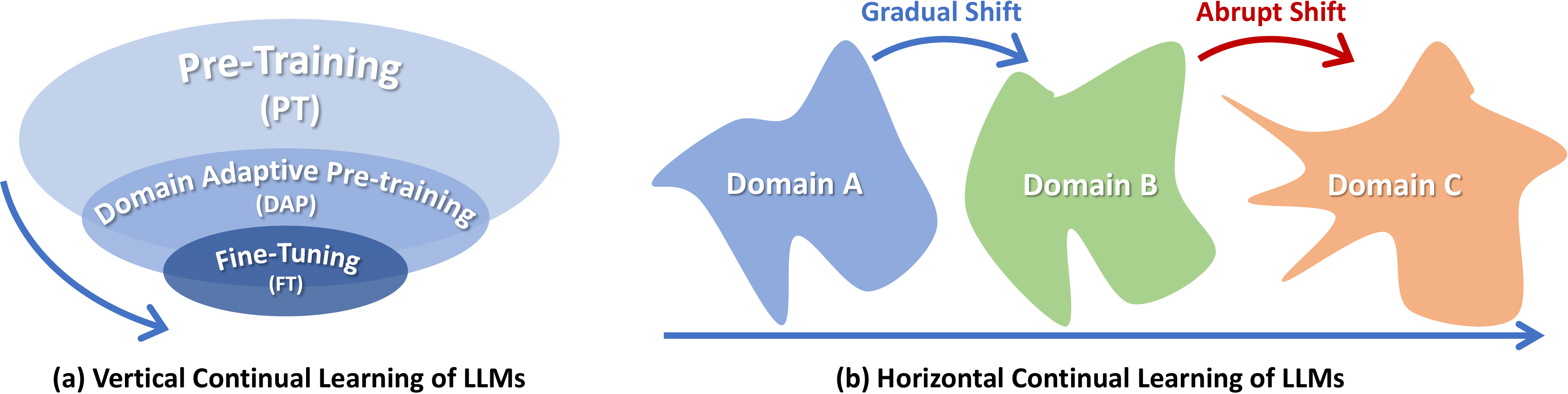}
	\end{center}
	\caption{
        A diagram showing two different directions of continual learning of LLMs. \textbf{(a) Vertical Continual Learning of LLMs:} in this case, the upstream data distribution usually partially covers the subsequent tasks' data distribution. \textbf{(b) Horizontal Continual Learning of LLMs:} No constraints on the data distributions are present on horizontal continual learning. The continual LLMs need to handle the challenge of abrupt distributional shifts and a longer sequence of training.
    }
	\label{fig:continuity}
    \vspace{-1em}
\end{figure*}

\textbf{Horizontal Forgetting.}\quad 
We informally define \emph{``horizontal forgetting''} as the performance degradation on the previous tasks when model is undergoing horizontal continual learning. As illustrated in \Figref{fig:continuity}, horizontal continual learning typically involves training stages of similar scales, with potential distributional overlap among their data. In summary, two main challenges need to be addressed for horizontal continual learning of LLMs:
\begin{itemize}
    \item \textbf{Long Task Sequences.}
    Horizontal continual learning ideally involves numerous incremental phases, particularly to accommodate temporal shifts in data distribution. A \emph{longer task sequence} entails more update steps of the model, leading to inevitable forgetting of previously learned tasks. To address this challenge, researchers employ established continual learning techniques with stronger constraints, such as continual model ensemble~\cite{ramesh2021model}.
    \item \textbf{Abrupt Distributional Shift.} 
    In contrast to vertical continuity, where distributional shifts are often predictable, horizontal continual learning does not impose constraints on task properties. Evidence suggests that abrupt changes in task distributions can result in significant horizontal forgetting of the model~\cite{sarfraz2023error}.
\end{itemize}

\section{Learning Stages of Continual Large Language Models}
\label{sec:stages}
\Figref{fig:overview} provides an overview of continually learning LLMs. Along the axis of vertical continuity, three main ``layers'' of modern continual learning emerge. 
The top layer, Continual Pre-Training~(CPT), involves continuous pre-training of LLMs by the supplier on newly-collected data alongside existing data (\Secref{sec:cpt}). 
The middle layer, Domain-Adaptive Pre-training~(DAP), prepares LLMs for domain-specific applications through additional pre-training on domain-specific \emph{unlabeled} data (\Secref{sec:dap}). 
The bottom layer, Continual Fine-Tuning~(CFT), targets models for final downstream tasks on the consumer side (\Secref{sec:cft}), where the model needs to be updated after deployment for the specified task. 

\subsection{Continual Pre-Training~(CPT)}
\label{sec:cpt}


\subsubsection{CPT: Effectiveness and Efficiency}
Before delving into the details of continual pre-training (CPT), it is important to address two fundamental questions: Firstly, regarding \emph{effectiveness}, can CPT enhance performance on downstream tasks beyond that of the initial training on a wide range of data domains? Extensive studies
have not only demonstrated the necessity of CPT for improved downstream performance~\cite{qin2022elle,gururangan2022demix,jang2022towards,jang2022temporalwiki,jin2022lifelong,chen2023lifelong}, but also shown that when distributional shifts are gradual~\cite{jang2022temporalwiki,yildiz2024investigating} or somewhat correlated~\cite{gururangan2022demix}, CPT can effectively help model generalize to unseen data.
The second question is about \emph{efficiency}: given the large size of an LLM' parameters and data, both old and new, can we achieve adaptation and knowledge retention in a computationally efficient way? Concerning efficiency, most studies focus on techniques for efficient knowledge retention~\cite{jin2022lifelong,jang2022towards,jang2022temporalwiki,li2024examining}, which significantly overlap with the CL literature addressing catastrophic forgetting~\cite{schwarz2018progress,riemer2018learning,buzzega2020dark,shi2024unified,rebuffi2017icarl,ritter2018online,aljundi2018memory,rusu2016progressive,ramesh2021model,wang2022coscl}. 
In contrast to prior approaches that fully utilize emergent data, some studies recognize the impracticality of this approach in real production environments. Instead, they concentrate on further improving the efficiency of adaptation. For instance, ELLE~\cite{qin2022elle} employs a function-preserved model expansion to facilitate efficient knowledge growth; \cite{amba2021dynamic} and \cite{xie2023efficient} sub-sample training data based on novelty and diversity to enhance training efficiency, achieving superior performance compared to full-data training. Though currently underexplored, efficient adaptation in continual pre-training is poised to become significant, given recent findings emphasizing data quality over quantity for LLM generalization~\cite{xie2024data,soldaini2024dolma}.

\begin{table*}[t]
\caption{
    \textbf{Summary of existing studies on Continual Pre-training of LLMs.} The papers are organized based on their relation to CL: (i)~no CL techniques are studied, (ii)~CL techniques are studied as solely baselines, and (iii)~new CL approaches are proposed. 
    In the table, \emph{Dist.~Shift} denotes what type(s) of distributional shifts this particular study considers and is dedicated to solve.
    In the section of \textbf{Continual Learning Tech.}, we mainly categorize three types of continual learning techniques that are studied in the paper: rehearsal~(\emph{Rehearsal}), parameter regularization~(\emph{Param.~Reg.}), and architecture expansion~(\emph{Arch. Exp.}). We use ``\cmark'', ``\xmark'', and ``\club'' to denote ``deployed in the proposed method'', ``not studied in the paper'', and ``studied as a baseline method'', respectively. 
    Note that we do not include naive sequential fine-tuning in this table, as it is universally studied as the important baseline method in all of the papers in the table. 
    The papers with only ``\club''~\cite{jin2022lifelong,jang2022temporalwiki,jang2022towards} means that 
    only existing CL techniques are studied, without proposing new ones, 
    and the papers with only ``\xmark''~\cite{gupta2023continual,gogoulou2024continual} means that 
    special aspects of fine-tuning are studied, without using CL techniques. 
}
\vspace{-0.5em}
\begin{center}
\resizebox{1\linewidth}{!}{%
\setlength{\tabcolsep}{9pt}
\begin{tabular}{cCCCC CCcc}
	\toprule[0.12em]
	\multirow{2}{*}{\textbf{Method}} & 
    \multicolumn{2}{c}{\textbf{Scenario}} & 
    \multicolumn{3}{c}{\textbf{{Continual Learning Tech.}}} & 
    \multirow{2}{*}{\textbf{{LLM Arch.}}} & 
    \multicolumn{2}{c}{\textbf{{Evaluation}}} \\
    \cmidrule(lr){2-3}\cmidrule(lr){4-6}\cmidrule(lr){8-9}
    & \emph{Dist. Shift} & \emph{\#Domains} & \emph{Rehearsal} & \emph{Param. Reg.} & \emph{Arch. Exp.} & & \emph{Pre-Training} & \emph{Downstream} \\
    \cmidrule[0.12em]{1-9}
    
    TimeLMs~\cite{loureiro2022timelms} & Temporal & 8 & \xmark & \xmark & \xmark & RoBERTa & \cmark & \cmark \\
    \hline

    \cite{yildiz2024investigating} 
    & Content & 159 
    & \xmark 
    & \xmark 
    & \xmark
    & RoBERTa GPT-2
    & \cmark 
    & \xmark \\
    \hline 

    \cite{gupta2023continual}  & Content & 1 & \xmark & \xmark & \xmark & Pythia & \cmark & \xmark \\
    \hline

    \cite{gogoulou2024continual} & Language & 3 & \xmark & \xmark & \xmark & GPT & \cmark & \xmark \\
    \hline

    RHO-1~\cite{lin2024rho} 
    & Other
    & 1
    & \xmark 
    & \xmark 
    & \xmark
    & TinyLlama Mistral
    & \cmark & \cmark
    \\

    \midrule
    \midrule

    \cite{li2024examining} 
    & Language & 1 
    & \cellcolor{gray2}\xmark 
    & \cellcolor{gray2}P-Freeze$^\text{\club}$ 
    & \cellcolor{gray2}Adapter$^\text{\club}$ LoRA$^\text{\club}$ 
    & Llama2 
    & \cmark & \cmark \\
    \hline

    CKL~\cite{jang2022towards} 
    & Temporal & 1 
    & \cellcolor{gray2} Mix-Review$^\text{\club}$ 
    & \cellcolor{gray2} P-Freeze$^\text{\club}$ RecAdam$^\text{\club}$  
    & \cellcolor{gray2} LoRA$^\text{\club}$ K-Adapter$^\text{\club}$ 
    & T5 
    & \xmark & \cmark \\
    \hline

    LLPT~\cite{jin2022lifelong} 
    & \makecell{Temporal \\ \rule[0.5ex]{1\linewidth}{0.4pt} \\ Content} 
    & \makecell{4 \\ \rule[0.5ex]{1\linewidth}{0.4pt} \\ 8} 
    & \cellcolor{gray2} ER$^\text{\club}$ Logit-KD$^\text{\club}$ Rep-KD$^\text{\club}$ Contrast-KD$^\text{\club}$ SEED-KD$^\text{\club}$ 
    & \cellcolor{gray2} oEWC$^\text{\club}$ 
    & \cellcolor{gray2} Adapter$^\text{\club}$ Layer~Exp.$^\text{\club}$ 
    & RoBERTa 
    & \cmark & \cmark \\
    \hline

    TemporalWiki~\cite{jang2022temporalwiki} 
    & Temporal & 5 
    & \cellcolor{gray2} Mix-Review$^\text{\club}$ 
    & \cellcolor{gray2} P-Freeze$^\text{\club}$ RecAdam$^\text{\club}$  
    & \cellcolor{gray2} LoRA$^\text{\club}$ K-Adapter$^\text{\club}$ 
    & GPT-2 
    & \cmark & \cmark\\
    \midrule
    \midrule

    CPT$^*$~\cite{ke2022continual-train} 
    & Content & 4 
    & \cellcolor{gray3} DER++$^\text{\club}$\qquad KD$^\text{\club}$ 
    & \cellcolor{gray3}CPT$^\text{\cmark}$\qquad EWC$^\text{\club}$\qquad HAT$^\text{\club}$ 
    & \cellcolor{gray3}Adapter$^\text{\club}$ DEMix$^\text{\club}$ 
    & RoBERTa 
    & \cmark & \xmark \\
    \hline

    ERNIE~2.0~\cite{sun2020ernie} 
    & Content & 4 
    & \cellcolor{gray3} ER$^\text{\cmark\club}$ 
    & \cellcolor{gray3} \xmark 
    & \cellcolor{gray3} \xmark 
    & ERNIE 
    & \xmark & \cmark \\
    \hline

	\cite{amba2021dynamic} 
    & Temporal & 7
    & \cellcolor{gray3} \xmark 
    & \cellcolor{gray3} P-Freeze$^\text{\cmark}$ 
    & \cellcolor{gray3} Vocab. Exp.$^\text{\cmark}$ 
    & BERT 
    & \xmark & \cmark \\
    \hline

    \cite{cossu2022continual} 
    & Content & 5 
    & \cellcolor{gray3} \xmark 
    & \cellcolor{gray3} \xmark 
    & \cellcolor{gray3} Vocab. Exp.$^\text{\cmark}$ 
    & BERT\qquad RoBERTa 
    & \xmark & \cmark \\
    \hline

    DEMix~\cite{gururangan2022demix} 
    & Content & 8 
    & \cellcolor{gray3} \xmark 
    & \cellcolor{gray3} \xmark 
    & \cellcolor{gray3} MoE$^\text{\cmark}$ 
    & GPT-3 
    & \cmark & \cmark \\
    \hline

    TempoT5~\cite{dhingra2022time} 
    & Temporal & 1 
    & \cellcolor{gray3} \xmark 
    & \cellcolor{gray3} \xmark 
    & \cellcolor{gray3} Vocab. Exp.$^\text{\cmark}$ Prompt$^\text{\cmark}$ 
    & T5 
    & \xmark & \cmark \\
    \hline

    RecTuning~\cite{qin2023recyclable} 
    & Content & 4 
    & \cellcolor{gray3} ER$^\text{\cmark}$\qquad \qquad KD$^\text{\cmark}$ 
    & \cellcolor{gray3} \xmark 
    & \cellcolor{gray3} Adapter$^\text{\cmark}$ 
    & RoBERTa  
    & \xmark & \cmark\\
    \hline

    Lifelong-MoE~\cite{chen2023lifelong} 
    & Content & 3 
    & \cellcolor{gray3} ER$^\text{\club}$\qquad \qquad KD$^\text{\cmark}$ 
    & \cellcolor{gray3} P-Freeze$^\text{\cmark}$\qquad L2$^\text{\club}$ 
    & \cellcolor{gray3} MoE$^\text{\cmark}$ 
    & GLaM 
    & \cmark & \cmark\\
    \hline

    ELLE~\cite{qin2022elle} 
    & Content & 5 
    & \cellcolor{gray3} ER$^\text{\cmark\club}$\qquad KD$^\text{\club}$ 
    & \cellcolor{gray3} P-Freeze$^\text{\cmark}$ 
    & \cellcolor{gray3} Prompt$^\text{\cmark}$\qquad Layer~Exp.$^\text{\cmark}$ Adapter$^\text{\club}$ 
    & BERT\qquad \qquad GPT 
    & \cmark & \cmark \\
    \hline

    \cite{ibrahim2024simple} 
    & Content Language & 2
    & \cellcolor{gray3} ER$^\text{\cmark}$ 
    & \cellcolor{gray3} \xmark
    & \cellcolor{gray3} \xmark
    & GPT-NeoX
    & \cmark & \cmark\\
    \hline

    CEM~\cite{zhao2024large} 
    & Other
    & 1 
    & \cellcolor{gray3} ER$^\text{\cmark}$
    & \cellcolor{gray3} \xmark
    & \cellcolor{gray3} \xmark
    & CuteGPT ChatGLM Qwen-Chat
    & \xmark
    & \cmark
    \\
    \hline
    
    IR-DRO~\cite{chen2024take}
    & Other
    & 1
    & \cellcolor{gray3} ER$^\text{\cmark}$
    & \cellcolor{gray3} \xmark
    & \cellcolor{gray3} \xmark
    & OPT
    & \xmark
    & \cmark 
    \\
    
	\bottomrule[0.12em]
	\end{tabular}
	}
\end{center}
\vspace{-1.5em}
\label{tab:cpt-small}
\end{table*}

\subsubsection{General Observations on CPT}
\Tabref{tab:cpt-small} summarizes the existing studies on continual pre-training~(CPT), and here are some key observations we make about CPT.
\begin{itemize}
    \item \textbf{OBS-1: The development of advanced techniques tailored specifically for CPT is at the starting stage and warrants further exploration.} 
    Only about half of the examined papers propose novel techniques for CPT~\cite{ke2022continual-train,sun2020ernie,amba2021dynamic,cossu2022continual,gururangan2022demix,dhingra2022time,qin2023recyclable,chen2023lifelong,qin2022elle}, while the remaining half either focus solely on the effects of pure adaptation without considering CL techniques~\cite{loureiro2022timelms,gupta2023continual,gogoulou2024continual}, or conduct empirical studies on the straightforward application of existing CL techniques~\cite{li2024examining,jin2022lifelong,jang2022towards,jang2022temporalwiki}.
    \item \textbf{OBS-2: The diversity of CL techniques incorporated in CPT remains limited.} 
    Most practical implementations of CL techniques for CPT primarily focus on architecture expansion of LLMs~\cite{amba2021dynamic,cossu2022continual,gururangan2022demix,dhingra2022time,qin2023recyclable,chen2023lifelong}, with only a few explicitly utilizing replay~\cite{qin2023recyclable,chen2023lifelong} and parameter regularization~\cite{amba2021dynamic,chen2023lifelong}.
    \item \textbf{OBS-3: There is an apparent gap between the existing studies and the real production environment of CPT.} 
    Except for the recent study~\cite{yildiz2024investigating} which conducts CPT over 159 domains, the longest sequence of pre-training stages explored is 8~\cite{jin2022lifelong,gururangan2022demix}. However, this falls short of real-world scenarios where continual pre-training occurs more frequently and persists for months or years. The efficacy of CPT methods in such prolonged scenarios remains uncertain. Additionally, investigating CPT in a task-boundary-free data stream setting is an important avenue for research to be explored in the future as well.
\end{itemize}

\subsubsection{Distributional Shifts in CPT} 
This survey categorizes distributional shifts of CPT into three main types:
(i)~\emph{Language Shift}: LLMs sequentially learn different language corpora, e.g., English $\rightarrow$ Chinese~\cite{gogoulou2024continual,li2024examining}.
(ii)~\emph{Content Shift}: LLMs sequentially learn corpora from different fields, e.g., chemistry $\rightarrow$ biology~\cite{gururangan2022demix,cossu2022continual,jin2022lifelong,qin2023recyclable,chen2023lifelong,gupta2023continual}.
(iii)~\emph{Temporal Shift}: Distributional shifts occur over time, e.g., news in 2021 $\rightarrow$ news in 2022, with a major focus on timestamp-sensitive knowledge retention and update~\cite{amba2021dynamic,jin2022lifelong,dhingra2022time,jang2022towards,jang2022temporalwiki}.

\textbf{Language Shift.}\quad 
\cite{gogoulou2024continual} focuses on assessing LLMs' natural ability to learn new languages sequentially. With no explicit CL techniques employed, the study observes consistent positive forward transfer of the knowledge, facilitating new language acquisition regardless of the learning order. Forgetting, on the other hand, emerges as a significant challenge that cannot be mitigated by the increasing size of LLMs.
In \cite{li2024examining}, the degree of forgetting of previously learned language when adapting LLMs to a new language is investigated. Various CL techniques, including parameter freezing, LoRA~\cite{hu2021lora}, and (IA)$^3$~\cite{liu2022few}, are evaluated across multiple dimensions. Preliminary experimental results highlight the non-trivial nature of addressing horizontal forgetting for CPT under the language shift as well.

\textbf{Content Shift.}\quad
\cite{yildiz2024investigating} explores the large-scale CPT over 159 content domains, and shows that CPT on various domains can effectively improve models' adaptation ability compared to DAP on single domain.
Similarly, \cite{gupta2023continual} continues the pre-training phase of Pythia~\cite{biderman2023pythia} with no complex CL techniques and discovers that learning rate re-warming consistently improves models trained from scratch. 
Built upon this simple observation, \cite{ibrahim2024simple} further shows that proper combination of learning rate re-warming and re-decay, and replay of the previous data is sufficient to achieve a comparable performance to full re-training. 
LLPT~\cite{jin2022lifelong} establishes a comprehensive training and evaluation protocol for a series of content-level distributional shifts. They assess multiple CL methods and, similar to \cite{gogoulou2024continual}, find consistent forward knowledge transfer, yet horizontal forgetting remains significant. Besides, contrary to the common understanding that experience replay~\cite{chaudhry2019tiny} is the most efficient approach to preventing forgetting, the authors find it ineffective in the case of CPT, due to the potential overfitting issue.
Recyclable Tuning~\cite{qin2023recyclable} shows that if the upstream supplier continually pre-trains LLMs, with or without replay, consumer-side efficiency can be boosted by recycling previously learned update components when proper CL techniques are applied. 

DEMix~\cite{gururangan2022demix} incrementally trains and integrates new experts (DEMix layer) for new domains during CPT. To ensure reasonable inference performance during testing when no domain information is available, it proposes a parameter-free probabilistic approach
to dynamically estimate a weighted mixture of domains. 
DEMix's modularization has been shown to facilitate efficient domain-adaptive pre-training, promote relevant knowledge during inference, and allow for removable components.
Lifelong-MoE~\cite{chen2023lifelong}, similar to DEMix~\cite{gururangan2022demix}, incrementally trains domain experts for new domains. However, Lifelong-MoE differs from DEMix in utilizing a \emph{token-level gating function} to activate multiple experts for intermediate embedding calculation. During training, previous experts' parameters and gating functions remain frozen, and knowledge distillation loss is employed to regulate parameter updates, which thereby makes Lifelong-MoE robust against the issue of horizontal forgetting.

It is noteworthy that some papers draw almost opposite conclusions regarding the significance of CPT for content shifts. For instance, \cite{cossu2022continual} continually pre-trains BERT-based models~\cite{devlin2018bert,liu2019roberta} on five scientific domains and evaluates performance on downstream sentiment analysis. They observe that even the trivial sequential pre-training does not exhibit severe forgetting, prompting reasonable questions about the necessity of CPT. 

\textbf{Temporal Shift.}\quad
In the context of CPT amid content shifts, Multi-Task Learning~(MTL) is often regarded as the upper bound achievable~\cite{pentina2016theoretical, wang2024comprehensive, shi2024unified}. However, this belief does not fully hold when considering CL under temporal shifts~\cite{jang2022towards,jang2022temporalwiki,dhingra2022time}, as temporal shifts can introduce conflicting information, posing challenges for LLMs. For instance, the statement \emph{``Lionel Messi plays for team Barcelona''} remains accurate from 2004 to 2021 but becomes false by 2024, as \emph{``Lionel Messi plays for team Inter Miami''} becomes the correct statement.

Hence, as advocated by CKL~\cite{jang2022towards} and TemporalWiki~\cite{jang2022temporalwiki}, LLMs undergoing continual adaptation to temporal shifts must simultaneously achieve three objectives: (i)~retention of old knowledge, (ii)~acquisition of new knowledge, and (iii)~update of the outdated knowledge.
They evaluate the same set of continual learning baseline methods~\cite{chen2020recall,he2021analyzing,hu2022lora,wang2021kadapter}, 
each highlighting distinct aspects of their impact. CKL~\cite{jang2022towards} 
observes that parameter expansion consistently exhibits robust performance across all experimental conditions. In contrast, replay-based methods struggle to efficiently adapt to new knowledge acquisition and outdated knowledge update, leading to rapid forgetting of newly learned information during training.
TemporalWiki~\cite{jang2022temporalwiki} constructs a series of temporal corpora and their differential sets from sequential snapshots of Wikipedia, revealing that updating LLMs on these differential sets substantially enhances new knowledge acquisition and updates, requiring significantly less computational resources, and various CL techniques prove effective in mitigating horizontal forgetting during this process. 
LLPT~\cite{jin2022lifelong} introduces temporal generalization evaluation for LLMs pre-trained on sequential corpora. Through experiments on a large-scale chronologically-ordered Tweet Stream, the authors demonstrate the superiority of CPT combined with CL techniques to task-specific LMs, in terms of both knowledge acquisition and temporal generalization. Nonetheless, these preliminary experiments do not conclusively determine which specific CL method is more preferable than the others.

Another line of work, Temporal Language Models (TLMs), takes a different approach to address knowledge retention, acquisition, and update under temporal shifts by integrating temporal information into the model~\cite{rosin2022time,dhingra2022time,su2023efficient}. During training, they inject temporal information into training examples as prefixes of prompts, using special tokens~\cite{rosin2022time}, explicit year information~\cite{dhingra2022time}, or syntax-guided structural information~\cite{su2023efficient}. In sequential training experiments conducted by TempoT5~\cite{dhingra2022time}, comparison between continually and jointly pre-trained LMs demonstrates that CPT better balances adaptation and forgetting when the replay rate of past data is appropriately set.

\textbf{Others.}\quad
CPT as a technique to progressively attain novel knowledge, can be used to refine LLMs' behavior.
CEM~\cite{zhao2024large} collects examples where the model's response is incorrect and continually trains the model on these examples, along with a supplemental dataset.
RHO-1~\cite{lin2024rho} proposes Selective Language Modeling (SLM), which employs a reference model to evaluate the perplexity of each token in the training corpus, and continually pre-trains the model on high-perplexity tokens. 
Similarly, IR-DRO~\cite{chen2024take} re-trains the model on re-weighted examples from the original pre-training dataset, focusing more on higher-loss sequences.

The significance of addressing temporal shifts through CPT is underscored by several industrial studies. For instance, \cite{amba2021dynamic} employs a dynamic vocabulary expansion algorithm and an efficient sub-sampling procedure to conduct CPT on large-scale emerging tweet data. Conversely, \cite{loureiro2022timelms} adopts CPT without explicit measures to constrain model updates, releasing a series of BERT-based LMs incrementally trained on new tweet data every three months. Preliminary experimental results demonstrate substantial improvements of continually pre-trained LMs over the base BERT model across downstream tasks. While some studies question the necessity of continually adapting LLMs along the temporal axis for environmental reasons, such as reducing CO$_2$ emissions~\cite{attanasio2023worth}, the community commonly embraces CPT as a more efficient learning paradigm compared to the traditional ``combine-and-retrain'' approach.

\subsection{Domain-Adaptive Pre-training~(DAP)}
\label{sec:dap}
\textbf{Background of DAP.}\quad
Institutions, regardless of size, often possess significant amounts of unlabeled, domain-specific data. This data bridges the gap between general-purpose LLMs trained on diverse corpora and fine-tuned LLMs designed for specific downstream tasks. Leveraging this data as a preparatory stage can facilitate effective adaptation of LLMs to downstream tasks.
Such process of ``continued/continual/continuous pre-training''~\cite{yan2023af,guo2023continuous,ma2023ecomgptct,han2021econet,xie2023efficient,xie2023quert,huang2023lawyer,Lu2023BBTFin,Xie2023PIXIU,Azerbayev2023LLEMMA,yue2023mammoth,colombo2024saullm7b,Zhang2024SciGLM,shen2024tag}, 
``further pre-training''~\cite{song2024code,lin2023geogalactica,deng2023learning,Rubungo2023LLM-Prop,agarwal2024structured}, 
``domain tuning''~\cite{rongali2021continual},
``knowledge enhancement pre-training''~\cite{Lu2023BBTFin}, 
and ``knowledge injection training''~\cite{wu2023pmc} 
is unified and termed ``\textbf{\emph{Domain Adaptive Pre-training~(DAP)}}''~\cite{gururangan2020dont} for clarity and consistency throughout this survey.
In the pioneering work of domain-adaptive pre-training~(DAPT)~\cite{gururangan2020dont}, the authors continuously pre-train the language models on a larger domain-specific dataset before fine-tuning them to the downstream tasks, resulting in universally improved performance aross various tasks. 
As the observation above has been validated on multiple domains in parallel, including BioMed, CS, News, and Reviews~\cite{gururangan2020dont}, practitioners commonly accept that employing DAP on additional unlabeled domain-specific data benefits downstream tasks. Consequently, this technique has become widely deployed in many modern LLMs. 

\textbf{Summary of LLMs with DAP.}\quad
We provide a summary of the existing 41 studies utilizing DAP for LLMs in \Tabref{tab:dap}. Each entry is characterized by three main features: (i)~training process specifications, encompassing the vertical domain for which LLMs are trained, the training pipeline preceding release, and the LLM architecture employed; (ii)~adopted continual learning techniques, including rehearsal, parameter regularization, and architecture expansion; and (iii) evaluation metrics for CL, such as backward transfer (forgetting) and forward transfer (adaptation to downstream data). 

\subsubsection{General Observation on DAP}\label{sec:dap-obs}
Several key observations emerge regarding the research landscape of DAP~(\Tabref{tab:dap}). 
\begin{itemize}
    \item \textbf{OBS-1: DAP predominantly occurs in a single stage.}
    Continual DAP which involves more than one stage is seldom explored: among all papers listed in \Tabref{tab:dap}, only one employs two stages of DAP~(``PT $\rightarrow$ DAP $\rightarrow$ DAP $\rightarrow$ FT'' in Code Llama \cite{rozière2024code}). 
    It is arguably reasonable to categorize studies that conduct only one stage of DAP and nothing more~\cite{Lu2023BBTFin,Nguyen2023AstroLLaMA,song2024code,xie2023efficient,li2023starcoder,guo2024deepseekcoder,Xue2023WeaverBird,paul2024ircoder,Azerbayev2023LLEMMA,cheng2024adapting} into CPT rather than DAP. Nevertheless, considering that they aim to adapt a general-purpose LLM to a specific domain, we include them in this section.
    \item \textbf{OBS-2: The notion of interpreting DAP through the lens of CL, whether intentional or not, is widely embraced.}
    As shown in~\Tabref{tab:dap}, except for the first section~(white, 13/41), where papers overlook any potential side effects of DAP leading to vertical forgetting, the remaining sections~(all gray, 28/41) either evaluate the potential negative impacts of DAP or proactively employ CL techniques to mitigate the risk of vertical forgetting.
    \item \textbf{OBS-3: Further research of more sophisticated CL techniques for not just DAP, but general vertical continual learning is much needed.} It is supported by the widespread adoption of CL techniques (22/41) for training domain-specific LLMs. However, the diversity of these techniques is limited, with only replay~\cite{colombo2024saullm7b,wu2023pmc,Azerbayev2023LLEMMA,rongali2021continual,Chen2023HuatuoGPTII,Zhang2023xuanyuan,Yang2023PLLaMa,ma2023ecomgptct,huang2023lawyer,cheng2024adapting} and parameter expansion (LoRA~\cite{Xue2023WeaverBird,paul2024ircoder,wu2024llama,yan2023af}) or Layer/Block expansion~\cite{wu2024llama,yan2023af} utilized. 
    In fact, it appears that individuals may not explicitly recognize that DAP should be viewed from the perspective of vertical continuity, as they often employ CL techniques unknowingly, e.g., studies deploying replay terming the technique as ``data combination''~\cite{wu2023pmc} or ``data mixing/mixture''~\cite{Azerbayev2023LLEMMA,Yang2023PLLaMa,ma2023ecomgptct,cheng2024adapting}, without recognizing it as a typical CL solution to vertical continual learning. 
\end{itemize}

\subsubsection{Different Domains of DAP}
\label{sec:dap-domains}
We include work aimed at establishing vertical LLMs across various domains, including legal, medical, financial, scientific, and code. Additionally, we cover other domains such as language and e-commerce.

\textbf{Legal Domain.}\quad 
In Layer Llama~\cite{huang2023lawyer}, the authors gathered publicly available legal texts from China Courts websites, 
totaling approximately 10 billion tokens as noted in a GitHub issue. 
In SaulLM~\cite{colombo2024saullm7b}, the authors collected the DAP corpus from various jurisdictions in different countries, 
resulting in a corpus of 30 billion tokens to cover diverse aspects of legal texts. When combined with previously available datasets, the total number of tokens used for legal-domain DAP reaches 94 billion. 
The substantial volume of DAP data, while offering valuable insights into specific domains, increases the risk of vertical forgetting of the general knowledge due to the large number of update steps involved. To mitigate this issue, SaulLM incorporates general data from Wikipedia, StackExchange, and GitHub into the DAP data, constituting about 2\% of the final dataset~\cite{colombo2024saullm7b}. 
Similarly, Lawyer Llama incorporates replaying general-domain data during DAP, but the replay rate is not disclosed~\cite{huang2023lawyer}. 
\cite{takahashi2024pretraining} also replays of non-latest business documents during DAP when building a Japanese business-specific LLM. 


\textbf{Medical Domain.}\quad 
Efforts have been made to develop medical specialists by either training an LLM from scratch~\cite{gu2021domain,luo2022biogpt} or fine-tuning publicly-available LLMs to meet specific medical needs~\cite{luo2023biomedgpt,wu2023pmc,Chen2023HuatuoGPTII}. Among these approaches, DAP techniques have been extensively utilized to preserve the communication and instruction-following abilities of a general LLM, preparing it for subsequent medical applications~\cite{luo2023biomedgpt,wu2023pmc,Chen2023HuatuoGPTII}.
BioMedGPT~\cite{luo2023biomedgpt} is a multi-modal biomedical language model that integrates representations of human language and the language of life (molecules, proteins, cells, genes, etc.). Prior to final multi-modal supervised fine-tuning, the authors initialize the model from Llama2-Chat~\cite{touvron2023llama2} and conduct DAP using extensive biomedical documents from S2ORC~\cite{lo2020s2orc}, without considering any CL techniques or evaluations.
In \cite{guo2023continuous}, DAP is performed using Chinese medical encyclopedias and online expert articles, 
with next-token prediction as the training objective. During DAP, the performance gradually deteriorates on general-domain datasets as the training step increases, but improves on the downstream medical examination tasks~\cite{hendryckstest2021}.
PMC-LLama~\cite{wu2023pmc} gathers biomedical papers from S2ORC~\cite{lo2020s2orc} and medical textbooks for ``knowledge injection training.'' During this phase, a general language corpus from RedPajama-Data~\cite{together2023redpajama} is replayed at a 5\% rate within a training batch. However, the paper does not analyze the effectiveness of this operation of mixing in general-domain data for DAP.

To mitigate vertical forgetting, AF Adapter~\cite{yan2023af} proposes an adapter structure extending the width of Attention layers and FFNs for acquiring domain knowledge and only the adapters are tuned during DAP. 
Similarly, Hippocrates~\cite{acikgoz2024hippocrates} deploys LoRA during DAP to both have medical-specific knowledge injected and general ability preserved. 
Me-Llama~\cite{xie2024me} mixes in about 25\% of the general-domain data for DAP on the clinical notes and biomedical articles, which achieves even positive backward transfer on MMLU~\cite{hendryckstest2021}. 
HuatuoGPT-II~\cite{Chen2023HuatuoGPTII} proposes to fuse the DAP into the final SFT, unifying the two stages into one single process. The challenge of such process mainly comes from the data heterogeneity of DAP's unlabeled corpus. The authors address this challenge by reformulating paragraphs of data into \emph{(instruction, output)} format using existing large language models. They further employ a priority sampling strategy to avoid compromising downstream ability, a pitfall observed in the fixed-rate data mixing strategy~\cite{touvron2023llama2}. This paper empirically demonstrates the superiority of unified one-stage SFT over two-stage training, questioning the reasonability of the current DAP.
On medical-domain data, \cite{rongali2021continual}
finds that LMs constrained by CL techniques on source domains exhibit greater robustness to future domain shifts. Specifically, they identify that parameter regularization techniques like EWC~\cite{kirkpatrick2017overcoming}, despite slightly higher cost, can facilitate positive forward and backward transfer.

\begin{table*}[htbp]
    \centering
    \caption{
    \textbf{Summary of the existing studies that leverage Domain-Adaptive Pre-training of LLMs,} 
    where the papers are organized in four main categories based on whether they (i) adopt the \emph{continual learning techniques} and (ii) perform the evaluation for \emph{backward transfer~(forgetting)}.
    In the column of \textbf{Train Proc.}~(Training Process), we omit the phase of general Pre-Training. DAP represents Domain-Adaptive Pre-Training; SFT represents Supervised Fine-Tuning; IT represents Instruction Tuning. The prefix G- and D- represent General and Domain-Specific training process~\cite{lin2023geogalactica,huang2023lawyer}, and the prefix U- represents them unified~\cite{wu2024llama,Chen2023HuatuoGPTII}. The prefix MM- and LC- represents Multi-Modal and Long-Context training phases~\cite{luo2023biomedgpt,Zheng2023MarineGPT,rozière2024code}.
    In the column of \textbf{Continual Learning Eval}., we consider two criteria: (i)~\emph{Backward Transfer}, i.e., performance degradation on the previous tasks, which is also known as catastrophic forgetting, (ii)~\emph{Forward Transfer}, i.e., the performance gained by DAP while transferring the LLMs to the downstream tasks. We use L and Perp. to denote Loss and Perplexity, FT to denote Fine-Tuning, ZS and FS to denote Zero-Shot and Few-Shot Accuracy, HE and LLM to denote the Human Evaluation and LLM Evaluation for generative tasks.
    }
    \label{tab:dap}
    \resizebox{1\linewidth}{!}{%
    \setlength{\tabcolsep}{5pt}
    \begin{tabular}{ccccC CCcc}
    	\toprule[0.15em]
    	\multirow{2}{*}[-0.25em]{\textbf{Domain}} & 
        \multirow{2}{*}[-0.25em]{\textbf{Method}} & 
        \multirow{2}{*}[-0.25em]{\textbf{Train Proc.}} & 
        \multirow{2}{*}[-0.25em]{\textbf{LLM Arch.}} & 
        \multicolumn{3}{c}{\textbf{{Continual Learning Tech.}}} & 
        \multicolumn{2}{c}{\textbf{{Continual Learning Eval.}}} \\
        \cmidrule(lr){5-7}\cmidrule(lr){8-9}
        & & & & \emph{Rehearsal} & \emph{Param. Reg.} & \emph{Arch. Exp.} & \emph{Backward Transfer} & \emph{Forward Transfer} \\
        \midrule
        \midrule

        Medical & BioMedGPT~\cite{luo2023biomedgpt} 
        & \small{DAP $\rightarrow$ MM-SFT} & Llama2 
        & \xmark & \xmark & \xmark 
        & \xmark & FT \\
        \hline

        Financial & BBT-Fin~\cite{Lu2023BBTFin} 
        & \small{DAP} & T5 
        & \xmark & \xmark & \xmark 
        & \xmark & FT \\
        \hline

        Financial & CFGPT~\cite{li2023cfgpt} 
        & \small{DAP $\rightarrow$ SFT} & InternLM 
        & \xmark & \xmark & Q-LoRA$_{\text{(SFT)}}$ 
        & \xmark & HE$^1$ \\
        \hline

        Scientific & AstroLlama~\cite{Nguyen2023AstroLLaMA} 
        & \small{DAP} & LlaVa 
        & \xmark & \xmark & \xmark 
        & \xmark & Perp. \\
        \hline

        Scientific & OceanGPT~\cite{Bi2023OCEANGPT} 
        & \small{DAP $\rightarrow$ IT} & \makecell{Vicuna \\ Llama2-chat \\ ChatGLM2} 
        & \xmark & \xmark & LoRA$_{\text{(IT)}}$ 
        & \xmark & HE\\
        \hline

        Scientific & K2~\cite{deng2023learning} 
        & \small{DAP $\rightarrow$ SFT} & Llama 
        & \xmark & \xmark & LoRA$_{\text{(SFT)}}$ 
        & \xmark & Perp. | ZS | LLM \\
        \hline

        Scientific & MarineGPT~\cite{Zheng2023MarineGPT} 
        & \small{MM-DAP $\rightarrow$ MM-IT} & Llama 
        & \xmark & \xmark & \xmark 
        & \xmark & HE \\
        \hline

        Code & CodeGen~\cite{nijkamp2022codegen} 
        & \small{DAP $\rightarrow$ DAP} & CodeGen 
        & \xmark & \xmark & \xmark 
        & \xmark & \makecell{Perp. | ZS}\\
        \hline

        Code & Comment-Aug~\cite{song2024code} 
        & \small{IT $\rightarrow$ DAP} & \makecell{Llama2 \\ Code~Llama \\ InternLM2}
        & \xmark & \xmark & \xmark 
        & \xmark & ZS \\
        \hline

        EventTemporal & EcoNet~\cite{han2021econet}$^1$ 
        & \small{DAP $\rightarrow$ FT} & \makecell{BERT \\ RoBERTa}
        & \xmark & \xmark & \xmark 
        & \xmark & FT \\
        \hline

        CommonSense & CALM~\cite{zhou2020pre} 
        & \small{DAP $\rightarrow$ FT} & T5 
        & \xmark & \xmark & \xmark 
        & \xmark & FT \\
        \hline
        
        Multi-Domain & BLADE~\cite{li2024blade} 
        & \small{DAP $\rightarrow$ IT} & BLOOMZ 
        & \xmark & \xmark & \xmark
        & \xmark & ZS 
        \\
        \hline

        Scientific & ClimateGPT~\cite{thulke2024climategpt} 
        & \small{DAP $\rightarrow$ IT $\rightarrow$ RAG} & Llama2
        & \xmark & \xmark & \xmark 
        & \xmark & FS | Ret.
        \\
        \midrule
        \midrule

        Medical & \cite{guo2023continuous} 
        & \small{DAP $\rightarrow$ FT} & Llama2 
        & \xmark & \xmark & \xmark 
        & \cellcolor{gray1} FS | FT & FS | FT\\
        \hline

        Financial & \cite{xie2023efficient} 
        & \small{DAP} & Pythia 
        & \xmark & \xmark & \xmark 
        & \cellcolor{gray1} L | FS & L | FS \\
        \hline

        Scientific & GeoGalactica~\cite{lin2023geogalactica} 
        & \small{DAP $\rightarrow$ G-SFT $\rightarrow$ D-SFT} & GAL 
        & \xmark & \xmark & \xmark 
        & \cellcolor{gray1} ZS & Perp. | ZS | LLM \\
        \hline

        Code & StarCoder~\cite{li2023starcoder} 
        & \small{DAP} & StarCoder 
        & \xmark & \xmark & \xmark 
        & \cellcolor{gray1} Perp. | ZS | FS &  Perp. | ZS | FS \\
        \hline

        Code & DeepSeek-Coder~\cite{guo2024deepseekcoder} 
        & \small{DAP} & DeepSeek-LLM 
        & \xmark & \xmark & \xmark 
        & \cellcolor{gray1} ZS | FS & ZS \\
        \hline

        Multi-Domain & DAPT~\cite{gururangan2020dont} 
        & \small{DAP $\rightarrow$ FT} & RoBERTa 
        & \xmark & \xmark & \xmark 
        & \cellcolor{gray1} Loss & L | FT \\
        \hline
        \hline

        Financial & WeaverBird~\cite{Xue2023WeaverBird} 
        & \small{DAP} & GLM2 
        & \cellcolor{gray2} \xmark & \cellcolor{gray2} \xmark & \cellcolor{gray2} LoRA 
        & \xmark & HE \\
        \hline

        Code & IRCoder~\cite{paul2024ircoder} 
        & \small{DAP} & \makecell{StarCoder \\ DeepSeek-Coder \\ Code~Llama}
        & \cellcolor{gray2} \xmark & \cellcolor{gray2} \xmark & \cellcolor{gray2} LoRA 
        & \xmark & ZS \\
        \hline

        Code & Code~Llama~\cite{rozière2024code} 
        & \makecell{\small{DAP $\rightarrow$ LC-FT $\rightarrow$ IT} \\ \small{DAP $\rightarrow$ DAP $\rightarrow$ LC-FT}} & Llama2
        & \cellcolor{gray2} Replay & \cellcolor{gray2} \xmark & \cellcolor{gray2} \xmark 
        & \xmark & Perp. | ZS \\
        \hline

        Legal & SaulLM~\cite{colombo2024saullm7b} 
        & \small{DAP $\rightarrow$ U-IT} & Mistral 
        & \cellcolor{gray2} Replay & \cellcolor{gray2} \xmark & \cellcolor{gray2} \xmark 
        & \xmark & Perp. | ZS \\
        \hline

        Medical & PMC-Llama~\cite{wu2023pmc} 
        & \small{DAP $\rightarrow$ IT} & Llama 
        & \cellcolor{gray2} Replay & \cellcolor{gray2} \xmark & \cellcolor{gray2} \xmark 
        & \xmark & ZS | FT \\
        \hline

        Scientific & Llema~\cite{Azerbayev2023LLEMMA} 
        & \small{DAP} & Code~Llama 
        & \cellcolor{gray2} Replay & \cellcolor{gray2} \xmark & \cellcolor{gray2} \xmark 
        & \xmark & Perp. | FS \\
        \hline

        Multi-Domain & DAS~\cite{ke2022continual-pre} 
        & \small{[DAP]$_n$} & RoBERTa 
        & \cellcolor{gray2} DER++$^\text{\club}$ & \cellcolor{gray2} EWC$^\text{\club}$\qquad HAT$^\text{\club}$\qquad Soft-Masking & \cellcolor{gray2} Adapter$^\text{\club}$\qquad DEMix$^\text{\club}$ 
        & \xmark & FT \\
        \hline

        Medical & Hippocrates~\cite{acikgoz2024hippocrates} 
        & \small{DAP $\rightarrow$ IT $\rightarrow$ MA} & \makecell{Llama2 \\ Mistral}
        & \cellcolor{gray2} \xmark & \cellcolor{gray2} \xmark & \cellcolor{gray2} LoRA
        & \xmark & ZS | FS
        \\
        \hline

        Language & Sailor~\cite{dou2024sailor} 
        & \small{DAP} & Qwen1.5 
        & \cellcolor{gray2} Replay & \cellcolor{gray2} \xmark & \cellcolor{gray2} \xmark
        & \xmark & ZS  
        \\

        \midrule
        \midrule

        Code \& Math & Llama Pro~\cite{wu2024llama} 
        & \small{DAP $\rightarrow$ U-SFT} & Llama2 
        & \cellcolor{gray3} \xmark & \cellcolor{gray3} \xmark & \cellcolor{gray3} Block~Exp. LoRA$^{\text{\club}}$ 
        & \cellcolor{gray3} ZS | FS & Perp. | ZS | FS \\
        \hline

        Medical & AF Adapter~\cite{yan2023af} 
        & \small{DAP $\rightarrow$ FT} & RoBERTa 
        & \cellcolor{gray3} \xmark & \cellcolor{gray3} \xmark & \cellcolor{gray3} Layer~Exp. LoRA$^{\text{\club}}$ 
        & \cellcolor{gray3} Acc. & L | FT\\
        \hline

        Medical & \cite{rongali2021continual} 
        & \small{DAP $\rightarrow$ FT} & \makecell{BERT \\ RoBERTa \\ DistilBERT}
        & \cellcolor{gray3} Replay$^{\text{\club}}$ GEM$^{\text{\club}}$ & \cellcolor{gray3} L2~Reg.$^{\text{\club}}$ EWC$^{\text{\club}}$ & \cellcolor{gray3} \xmark 
        & \cellcolor{gray3} L | FT & L | FT\\
        \hline

        Medical & HuatuoGPT-II~\cite{Chen2023HuatuoGPTII} 
        & \small{DAP + U-SFT} & Baichuan2 
        & \cellcolor{gray3} Replay & \cellcolor{gray3} \xmark & \cellcolor{gray3} \xmark 
        & \cellcolor{gray3} ZS & ZS | HE\\
        \hline

        Financial & XuanYuan 2.0~\cite{Zhang2023xuanyuan} 
        & \small{DAP + SFT} & BLOOM 
        & \cellcolor{gray3} Replay & \cellcolor{gray3} \xmark & \cellcolor{gray3} \xmark 
        & \cellcolor{gray3} HE & HE\\
        \hline

        Scientific & PLlama~\cite{Yang2023PLLaMa} 
        & \small{DAP $\rightarrow$ IT} & GAL 
        & \cellcolor{gray3} Replay & \cellcolor{gray3} \xmark & \cellcolor{gray3} \xmark 
        & \cellcolor{gray3} L &  L | ZS \\
        \hline

        E-Commerce & EcomGPT-CT~\cite{ma2023ecomgptct} 
        & \small{DAP $\rightarrow$ SFT} & BLOOM 
        & \cellcolor{gray3} Replay & \cellcolor{gray3} \xmark & \cellcolor{gray3} \xmark 
        & \cellcolor{gray3} ZS | FS & ZS | FS \\
        \hline

        Legal & Layer~Llama~\cite{huang2023lawyer} 
        & \small{DAP $\rightarrow$ G-IT $\rightarrow$ D-IT} & Llama 
        & \cellcolor{gray3} Replay & \cellcolor{gray3} \xmark & \cellcolor{gray3} \xmark 
        & \cellcolor{gray3} ZS & ZS\\
        \hline

        Multi-Domain & AdaptLLM~\cite{cheng2024adapting} 
        & \small{DAP} & Llama 
        & \cellcolor{gray3} Replay & \cellcolor{gray3} \xmark & \cellcolor{gray3} \xmark 
        & \cellcolor{gray3} ZS & ZS | FT\\
        \hline

        Language & Swallow~\cite{fujii2024continual} 
        & \small{DAP} & Llama2
        & \cellcolor{gray3} Replay & \cellcolor{gray3} \xmark  & \cellcolor{gray3} \xmark 
        & \cellcolor{gray3} FS & FS 
        \\
        \hline

        Financial & \cite{takahashi2024pretraining} 
        & \small{DAP} & Llama2 
        & \cellcolor{gray3} Replay & \cellcolor{gray3} \xmark & \cellcolor{gray3} \xmark
        & \cellcolor{gray3} Loss | ZS & Loss | ZS | FS | RAG
        \\
        \hline

        Medical & Me-Llama~\cite{xie2024me} 
        & \small{DAP $\rightarrow$ IT} & Llama2
        & \cellcolor{gray3} Replay & \cellcolor{gray3} \xmark & \cellcolor{gray3} \xmark
        & \cellcolor{gray3} ZS | FS & ZS | FS | FT
        \\
        \hline

        Language & Aurora-M~\cite{nakamura2024aurora} 
        & \small{DAP $\rightarrow$ IT} & StarCoder 
        & \cellcolor{gray3} Replay & \cellcolor{gray3} \xmark  & \cellcolor{gray3} \xmark
        & \cellcolor{gray3} ZS & ZS | FS | HE
        \\
        \hline
        
    	\bottomrule[0.15em]
	   \end{tabular}
	}
\end{table*}

\textbf{Financial Domain.}\quad
A gap persists between general-purpose LLMs and existing domain-specific smaller-scale LLMs~\cite{araci2019finbert,Wu2023BloombergGPT}, underscoring the urgent need for more powerful financial-domain experts through the integration of LLMs. Notably, DAP techniques have emerged as crucial tools for tailoring LLMs to the intricacies of the financial domain while mitigating the negative effects of abrupt domain shifts from general to finance~\cite{Lu2023BBTFin,li2023cfgpt,xie2023efficient,Xue2023WeaverBird,Zhang2023xuanyuan}.

BBT-Fin~\cite{Lu2023BBTFin} collects a Chinese financial DAP dataset comprising 80 billion tokens sourced from corporate reports, analyst reports, social media, and financial news. In addition to the conventional masked language modeling (MLM) training objective, BBT-Fin further incorporates triplet masking and span masking techniques during DAP. 
CFGPT~\cite{li2023cfgpt} creates CFData, a financial dataset for DAP and SFT, comprising 141 billion tokens. 
During DAP, CFGPT does not employ CL techniques but utilizes QLoRA~\cite{dettmers2023qlora} for preventing overfitting to downstream data and balancing general response ability and domain-specific ability during SFT. These two methods are typical domain-specific LLMs focusing solely on adaptation to target domains without explicit CL measures or evaluation of vertical forgetting.

In \cite{xie2023efficient}, the authors aim to enhance the data efficiency of DAP. 
When the downstream tasks' data distribution $\gT$ are known, based on the generalization bound~\cite{ben2010theory,ganin2016domain,shi2024unified}, the authors propose to sample the subset of DAP data whose distribution $\gD$ is similar to the downstream task's data, i.e., $d_{\gH\Delta \gH}(\gD, \gT)$ is low. 
When the downstream data distribution is unknown, the authors suggest ensuring \emph{novelty} and \emph{diversity} in the sampled corpus for DAP. 
This approach significantly enhances DAP efficiency: it utilizes only 10\% of the originally collected data yet outperforms models trained on the entire DAP dataset, underscoring the importance of data quality over quantity. 
WeaverBird~\cite{Xue2023WeaverBird} introduces an intelligent finance dialogue system, where the encoder is trained on Chinese and English financial documents, alongside expert-annotated financial query-response pairs, using LoRA~\cite{hu2022lora}. Xuanyuan 2.0~\cite{Zhang2023xuanyuan}, akin to HuatuoGPT-II~\cite{Chen2023HuatuoGPTII}, proposes the technique of hybrid-tuning, which fuses the stages of DAP and SFT into one, general-domain data and financial-domain data into one. Notably, the distribution of data in hybrid-tuning is unconventional: financial DAP data comprises only a small portion of 13\%. This prompts a pertinent question in line with the investigation on efficient DAP in \cite{xie2023efficient}: Is a large DAP dataset necessary for developing a domain-specific LLM?


\textbf{Scientific Domain.}\quad
Vertical scientific LLMs span many subjects~\cite{Zhang2024SciGLM,Nguyen2023AstroLLaMA,Azerbayev2023LLEMMA,luo2023wizardmath,lin2023geogalactica,Bi2023OCEANGPT,Zheng2023MarineGPT}.
However, among all the studies listed above, only a small fraction of them adopt the technique of DAP. 
OceanGPT~\cite{Bi2023OCEANGPT} is the first LLM tailored specifically for the ocean domain. It performs DAP on a raw corpus of ocean science literature, prioritizing recent research and historically significant works.
K2~\cite{deng2023learning} pioneers the development of a foundational language model tailored specifically for geoscience. It aggregates geoscience open access literature and Earth science-related Wikipedia pages for DAP. Following this, it undergoes multi-task instruction tuning utilizing LoRA~\cite{hu2022lora} on both a general instruction tuning dataset and the GeoSignal benchmark introduced within the K2 framework.
AstroLlama~\cite{Nguyen2023AstroLLaMA} gathers abstracts solely from astronomy papers on arXiv and proceeds pre-training. It observes an improved perplexity on the domain of scholarly astronomy, without providing more quantitative evaluation. 
MarineGPT~\cite{Zheng2023MarineGPT} is a multi-modal LLM designed specifically for the marine domain. 
During DAP, MarineGPT incorporates 5 million marine image-text pairs to imbue domain knowledge. This involves training a Q-Former~\cite{li2023blip2} between the frozen visual and text decoder~\cite{dosovitskiy2020image,touvron2023llama}. 

Another branch of methods proactively integrate in the replay of the general-domain data to mitigate vertical forgetting.
GeoGalactica~\cite{lin2023geogalactica} introduces a series of LLMs tailored for geoscience. In the DAP phase, besides the 52-billion-token geoscience corpus, Arxiv papers and Codedata are incorporated, with a mixing ratio of 8:1:1. The authors believe that the inclusion of the Codedata during the model's pre-training can significantly boost the reasoning ability of the LLMs. 
Although GeoGalactica pinpoints challenges of DAP, including overfitting, catastrophic forgetting,  maintaining the training stability, and convergence speed, it does not further provide empirical evidence supporting the inclusion of the Codedata, or deploying specific measures to address the challenges proposed above. 
Llemma~\cite{Azerbayev2023LLEMMA} focuses on mathematics, initialized from Code~Llama~\cite{rozière2024code}, and undergoes DAP on a blend of the 55-billion-token mathematical pre-training dataset and general domain data at the ratio of 19:1. 
In contrast, PLlama~\cite{Yang2023PLLaMa}, designed for plant science, mixes domain-specific and general-domain data at the ratio of 9:1.

\textbf{Code Domain.}\quad
The development of LLMs for automatic code filling, debugging, and generation holds significant practical importance~\cite{moradidakhel2023github,sun2024survey}. These advancements cover various frameworks, including encoder-only~\cite{moradidakhel2023github}, encoder-decoder~\cite{wang2021codet5,wang2023codet5plus}, and decoder-only~\cite{nijkamp2022codegen,lozhkov2024starcoder,guo2024deepseekcoder}.
There is a growing trend towards decoder-only architectures~\cite{sun2024survey}, leveraging models pre-trained on general natural language like Llama~\cite{touvron2023llama,touvron2023llama2}. Consequently, there is a shift in the training objective from utilizing code structures to simpler tasks like next token prediction and infilling.

From the perspective of CL, the code domain presents unique advantages and challenges for DAP, compared to other domains. On one hand, its hierarchical structure (\emph{general domain corpus $\rightarrow$ multi-language code $\rightarrow$ specific programming language}) provides an ideal training pipeline for DAPs~\cite{rozière2024code}, offering potential for more efficient training strategies. On the other hand, programming languages adhere to strict grammars, unlike the fuzzy and context-dependent natural language. Consequently, language models should ideally leverage these structures through tailored designs, and adopting the same training objectives as for natural languages may yield sub-optimal results. Therefore, many existing studies omit DAP~\cite{wang2021codet5,wang2023codet5plus,luo2023wizardcoder}.
In the following section, we will introduce existing code LLMs that employ DAP before the final downstream tasks, discussing both their common attributes and unique characteristics.

Representing a series of notable works that focus solely on adaptation to target domains, 
CodeGen~\cite{nijkamp2022codegen} comprises a suite of LLMs designed for natural language~(CodeGen-NL), multi-lingual programming languages~(CodeGen-Multi), and mono-lingual programming languages~(CodeGen-Mono). These models are trained sequentially, with each subsequent model initialized from the previous one trained on more general-domain data. 
Comment-Aug~\cite{song2024code} addresses the challenge of aligning programming languages with natural languages (PL-NL alignment) by performing DAP on the code augmented with generated additional comments. 
StarCoder~\cite{li2023starcoder} introduces two models: StarCoderBase and StarCoder. StarCoderBase is initially trained on a mixed dataset comprising various programming languages without significant reweighting on the data. Subsequently, StarCoderBase undergoes further fine-tuning on additional 35 billion tokens of Python code, resulting in the development of StarCoder.
DeepSeek-Coder-v1.5~\cite{guo2024deepseekcoder} originates from DeepSeek-LLM~\cite{deepseekai2024deepseek} and undergoes pre-training on 2 trillion tokens, comprising 87\% source code, 10\% English code-related natural language, and 3\% Chinese natural language corpus. 
Initialization from a general-domain LLM results in improved performance across various tasks, including natural language and mathematical reasoning, with minimal performance degradation on coding tasks, which underscores the efficacy of DAP.

As the only work that utilizes the general data replay to mitigate vertical forgetting in the code domain,
Code~Llama~\cite{rozière2024code} introduces a sophisticated training framework tailored for various coding tasks and model sizes. Initialized from Llama~2 weights, these models undergo DAP on a dataset composed of deduplicated public code, discussions about code, and a subset of natural language data. This mix of natural language data serves as a form of pseudo-replay to maintain the models' proficiency in understanding natural language. 
Besides replay, architecture expansion has proven effective in acquiring robust coding abilities and preventing vertical forgetting simultaneously.
IRCoder~\cite{paul2024ircoder} utilizes compiler intermediate representations to enhance the multilingual transferability of Code LLMs. By conducting DAP on code grounded in intermediate representations with LoRA~\cite{hu2021lora}, IRCoder achieves superior multilingual programming instruction following, enhanced multilingual code understanding, and increased robustness to prompt perturbations.
Llama~Pro~\cite{wu2024llama} undergoes DAP on a combination of code and math data. It expands the original Llama2 architecture by dynamically adding multiple identity copies of the transformer blocks. These added blocks initially preserves the original functionality, and will be tuned for DAP. The proposed expansion method is shown to be more resilient against vertical forgetting compared to other parameter-efficient tuning methods like LoRA.

The three aforementioned studies highlight the importance of DAP for code LLMs. However, it is crucial to note that the problem definition and conventional architectures of existing Code LLMs may present challenges of compatibility for DAP deployment, and need to be addressed in the future.

\textbf{Other Domains.}\quad
ECONET~\cite{han2021econet} enhances the model's ability to reason about event temporal relations through a dedicated DAP phase. Temporal and event indicators are masked out, and a contrastive loss is applied to the recovered masked tokens. Results demonstrate that incorporating this DAP stage significantly improves performance on final tasks compared to direct fine-tuning.
Concept-Aware Language Model~(CALM)~\cite{zhou2020pre} introduces a data-efficient DAP approach for enhancing the concept-centric commonsense reasoning ability of LLMs. It incorporates both generative and discriminative commonsense reasoning tasks specifically tailored for concept-centric reasoning tasks. Consequently, even a small number of data examples for DAP can lead to notable improvements for downstream tasks.

Aurora-M~\cite{nakamura2024aurora} and Swallow~\cite{fujii2024continual} adopt the simple replay strategy that mixes in a small portion of general data during DAP for their multi-lingual ability.
Furthermore, Sailor~\cite{dou2024sailor} studies the optimal strategy of data mixing for DAP, balancing the general knowledge and capacity of different languages. 
EcomGPT-CT~\cite{ma2023ecomgptct} employs a data mixing strategy for DAP which transforms semi-structured E-commerce data into a set of nodes and edges, samples a cluster of nodes, and then extracts and concatenates them into a training example. It combines the general-domain corpus with E-commerce data at a ratio of 2:1, which is significantly lower than the common setting adopted by other works.

Notably, there are some papers studying other effective ways of DAP. 
AdaptLLM~\cite{cheng2024adapting} transforms raw corpora into \emph{(raw text, question, answer)} format, creating intrinsic reading comprehension tasks. 
AdaptLLM demonstrates superior domain-specific knowledge adaptation and minimal vertical forgetting, thereby challenging the data efficiency of conventional DAP. 
Tag-LLM~\cite{shen2024tag} re-purposes the general-domain LLM into domain-specific one by multi-stage training of domain tags and function tags, without modifying the base LLM's weights and thereby mitigates forgetting.

\subsection{Continual Fine-Tuning~(CFT)}
\label{sec:cft}

\textbf{Background of Continual Fine-Tuning~(CFT).}\quad
Continual Fine-Tuning~(CFT) lies at the bottom layer of the vertical continuity, where models are trained on successive homogeneous tasks drawn from an evolving data distribution. As the service-oriented layer of LLM, it does not require consideration of further adaptation to another downstream tasks, simplifying optimization objectives to a great extent: better adaptation and less forgetting\footnote{We direct interested readers to additional survey literature on the topic of general CFT~\cite{biesialska2020continual,ke2023continual}.}. 
In the era of LLMs, new computational paradigms in CFT have emerged and attracted significant attention within the research community. 
These topics include (i)~Continual Instruction Tuning~(CIT)~\cite{zhang2023citb}, (ii)~Continual Model Refinement~(CMR)~\cite{hartvigsen2023aging}, (iii)~Continual Model Alignment~(CMA)~\cite{lin2024mitigating,zhangcppo}, and (iv)~Continual Learning for Multimodal Language Models~(CMLLMs)~\cite{he2023continual,ni2023continual}.
We summarize existing studies on CFT in \Tabref{tab:cft}, categorizing studies into sub-categories as listed above. The table includes details on incremental learning types (X-IL), LLM architecture, and employed CL techniques and evaluation metrics. After discussing general observations on CFT in \Secref{sec:cft-obs}, we will delve into each sub-category in detail.

\setlength{\aboverulesep}{0pt}
\setlength{\belowrulesep}{0pt}
\begin{table*}[htbp]
    \centering
    \caption{
    \textbf{Summary of the existing studies on Continual Fine-Tuning LLMs,} where the papers are organized in five main categories based on what downstream tasks they are designed to tackle, including (i)~General Continual Fine-Tuning~(CFT); (ii)~Continual Instruction Tuning~(CIT); (iii)~Continual Model Refinement~(CMR); (iv)~Continual Model Alignment~(CMA); (v)~Continual Multimodal LLMs~(CMLLMs), which is shown in the column of \textbf{CFT Type}.
    The column of \textbf{X-IL} shows what continual learning paradigm the study includes~\cite{van2022three}, where 
    \emph{TIL} represents task-incremental learning, meaning task ID/information is provided during inference; 
    \emph{DIL} represents domain-incremental learning, meaning the tasks are defined in the same format, and no task ID/information is available during inference; 
    \emph{CIL} represents class-incremental learning, meaning the task ID needs to be further inferred when testing. 
    }
    \label{tab:cft}
    \resizebox{1\linewidth}{!}{%
    \setlength{\tabcolsep}{2pt}
\begin{tabular}{ccccc ccccc c}
	\toprule[0.15em]
    \multirow{2}{*}[-0.6em]{\textbf{CFT Type}} & 
    \multirow{2}{*}[-0.6em]{\textbf{Method}} & 
    \multirow{2}{*}[-0.6em]{\textbf{X-IL}} & 
    \multirow{2}{*}[-0.6em]{\textbf{LLM Arch.}} & 
    \multicolumn{4}{c}{\textbf{{Continual Learning Tech.}}} & 
    \multicolumn{3}{c}{\textbf{{Continual Learning Eval.}}} \\
    \cmidrule(lr){5-8}\cmidrule(lr){9-11}
    & & & & \emph{Rehearsal} & \emph{Param. Reg.} & \emph{Arch. Exp.} & \emph{Others} & \emph{\makecell{Avg. \\Acc.}} & \emph{\makecell{Bwd. \\Trans.}} & \emph{\makecell{Fwd. \\Trans.}} \\
    \midrule
    \midrule
    
    \multirow{11}{*}[0em]{\makecell{General}} & CTR~\cite{ke2021achieve} & DIL | CIL & BERT 
    & \xmark & \xmark & Adapter & \xmark 
    & \cmark & \cmark & \cmark \\
    \cmidrule{2-11}

    & \cite{tao2022can} & TIL & BERT 
    & S-Replay & \xmark & \xmark & \xmark 
    & \club & \club & \club \\
    \cmidrule{2-11}

    & CIRCLE~\cite{wei2022circle} & DIL & T5
    & Replay & EWC & Prompt & \xmark 
    & \cmark & \cmark & \cmark \\
    \cmidrule{2-11}

    & ConPET~\cite{song2023conpet} & DIL & Llama 
    & Replay & \xmark & LoRA & \xmark 
    & \cmark & \cmark & \cmark \\
    \cmidrule{2-11}

    & \cite{bai2023enhancing} & DIL | CIL & BERT 
    & \xmark & \xmark & \xmark & G-Prompt 
    & \cmark & \cmark & \xmark \\
    \cmidrule{2-11}

    & \cite{luo2023investigating} & TIL & \makecell{DistilBERT \\ ALBERT | RoBERTa} 
    & \makecell{ER | DER | LwF} & \xmark & \xmark & \xmark 
    & \club & \club & \xmark \\
    \cmidrule{2-11}

    & SEQ$^*$~\cite{zheng2023learn} & TIL | CIL & \makecell{Pythia | BERT | GPT2} & \xmark & P-Freeze & \xmark & Tricks for Classifiers 
    & \xmark & \cmark & \xmark \\
    \cmidrule{2-11}

    & LFPT5~\cite{qin2021lfpt5} & DIL & T5 
    & P-Replay & \xmark & \xmark & \xmark 
    & \cmark & \cmark & \xmark\\
    \cmidrule{2-11}

    & \cite{weyssow2023usage} & DIL & \makecell{RoBERTa | GPT2} 
    & Replay & \makecell{EWC | SI | RWalk} & \xmark & \xmark 
    & \cmark & \cmark & \xmark \\
    \cmidrule{2-11}

    & LR~ADJUST~\cite{winata2023overcoming} & DIL & XLM-R 
    & \xmark & \xmark & \xmark & LR Scheduling 
    & \cmark & \cmark & \cmark \\
    \cmidrule{2-11}

    & C3~\cite{chen2024parameterizing} & TIL & T5 
    & KD & \xmark & Prompt Tuning & \xmark 
    &\cmark & \cmark & \xmark \\

    \midrule
    \midrule

    & CT0~\cite{scialom2022fine} & TIL & T0 
    & S-Replay & \xmark & \xmark & \xmark 
    & \cmark & \cmark & \cmark \\
    \cmidrule{2-11}

    & RCL~\cite{wang2023trace} & TIL & \makecell{LLaMA \\ Vicuna | Baichuan} & Replay & \xmark & \xmark & \xmark 
    & \cmark & \cmark & \cmark \\
    \cmidrule{2-11}

    & DynaInst~\cite{mok2023large} & TIL & BART 
    & Replay & \xmark & \xmark & \xmark 
    & \cmark & \cmark & \cmark \\
    \cmidrule{2-11}

    & CITB~\cite{zhang2023citb} & TIL & T5 
    & \makecell{Replay | AGEM} & \makecell{L2 | EWC} & AdapterCL & \xmark 
    & \cmark & \cmark & \cmark \\
    \cmidrule{2-11}
    
    & SSR~\cite{huang2024mitigating} & TIL & \makecell{LLaMA | Alpaca} 
    & \makecell{RandSel | KMeansSel} & \xmark & \xmark & \xmark 
    & \cmark & \cmark & \cmark \\
    \cmidrule{2-11}

    & KPIG~\cite{he2024dont} & DIL | TIL & \makecell{LLaMA | Baichuan} 
    & \makecell{DynaInst | PCLL | DCL} & \makecell{L2 \\ EWC} & \makecell{DARE \\ LM-Cocktail} & KPIG 
    & \cmark & \cmark & \cmark \\
    \cmidrule{2-11}

    & ConTinTin~\cite{yin2022contintin} & TIL & BART 
    & Replay & \xmark & \xmark & InstructionSpeak 
    & \cmark & \cmark & \cmark \\
    \cmidrule{2-11}

    & O-LoRA~\cite{wang2023orthogonal} & TIL & \makecell{LLaMA | Alpaca} & \xmark  & \xmark &  O-LoRA & \xmark 
    & \cmark & \cmark & \cmark \\
    \cmidrule{2-11}

    \multirow{-9}{*}[1em]{{\makecell{CIT}}} & SAPT~\cite{zhao2024sapt} & TIL & \makecell{T5 | LLaMA} 
    & \xmark  & \xmark & \xmark & SAPT 
    & \cmark & \cmark & \cmark \\
    \cmidrule{2-11}

    & InsCL~\cite{wang2024inscl} & TIL & LLaMA 
    & Replay & \xmark & \xmark & InsCL 
    & \cmark & \cmark & \cmark \\
    \midrule
    \midrule

    & CMR~\cite{lin2022continual} & DIL & BART 
     & \makecell{ER | MIR | MLR} & \makecell{L2 | EWC} & \xmark & \xmark & \cmark & \cmark & \cmark \\
    \cmidrule{2-11}

    & GRACE~\cite{hartvigsen2023aging} & DIL & \makecell{T5 | BERT | GPT2} 
    & \xmark & \xmark & Adapter & \xmark 
    & \cmark & \cmark & \xmark \\
    \cmidrule{2-11}

    & WilKE~\cite{hu2024wilke} & DIL & \makecell{GPT2 | GPT-J} 
    & \xmark & \xmark & Adaptor & \xmark 
    & \cmark & \cmark & \cmark \\
    \cmidrule{2-11}

    & Larimar~\cite{das2024larimar} & DIL & \makecell{BERT | GPT-J} 
    & \xmark & \xmark & \xmark & Kanerva Memory 
    & \cmark & \cmark & \cmark \\
    \cmidrule{2-11}

    & MELO~\cite{yu2023melo} & DIL & \makecell{BERT | GPT2 | T5} 
    & \xmark & \xmark & LoRA & \xmark 
    & \cmark & \cmark & \cmark \\
    \cmidrule{2-11}

    & CME~\cite{li2023continual} & DIL & BERT 
    & Replay & \xmark & \xmark & Inner-Prod. Reg. 
    & \cmark & \cmark & \cmark \\
    \cmidrule{2-11}

    \multirow{-7}{*}[0em]{{\makecell{CMR}}} & WISE~\cite{wang2024wise} & DIL & \makecell{GPT-J | Llama2 | Mistral} 
    & \xmark & \xmark & \xmark & \makecell{Side Memory}
    & \cmark & \cmark & \cmark \\
    \midrule
    \midrule

    & COPF~\cite{zhang2023copf} & TIL | DIL & Llama 
    & Replay & Function Reg. & Prompt & \xmark 
    & \checkmark & \xmark & \checkmark \\
    \cmidrule{2-11}

    & AMA~\cite{lin2024mitigating} & DIL & \makecell{OpenLLaMA | Mistral} & Replay & L1 | L2 & LoRA & Adaptive Model Avg.
    &\club &\club &\club \\
    \cmidrule{2-11}

    \multirow{-3}{*}[0em]{{\makecell{CMA}}} & CPPO~\cite{zhangcppo} & TIL & GPT2 
    & \xmark & Weighting & Prompt & \xmark 
    & \checkmark & \checkmark & \checkmark \\ 
    \midrule
    \midrule
    
    & EProj~\cite{he2023continual} & TIL & InstructBLIP
    & \xmark & TSIR & Projector Exp. & \xmark 
    & \cmark & \xmark & \cmark\\
    \cmidrule{2-11}
       
    & Fwd-Prompt~\cite{zheng2024antiforgetting} & TIL & \makecell{InstructBLIP | BLIP2} & \xmark & \xmark & Projector Exp. & \xmark 
    & \cmark & \cmark & \cmark\\
    \cmidrule{2-11}

    & CoIN~\cite{chen2024coin} & TIL & LLaVA 
    & \xmark & \xmark & \makecell{MoE | LoRA} & \xmark 
    & \cmark & \xmark & \cmark
    \\
    \cmidrule{2-11}

    & Model~Tailor~\cite{zhu2024model} & TIL & 
    \makecell{InstructBLIP | LLaVA}
    & \xmark & Model~Tailor & \xmark & \xmark 
    & \cmark & \cmark & \cmark \\
    \cmidrule{2-11}

    \multirow{-5}{*}[0em]{{\makecell{CMLLMs}}} &  RebQ~\cite{zhao2024reconstruct} & TIL & ViLT
    & \xmark & \xmark &  Prompt Tuning & \xmark 
    & \cmark & \xmark & \cmark \\
	\bottomrule[0.15em]
	\end{tabular}
	}
\end{table*}

\subsubsection{General Observations on CFT}
\label{sec:cft-obs}

Examining the landscape of continual learning in the context of LLMs, and combined with the results shown in \Tabref{tab:cft}, we make several key observations about CFT. 
\begin{itemize}
    \item \textbf{OBS-1: There has been a noticeable transition in focus from CIL to TIL and DIL.}
    It has been a longstanding common sense in the CL community that CIL, as it requires the model to predict the context label and within-context label at the same time~\cite{van2022three,wang2024comprehensive,kim2022theoretical}, is the most challenging CL scenario and hence receives most of the attention from the community. 
    However, among all 35 papers presented in \Tabref{tab:cft}, only 3 papers study CFT of CIL.
    The transition of the research focus demonstrates the importance of TIL and DIL in the real-world applications of continual LLMs. More detailed discussion of this transition is included in \Secref{sec:discussion-xil}. 
    \item \textbf{OBS-2: In CFT, CL techniques enjoy broader adoption and explicit exploration compared to CPT and DAP.}
    In \Tabref{tab:cft}, all 35 papers explicitly deploy the CL techniques, 50\% of which develop new techniques that cannot be easily interpreted as trivial combination of existing classic CL techniques, e.g., shared attentive learning framework in SAPT~\cite{zhao2024sapt}, external memory deployed in Larimar~\cite{das2024larimar}, and adaptive model averaging method to achieve Pareto-optimal in AMA~\cite{lin2024mitigating}, etc.
    This underscores the recognition of continual learning as a pivotal component in the development of resilient and adaptive LLMs.
\end{itemize}
\setlength{\aboverulesep}{2pt}
\setlength{\belowrulesep}{2pt}

\subsubsection{General Continual Fine-Tuning~(General CFT)}
\label{sec:cft-general}

Researchers have long investigated the phenomenon of forgetting resilience in pre-trained LLMs when fine-tuned for downstream tasks \cite{ke2021achieve,tao2022can,luo2023investigating,zheng2023learn,mehta2023empirical}, despite some discover the opposite~\cite{luo2023investigating}. 
Although the pre-trained weights initially position the model in a flat-loss basin, aiding adaptation to future tasks without severely impacting previous ones \cite{mehta2023empirical}, zero or near-zero forgetting is only observed at the representation level. This implies that while the model retains its ability to distinguish between task-specific representations, it may still forget specific task details \cite{wu2021pretrained,tao2022can,luo2023investigating,zheng2023learn}. Therefore, additional measures are necessary when deploying these models in real-world applications \cite{ke2021achieve,wei2022circle,bai2023enhancing,qin2021lfpt5,weyssow2023usage,chen2024parameterizing}. 

Many studies advance beyond naive sequential fine-tuning, leveraging the inherent anti-forgetting nature of LLMs while avoiding the adoption of overly complex CL techniques~\cite{winata2023overcoming,zheng2023learn}. For instance, LR~ADJUST~\cite{winata2023overcoming} proposes a straightforward yet effective method of dynamically adjusting the learning rate to mitigate the overwriting of knowledge from new languages onto old ones. Building on the innate anti-forgetting ability of large language models like Pythia \cite{biderman2023pythia}, SEQ$^*$ \cite{zheng2023learn} introduces several strategies for fine-tuning LLMs on a sequence of downstream classification tasks, such as freezing the LLM and old classifier's parameters after warm-up, and pre-allocating future classifiers, etc. 

Given the minimal forgetting observed at the representation level in CL, some studies aim to tackle the misalignment between the representation space and the decision-making layers by introducing representation-level constraints during CFT. NeiAttn~\cite{bai2023enhancing} exemplifies this approach by formulating classification tasks as masked language modeling and proposing a neighboring attention mechanism to counteract negative representation drift. 

Another line of approaches refines the input/output format and network architectures of pre-trained LLMs to be better suited for CFT. 
For instance, CTR~\cite{ke2021achieve} incorporates two CL-plugin modules, i.e., a task-specific module~(TSM) for acquiring task-specific knowledge and a knowledge-sharing module~(KSM) for selectively transferring previously learned similar knowledge. 
CIRCLE~\cite{wei2022circle} manually designs diverse prompt templates for various types of buggy code, unifying them as the cloze task and employs difficulty-based replay to enhance continual program repair. 
LFPT5~\cite{qin2021lfpt5} addresses lifelong few-shot language learning by consolidating sequence labeling, text classification, and text generation into a text-to-text generation task. It undergoes prompt tuning on generated pseudo-examples from previous domains when adapting to new tasks. 
In \cite{zhang2022continual}, the authors propose a method for adaptively adding compositional adapters during continual sequence generation tasks. Before training on new domains, a decision stage determines which trained module can be reused. During training, this module also regenerates examples of the past for replay.
C3~\cite{chen2024parameterizing} merges PEFT and in-context learning (ICL) in a teacher-student framework. The teacher model undergoes in-context tuning focused solely on the current domain, while the student model, together with tunable prompts, minimizes the KL-divergence between the output distribution and the ground truth and teacher model simultaneously.


\subsubsection{Continual Instruction Tuning~(CIT)}
\label{sec:cft-cit}
When the instruction tuning data comes in as a stream, 
forgetting of the previously learned instructions should be addressed. 
CT0~\cite{scialom2022fine} represents the inaugural study on Continual Instruction Tuning~(CIT) of LLMs, applying the replay method on the base T0 model throughout the process. 
Many subsequent studies focus on enhancing the replay method used during CIT. For instance, \cite{he2024dont} improve replay efficiency by computing Key-Part Information Gain~(KPIG) on masked parts to dynamically select replay data, addressing the ``half-listening'' issue in instruction following. Similarly, SSR~\cite{huang2024mitigating} uses the LLM to generate synthetic instances for replay, achieving superior or comparable performance to traditional methods at a lower cost.

Other approaches introduce multiple CL techniques during CIT. DynaInst~\cite{mok2023large} merges parameter regularization with dynamic replay, selectively storing and replaying instances and tasks to enhance outcomes. InstructionSpeak~\cite{yin2022contintin} employs negative training and replay instructions to improve both forward transfer and backward transfer. Some methods incorporate PEFT. Orthogonal Low-Rank Adaptation~(O-LoRA) learns new tasks within an orthogonal subspace while preserving LoRA parameters for previous tasks~\cite{wang2023orthogonal} to minimize the interference among different tasks. Shared Attention Framework~(SAPT) combines a PET block with a selection module via a Shared Attentive Learning \& Selection module, tackling catastrophic forgetting and knowledge transfer concurrently~\cite{zhao2024sapt}.
While regularization-based and architectural-based methods require additional parameter storage and GPU memory, together with replay-based methods they remain for CIT due to the simplicity and effectiveness~\cite{wang2024inscl}. 


\subsubsection{Continual Model Refinement~(CMR)}
\label{sec:cft-cmr}
The concept of model editing was initially explored in \cite{sinitsin2020editable}, which introduced a \emph{``reliability-locality-efficiency''} principle and proposed a gradient descent editor to address it efficiently. Subsequent research, such as \cite{de2021editing} and \cite{fast_edit}, extended this principle to edit factual knowledge in BERT-based language models and larger models like GPT-J-6B~\cite{gpt-j} and T5-XXL~\cite{raffel2020exploring}, respectively, using gradient decomposition. These approaches typically update a subset of model parameters to alter the labels of specific inputs. Additionally, memory-based models, as discussed in \cite{mitchell2022memory} and \cite{hartvigsen2023aging}, incorporate editing through retrieval mechanisms.
Continual Model Refinement~(CMR) extends model refinement horizontally, presenting updated sample pairs ${(\vx_e, y_e, \hat{y}_e)}^{e=1}_N$ sequentially as a stream. \cite{lin2022continual} initially introduces this idea, evaluating various CL methods with a dynamic sampling algorithm. Many CMR methods employ a retrieval mechanism. For instance, \cite{hartvigsen2023aging} uses hidden activations of the language model as a ``key'' to activate updated parameters only when input $x_0$ resembles updated sample pairs; 
\cite{yu2023melo} improves this approach's efficiency by integrating LoRA \cite{hu2021lora}; \cite{das2024larimar} augments the LLM with an external episodic memory, modeling CMR as an ongoing memory refresh. 
Meanwhile, some methods focus solely on updating a subset of model parameters. For example, \cite{hu2024wilke} addresses the issue of ``toxicity buildup and flash'' in single-editing methods like ROME \cite{meng2022locating}, adapting it to the CL context with a knowledge-aware layer selection algorithm.
WISE~\cite{wang2024wise} addresses the ``impossible triangle'' of reliability, locality, and generalization in existing lifelong model refinement methods. It introduces a side memory system that enables knowledge sharding and merging, successfully achieving all three objectives simultaneously.

While all these works pioneer research in CMR, the exploration of CMR of LLMs remains open. \cite{hase2023does} highlights a potential problem: the location for storing the fact may not coincide with the best place for editing it. This challenges the classical ``locate and edit'' paradigm used by several existing methods~\cite{meng2022locating, meng2022mass}, and could become a significent concern for CMR~\cite{hu2024wilke}. Other questions, including whether such problem setting fits LLMs and whether more memory/computationally efficient methods of CMR could be developed for LLMs, are yet to be answered.

\subsubsection{Continual Model Alignment~(CMA)}
\label{sec:cft-cma}
When LLMs undergo the phase of MA, vertical forgetting of previous knowledge usually occurs. 
In \cite{lin2024mitigating}, the authors refer to this phenomenon of catastrophic forgetting induced caused by MA as the ``Alignment Tax.'' 
Notably, even a single stage of MA can diminish the model's performance capabilities, as it restricts the model's responses to a narrower subset of the training distribution. 

Continual Model Alignment~(CMA) aims to continuously refine LLMs to align with evolving human values, ethics, and data. 
The static nature of LLM training on historical data sets can lead to discrepancies between the models' outputs and current factual accuracies, societal norms, and standards, making CMA a crucial process for maintaining their adaptability and alignment with contemporary contexts. 
Likewise, there are two types of CMA frameworks: RL-based and SL-based. 
In the realm of RL-based CMA, two significant contributions have been noted.  \cite{lin2024mitigating} identifies the conflicts between the existing CL techniques and RLHF, and proposes Adaptive Model Averaging~(AMA), adaptively finding appropriate ratios for the combination of model layers to gain maximal rewards with minimal tax;  Continual Proximal Policy Optimization~(CPPO)~\cite{zhangcppo} proposes a weighting strategy for different examples deciding its usage of policy enhancement or knowledge retention, mitigating the alignment tax over time. 
For SL-based CMA, Continual Optimal Policy Fitting~(COPF)~\cite{zhang2023copf} presents a solution adapted from the Direct Policy Optimization~(DPO)~\cite{rafailov2024direct}, solving its potential risks of sub-optimal policy fitting and over-optimization in the context of CMA.

\subsubsection{Continual Multimodal Large Language Models~(CMLLMs)}
\label{sec:cft-cmllm}


Continually training multi-modal models like CLIP~\cite{radford2021learning} has been long studied~\cite{zheng2023preventing,ni2023continual}, while the problem of continually training MLLMs still remains underexplored.
Several existing studies have investigated the causes of catastrophic forgetting when continually training MLLMs. \cite{zheng2024antiforgetting} performs singular value decomposition on input embeddings, revealing a significant disparity among different input embeddings. This discrepancy causes the model to learn irrelevant information for previously trained tasks, resulting in catastrophic forgetting and negative forward transfer. 
\cite{zhai2023investigating} observes that minority collapse may lead to catastrophic forgetting, when the imbalance ratio between majority and minority classes approaches infinity during fine-tuning. It further identifies hallucination as a contributing factor to performance degradation in MLLMs. 


\textbf{Continual Fine-Tuning MLLMs.}\quad 
In contrast to traditional continual learning methods that involve full-model fine-tuning for new tasks, continual fine-tuning for MLLMs focuses on refining specific layers when adapting to new tasks~\cite{zhai2023investigating,he2023continual,zheng2024antiforgetting,chen2024coin,zhu2024model}. Given the strong capabilities of pre-trained models, training specific layers suffices, and can simultaneously reduce computational demands. 
\cite{zhao2024reconstruct} additionally considers an continual learning scenario, Continual Missing Modality Learning~(CMML), where different modalities are emerging throughout the incremental learning stages.
All the aforementioned studies collectively indicate that MLLMs still suffer from catastrophic forgetting, which manifests in two ways: along the direction of \emph{vertical continuity}, a performance decline on pre-trained tasks following fine-tuning for downstream tasks; and along the axis of \emph{horizontal continuity}, a performance degrade on previously fine-tuned tasks after fine-tuning for new tasks. \cite{zheng2024antiforgetting} also observes negative forward transfer, where the performance of unseen tasks degrades when learning new tasks, indicating a decline in model generalization capability.

While traditional CL methods are applicable, some may not yield optimal results, as evidenced by various experiments~\cite{he2023continual,zheng2024antiforgetting}. For instance,
\cite{he2023continual} observes a consistent efficacy of replay-based and model expansion strategies across diverse scenarios of continual fine-tuning MLLMs, but regularization-based methods only perform well on models that have been jointly instruction-tuned on multiple tasks. 
Other works seek to develop ad-hoc solutions for continual learning MLLMs.
\cite{he2023continual} proposes EProj to expand the projection layer in MLLMs for each new task and utilizes task-similarity-informed regularization~(TIR) to enhance performance. \cite{zheng2024antiforgetting} introduces Fwd-Prompt, a prompt tuning method that projects prompt gradient to both the residual space and the pre-trained subspace to minimize the interference between tasks and reuse pre-trained knowledge respectively, fostering positive forward transfer without relying on previous samples. 
\cite{zhu2024model} focuses on the forgetting of the pre-trained MLLMs after fine-tuned on specific tasks and proposes model tailor to compensate the selected subset that are critical for enhancing target task performance. 
\cite{zhao2024reconstruct} presents a novel method named Reconstruct before Query~(RebQ), leveraging  the multi-modal knowledge from a pre-trained model to reconstruct the absent information for the missing modality. Recently, MoE~(Mixture-of-Experts) framework has gained attention which resembles the architecture-based methods in CL. It provides the model with
the ability to learn different intentions from distinct experts, e.g., 
\cite{chen2024coin} first introduces MoELoRA to fine-tune LLaVA, effectively mitigate the catastrophic forgetting of MLLMs in CoIN and the results demonstrate the effectiveness.


\section{Evaluation Protocols and Datasets}
\label{sec:eval-and-data}
\textbf{Continual LLMs' Evaluation Protocols.}\quad
LAnguage Model Analysis~(LAMA) is an evaluation framework designed to \emph{probe the world knowledge} embedded in language models~\cite{petroni2019language}.
LAMA converts each world fact into a cloze statement, which is then input into the language models to predict the correct answer. It has been extensively utilized in work on CPT under the temporal shifts~\cite{jang2022temporalwiki,jang2022towards}.
{FUAR~(Forgotten / (Updated + Acquired) Ratio)} is proposed for CPT to address the \textbf{OP}'s drawback of not able to accurately reflect the model's behavior. 
A FUAR value of 1 represents an equal trade-off between the knowledge forgetting and knowledge learning, while a FUAR less than 1 suggests high learning efficacy.
In TRACE~\cite{wang2023trace}, the authors propose a set of ``\textbf{X-Delta}'' metrics for continual instruction tuning, quantifying the forward transfer on specific abilities of LLMs, which is a straightforward extension of \textbf{FWT}. Specifically, the authors construct three sets of evaluation tasks to benchmark the ability of LLMs, including \emph{general ability}, \emph{instruction following}, and \emph{safety}. 
For more detailed introduction to these evaluation protocols, please refer to \appref{app:eval-llm}.

\textbf{Datasets.}\quad
In this section, we provide a comprehensive review of the datasets available for benchmarking continual LLMs, as illustrated in \Tabref{tab:datasets}. 
We provide information about these datasets' types, what distributional shifts and semantic domains they include, and their sources and applications. 
We intentionally exclude datasets used for domain-adaptive pre-training LLMs in vertical domains such as legal, medical, and financial, unless they are specifically designed for continual domain-adaptive pre-training. Furthermore, we omit datasets used in general continual fine-tuning, as they have already been extensively studied in existing works~\cite{biesialska2020continual,ke2023continual}.
For details, please refer to \appref{app:data}.

\begin{sidewaystable}
    \centering
    \caption{
    \textbf{Summary of the existing benchmarks publicly available for Continual Learning LLMs.} 
    In the column of \textbf{Name}, we use the superscript ``$^*$'' to denote the lack of the dataset name and the name shown is that of the original paper. 
    In this table, we deliberately omit the datasets used for domain-adaptive pre-training the vertical LLMs, as their main focus of development is not on continual learning. We also omit the datasets used for general continual fine-tuning, as they are extensively discussed in other existing surveys~\cite{biesialska2020continual,ke2023continual}. 
    }
    \label{tab:datasets}
    \resizebox{1\linewidth}{!}{
\begin{tabular}{ccccc ccccc }
	\toprule[0.15em]
    \textbf{Name} & \textbf{Type} & \textbf{Shift}  & \textbf{Domain} & \textbf{\#Stages} & \textbf{Scale} & \textbf{Sources} & \textbf{Applications} & \textbf{Comment} \\
    \midrule
    \midrule
    $^*$TimeLMs~\cite{loureiro2022timelms} & CPT & Temporal & Social Media & 8 & \#Examples: 123.86M  & Tweets & \cite{loureiro2022timelms} & \href{https://github.com/cardiffnlp/timelms}{code}\\
    \midrule
    CC-RecentNews~\cite{jang2022towards} & CPT & Temporal & News & 1 & \#Tokens: $\sim$168M & Web & \cite{jang2022towards} & \href{https://github.com/joeljang/continual-knowledge-learning}{code}\\
    \midrule
    TWiki~\cite{jang2022temporalwiki} & CPT & Temporal & General Knowledge & 5 & \#Tokens: 4.7B & Wikipedia & \cite{jang2022temporalwiki} & \href{https://github.com/joeljang/temporalwiki}{code} \\
    \midrule
    $^*$DAPT~\cite{gururangan2020dont} & \makecell{CPT \\ DAP} & Content & Multi-Domain & 4 & Size: 160GB & \makecell{BioMed~\cite{lo2020s2orc}, CS~\cite{lo2020s2orc}, News~\cite{zellers2019defending}, Reviews~\cite{he2016ups}} & \makecell{\cite{gururangan2020dont} \\ \cite{qin2023recyclable} \\ \cite{qin2022elle}} & \href{https://github.com/allenai/dont-stop-pretraining}{code} \\
    \midrule
    $^*$CPT~\cite{ke2022continual-train} & CPT & Content & Multi-Domain & 4 & \#Examples: 3.12M & \makecell{Yelp~\cite{xu2019bert}, S2ORC~\cite{lo2020s2orc}, AG-News~\cite{zhang2015character}} & \cite{ke2022continual-train} & \href{https://github.com/UIC-Liu-Lab/CPT}{code} \\
    \midrule
    $^*$DEMix~\cite{gururangan2022demix} & CPT & Content & Multi-Domain & 8 & \#Tokens: 73.8B & \makecell{1B~\cite{chelba2014billion}, CS~\cite{lo2020s2orc}, Legal~\cite{caselaw2018}, Med~\cite{lo2020s2orc}\\ WebText~\cite{gokaslan2019OpenWeb}, RealNews~\cite{zellers2019defending}, Reddit~\cite{baumgartner2020pushshift}, Reviews~\cite{ni2019justifying}} & \cite{gururangan2022demix} & \href{https://github.com/kernelmachine/demix}{code} \\
    \midrule
    $^*$DAS~\cite{ke2022continual-pre} & \makecell{CPT \\ DAP} & Content & Multi-Domain & 6 & Size: 4.16GB & \makecell{Yelp~\cite{xu2019bert}, Reviews~\cite{ni2019justifying}, Papers~\cite{lo2020s2orc}, PubMed} & \cite{ke2022continual-pre} & \href{https://github.com/UIC-Liu-Lab/ContinualLM}{code} \\
    \midrule
    SuperNI~\cite{wang2022supernaturalinstructions} & CIT & Content & Mutli-Domain & 16 & \makecell{\#Tasks: 1616 \\ \#Examples: $\sim$5M} & GitHub & \cite{zhang2023citb,wang2024inscl} & \href{https://github.com/allenai/natural-instructions}{code} \\

    \midrule
    CITB~\cite{zhang2023citb} & CIT & Content & Mutli-Domain & 19 & \#Tasks: 38 & SuperNI~\cite{wang2022supernaturalinstructions}  & \cite{zhang2023citb} & \href{https://github.com/hyintell/CITB}{code} \\

    \midrule
    CoIN~\cite{chen2024coin} & CIT & Content & Multi-Domain & 8 &\#Examples: $\sim$1.14M & \makecell{RefCOCO~\cite{kazemzadeh-etal-2014-referitgame},RefCOCO+~\cite{mao2016generation},RefCOCOg~\cite{mao2016generation} \\ ImageNet~\cite{imagenet_cvpr09}, VQAv2~\cite{goyal2017making}, ScienceQA~\cite{lu2022learn} \\ 
    TextVQA ~\cite{singh2019vqa}, GQA~\cite{hudson2019gqa}, VizWiz ~\cite{gurari2018vizwiz}, OCR-VQA~\cite{mishraICDAR19}} & \cite{chen2024coin} & \href{https://github.com/zackschen/CoIN}{code} \\
    \midrule
    TRACE~\cite{wang2023trace} & CIT & Content & Mutli-Domain & 8 &  \#Examples: 56,000  & \makecell{ScienceQA~\cite{lu2022learn}, FOMC~\cite{shah2023trillion}, MeetingBank~\cite{hu2023meetingbank}\\ 
    C-STANCE~\cite{zhao-etal-2023-c}, 20Minuten~\cite{kew-etal-2023-20}, CodeXGLUE~\cite{lu2021codexglue}, NumGLUE\cite{mishra2022numglue}} & \cite{wang2023trace} & \href{https://github.com/BeyonderXX/TRACE}{code} \\
    \midrule
    NATURAL-INSTRUCTION~\cite{mishra2021natural} & CIT & Content & Mutli-Domain & 6 &   \#Examples: 193k & \makecell{CosmosQA~\cite{huang2019cosmos}, DROP~\cite{dua2019drop}, Essential-Terms~\cite{khashabi-etal-2017-learning} \\ MCTACO~\cite{zhou2019goingvacationtakeslonger}, MultiRC~\cite{khashabi-etal-2018-looking}, QASC~\cite{khot2020qasc} \\ Quoref\cite{dasigi-etal-2019-quoref}~, ROPES~\cite{lin2019reasoning} , Winogrande~\cite{sakaguchi2019winogrande}} & \cite{mishra2021natural} & \href{https://github.com/allenai/natural-instructions-v1}{code} \\
    \midrule
    IMDB~\cite{maas2011learning} & CMA & Content &Social Media&1&Size: 217.35 MB& IMDB& \cite{zhang2023copf} & \href{https://huggingface.co/datasets/stanfordnlp/imdb}{code}\\
    \midrule 
    HH-RLHF~\cite{bai2022training} & CMA & Content &General Knowledge&1&Size: 28.1 MB& Human Feedback& \cite{zhang2023copf} & \href{https://github.com/anthropics/hh-rlhf}{code}\\
    \midrule
    Reddit TL;DR~\cite{volske2017tl} & CMA & Content&Social Media&2&Size: 19.6 GB&Reddit & \cite{zhang2023copf,zhangcppo} &\href{https://zenodo.org/records/1043504}{code}\\
    \midrule
    \makecell{Common Sense QA~\cite{lin2024mitigating} \\ Reading Comprehension~\cite{lin2024mitigating}\\ 
    Translation~\cite{lin2024mitigating}} & CMA & Content &Multi-Domain&6& \#Examples: $\sim$ 41.16M&  \makecell{ARC Easy and Challenge~\cite{clark2018think}, Race~\cite{lai2017race}, PIQA~\cite{bisk2020piqa} \\ SQuAD~\cite{rajpurkar2018know}, DROP~\cite{dua2019drop} \\
    WMT 2014 French to English~\cite{bojar2014findings}
    }& \cite{lin2024mitigating} & see sources \\
    \midrule
    FEVER~\cite{fever} & CMR & Content & General Knowledge & 1 & \#Examples: 420k &  Wikipedia & \cite{de2021editing, hase2021language} & \href{https://fever.ai/resources.html}{code} \\
    \midrule
    VitaminC~\cite{vitaminC} & CMR & Content & General Knowledge & 1 & \#Examples: 450k &  Wikipedia & \cite{mitchell2022memory} & \href{https://github.com/TalSchuster/VitaminC}{code} \\
    \midrule
    zsRE~\cite{zsRE} & CMR & Content & General Knowledge & 1 & \#Examples: 120M &  Wikireading~\cite{Wikireading} & \cite{hase2021language, meng2022locating, meng2022mass, hase2023does, hartvigsen2023aging, das2024larimar} & - \\
    \midrule
    T-rex~\cite{T-rex} & CMR & Content & General Knowledge & 1 & \#Examples: 11M &  Dbpedia abstracts~\cite{Dbpedia} & \cite{li2022large, dong2022calibrating} & \href{https://hadyelsahar.github.io/t-rex/}{code} \\
    \midrule
    NQ~\cite{nq} & CMR & Content & General Knowledge & 1 & \#Examples: 320k &  Google queries, Wikipedia & \cite{hartvigsen2023aging} & \href{https://ai.google.com/research/NaturalQuestions}{code} \\
    \midrule
    CounterFact~\cite{meng2022locating} & CMR & Content & General Knowledge & 1 & \#Examples: 22k & zsRE \cite{zsRE} & \cite{meng2022locating, yu2023melo, hu2024wilke, das2024larimar} & \href{https://github.com/kmeng01/rome}{code} \\
    \midrule
    SCOTUS~\cite{scotus} & CMR & Temporal & Law & 1 & \#Examples: 9.2k &  Supreme Court Database & \cite{hartvigsen2023aging} & \href{https://github.com/coastalcph/fairlex}{code} \\
	\bottomrule[0.15em]
	\end{tabular}
	}
\end{sidewaystable}

\section{Discussion}
\label{sec:discussion}

\subsection{Intriguing Properties Emergent in Continual LLMs}
\label{sec:discussion-emergent}
Beyond the well-established resilience of pre-trained large language models~(LLMs) against catastrophic forgetting compared to downstream-specific models~\cite{ke2021achieve,tao2022can,luo2023investigating,zheng2023learn,mehta2023empirical}, there is a notable lack of exploration into other intriguing properties of LLMs when trained continually. In \cite{yang2024reawakening}, it is observed that when fine-tuned sequentially and cyclically on a series of documents, large models exhibit a phenomenon known as ``\emph{anticipatory recovering}.'' This refers to the LLMs' ability to recover forgotten information on documents even before encountering them again. 
This suggests that LLMs may possess the capability of sequential memorization, which could pave the way for research into more complex structured learning environments as model parameters scale up.

\subsection{Conventional Types of Incremental Learning}
\label{sec:discussion-xil}
As mentioned in \Secref{sec:background-cl-types}, three types of incremental learning are prevalent~\cite{van2022three}. Among them, class-incremental learning~(CIL) has historically attracted significant attention from the community~\cite{rebuffi2017icarl,wu2019large}. However, in the context of continually pre-training and adapting large language models~(LLMs), we observe a decreased interest in CIL but an increased focus on task-incremental learning~(TIL) and domain-incremental learning~(DIL).
Given that language models are inherently designed for content generation and are pre-trained with the pretext generative task of next-word prediction, it is natural to emphasize the patterns of generative tasks and integrate the traditional CIL paradigm into the broader framework of language modeling, discarding the incremental classification head~\cite{shao2023class,cao2024generative}. 
However, the declining attention to CIL does not suggest that it is not impactful in the field of continual learning for LLMs. 
Techniques such as vocabulary expansion~\cite{amba2021dynamic,cossu2022continual} and learning routing function in the MoE system~\cite{chen2023lifelong} can be seen as an extension of expanding the classification head in CIL, and previously validated techniques of CIL can be directly applied.

The importance of DIL is self-evident, given the shared task definition and input-output format in continual pre-training~(CPT) and domain-adaptive pre-training~(DAP). 
On the other hand, TIL attracts significant interest 
as it plays a crucial role in instruction tuning, where instructions can be seen as natural-language-encoded task indices~\cite{scialom2022fine,huang2024mitigating,mok2023large,he2024dont,yin2022contintin,wang2023orthogonal,zhao2024sapt,wang2024inscl}.
It is worth noting that the boundary between TIL and DIL becomes somewhat blurred in continual instruction tuning. Language models demonstrate the capability to infer domain information for unseen instructions, suggesting a convergence of TIL and DIL in certain contexts.

\subsection{Roles of Memory in Continual LLMs}
\label{sec:discussion-mem}
Previous continual learning research, drawing inspiration from human learning patterns, primarily emphasizes the storage efficiency of past data. However, this focus may no longer hold true in the context of continual LLMs. In the direction of relaxing memory constraints, institutions with access to training data may opt to retain full access without restricting memory size, given that the cost of memory storage is more than affordable. In such scenarios, as highlighted in \cite{verwimp2024continual}, the challenge shifts from storage efficiency to computational efficiency. To achieve continual learning goals, models must efficiently adapt to new data (efficient adaptation) and select key experiences for replay (efficient replay)~\cite{xie2023efficient,jin2024model}. Therefore, it is essential to reassess the existing memory constraint and prioritize optimizing computational efficiency for continual learning of LLMs by restricting the number of updates and FLOPs~\cite{prabhu2023computationally,wang2022sparcl}.

On the other end of the spectrum, studies with tightened memory constraints remain vital in modern continual learning of LLMs. As shown in \Figref{fig:overview}, upstream suppliers of LLMs typically do not provide training data with the released model weights. Consequently, consumers must adapt these models to downstream data without access to the actual replay data. Various rehearsal-free continual strategies are applied in this scenario, such as collecting data examples from alternate sources~\cite{rozière2024code,colombo2024saullm7b,wu2023pmc,Azerbayev2023LLEMMA}, leveraging the generative capabilities of LLMs to produce pseudo-examples for replay~\cite{qin2021lfpt5}, and implementing regularization techniques in the parameter space~\cite{ke2022continual-pre,rongali2021continual}. Continual learning under the strict memory constraint is also driven by data privacy concerns, where preserving data on the server side is prohibited. In these scenarios, researchers must rely on online continual learning methods~\cite{mai2022online,prabhu2023online}, where data examples are only utilized for training as they arrive in a stream, and numerous efforts are already underway to develop LLMs capable of operating under these constraints~\cite{bornschein2024transformers}.

\subsection{Prospective Directions}
\label{sec:discussion-future}
\textbf{Theories of Continual LLMs.}\quad 
It is widely recognized that the continual learning community tends to prioritize empirical research over theoretical exploration. Nevertheless, there are efforts to establish theoretical foundations for CL. In \cite{wang2024comprehensive}, the authors utilize second-order Taylor expansions around optimal parameters to derive an inter-task generalization error bound based on the maximum eigenvalue and $l_2$-norm of parameter differences. Another line of approaches leverages task/domain discrepancies to construct a multi-task generalization bound. For instance, Unified Domain Incremental Learning~(UDIL) in \cite{shi2024unified} proposes upper bounds for intra-domain and cross-domain distillation losses, unifying various replay-based DIL techniques under a single adaptive generalization bound. However, applying these existing theories directly to continual LLMs can be imprudent, given their pre-trained, large-scale nature. Consequently, there is a notable gap in research focusing on continually learning LLMs with robust theoretical guarantees and understanding the forgetting behaviors of LLMs from a theoretical perspective.

\textbf{Efficient Replay for Knowledge Retention for Continual LLMs.}\quad
While the storage budget can theoretically be infinite (\Secref{sec:discussion-mem}), replaying past experiences without specific design can lead to inefficient updates in current domain learning, resulting in slow convergence. Beyond sparse replay solutions that control data mixture ratios \cite{lin2023geogalactica,rozière2024code,Yang2023PLLaMa}, there is ongoing exploration of efficient replay for continual LLMs. 
For example, KPIG~\cite{he2024dont} enhances replay efficiency by calculating Key-Part Information Gain~(KPIG) on masked segments, enabling the dynamic selection of replay data. 
\cite{jin2024model} introduces a forgetting forecasting mechanism based on output changes during adaptation, later used for selective replay in continual model refinement~(CMR). 
More sophisticated and accurate data mixing strategies and efficient replay sample selection mechanisms are needed and hence we mark it as a significant research focus in the future.

\textbf{Continual LLMs with Controllable Memory.}\quad
The long-term memory inherent in the whole set of parameters of LLMs often lacks interpretability and explicit manipulability, which is crucial in certain application areas such as machine unlearning~\cite{bourtoule2020machine}, where the continually pre-trained models need to constantly roll back to a previous version predating the inclusion of the revoked data and retrain the model from that point onward.
This example illustrates the benefits of equipping LLMs with an external, controllable memory.
As part of continual model refinement~(CMR), memory systems for continual learning have been explored in several studies. Larimar~\cite{das2024larimar} suggests integrating the Kanerva Machine~\cite{wu2018kanerva} as an episodic memory for multi-fact model editing. This memory system supports basic operations like \emph{writing, reading, and generating}, as well as advanced operations such as \emph{sequential writing and forgetting}. It enables one-shot knowledge updates without costly retraining or fine-tuning. Other memory systems like Hopfield Networks~\cite{ramsauer2021hopfield} hold promise for future investigation as well.

\textbf{Continual LLMs with Custom Preferences.}\quad
In service-oriented contexts, users often require different trade-offs between domain expertise, ethics, values, or tones of expression. Efficiently building customized LLMs for individual users and offering flexible adjustment options is a challenging task. Early attempts in this direction include Imprecise Bayesian Continual Learning~(IBCL), which, under certain assumptions, guarantees the generation of Pareto-optimal models based on user preferences by combining two model posteriors in the parameter space~\cite{lu2023ibcl}. While empirical validation is limited in scale, this approach paves the way for future research in this area.

\section{Conclusion}
\label{sec:conclusion}
In this work, we offer a comprehensive survey on continual LLMs, summarizing recent advancements in their training and deployment from a continual learning standpoint. We categorize the problems and tasks based on their positions within our proposed broader framework of modern stratified continual learning of LLMs. While there is a widespread and growing interest in this area across the community, we also note several missing cornerstones, including algorithmic diversity and a fundamental understanding of large models' behaviors such as knowledge forgetting, transfer, and acquisition. With a holistic yet detailed approach, we aim for this survey to inspire more practitioners to explore continual learning techniques, ultimately contributing to the development of robust and self-evolving AI systems.

{
\bibliographystyle{abbrv}
\bibliography{main}

\begin{thebibliography}{100}

\bibitem{achiam2023gpt}
J.~Achiam, S.~Adler, S.~Agarwal, L.~Ahmad, I.~Akkaya, F.~L. Aleman, D.~Almeida, J.~Altenschmidt, S.~Altman, S.~Anadkat, et~al.
\newblock Gpt-4 technical report.
\newblock {\em arXiv preprint arXiv:2303.08774}, 2023.

\bibitem{acikgoz2024hippocrates}
E.~C. Acikgoz, O.~B. {\.I}nce, R.~Bench, A.~A. Boz, {\.I}.~Kesen, A.~Erdem, and E.~Erdem.
\newblock Hippocrates: An open-source framework for advancing large language models in healthcare.
\newblock {\em arXiv preprint arXiv:2404.16621}, 2024.

\bibitem{agarwal2024structured}
M.~Agarwal, Y.~Shen, B.~Wang, Y.~Kim, and J.~Chen.
\newblock Structured code representations enable data-efficient adaptation of code language models, 2024.

\bibitem{aljundi2018memory}
R.~Aljundi, F.~Babiloni, M.~Elhoseiny, M.~Rohrbach, and T.~Tuytelaars.
\newblock Memory aware synapses: Learning what (not) to forget.
\newblock In {\em Proceedings of the European conference on computer vision (ECCV)}, pages 139--154, 2018.

\bibitem{amba2021dynamic}
S.~Amba~Hombaiah, T.~Chen, M.~Zhang, M.~Bendersky, and M.~Najork.
\newblock Dynamic language models for continuously evolving content.
\newblock In {\em Proceedings of the 27th ACM SIGKDD Conference on Knowledge Discovery \& Data Mining}, pages 2514--2524, 2021.

\bibitem{anil2023palm}
R.~Anil, A.~M. Dai, O.~Firat, M.~Johnson, D.~Lepikhin, A.~Passos, S.~Shakeri, E.~Taropa, P.~Bailey, Z.~Chen, et~al.
\newblock Palm 2 technical report.
\newblock {\em arXiv preprint arXiv:2305.10403}, 2023.

\bibitem{araci2019finbert}
D.~Araci.
\newblock Finbert: Financial sentiment analysis with pre-trained language models, 2019.

\bibitem{attanasio2023worth}
G.~Attanasio, D.~Nozza, F.~Bianchi, and D.~Hovy.
\newblock Is it worth the (environmental) cost? limited evidence for temporal adaptation via continuous training, 2023.

\bibitem{Azerbayev2023LLEMMA}
Z.~Azerbayev, H.~Schoelkopf, K.~Paster, M.~D. Santos, S.~McAleer, A.~Q. Jiang, J.~Deng, S.~Biderman, and S.~Welleck.
\newblock Llemma: An open language model for mathematics.
\newblock {\em CoRR}, abs/2310.10631, 2023.

\bibitem{bai2023enhancing}
X.~Bai, J.~Shang, Y.~Sun, and N.~Balasubramanian.
\newblock Enhancing continual learning with global prototypes: Counteracting negative representation drift, 2023.

\bibitem{bai2022training}
Y.~Bai, A.~Jones, K.~Ndousse, A.~Askell, A.~Chen, N.~DasSarma, D.~Drain, S.~Fort, D.~Ganguli, T.~Henighan, et~al.
\newblock Training a helpful and harmless assistant with reinforcement learning from human feedback.
\newblock {\em arXiv preprint arXiv:2204.05862}, 2022.

\bibitem{banerjee2005meteor}
S.~Banerjee and A.~Lavie.
\newblock Meteor: An automatic metric for mt evaluation with improved correlation with human judgments.
\newblock In {\em Proceedings of the acl workshop on intrinsic and extrinsic evaluation measures for machine translation and/or summarization}, pages 65--72, 2005.

\bibitem{baumgartner2020pushshift}
J.~Baumgartner, S.~Zannettou, B.~Keegan, M.~Squire, and J.~Blackburn.
\newblock The pushshift reddit dataset, 2020.

\bibitem{ben2010theory}
S.~Ben-David, J.~Blitzer, K.~Crammer, A.~Kulesza, F.~Pereira, and J.~W. Vaughan.
\newblock A theory of learning from different domains.
\newblock {\em Machine learning}, 79:151--175, 2010.

\bibitem{Bi2023OCEANGPT}
Z.~Bi, N.~Zhang, Y.~Xue, Y.~Ou, D.~Ji, G.~Zheng, and H.~Chen.
\newblock Oceangpt: {A} large language model for ocean science tasks.
\newblock {\em CoRR}, abs/2310.02031, 2023.

\bibitem{biderman2023pythia}
S.~Biderman, H.~Schoelkopf, Q.~G. Anthony, H.~Bradley, K.~O’Brien, E.~Hallahan, M.~A. Khan, S.~Purohit, U.~S. Prashanth, E.~Raff, et~al.
\newblock Pythia: A suite for analyzing large language models across training and scaling.
\newblock In {\em International Conference on Machine Learning}, pages 2397--2430. PMLR, 2023.

\bibitem{biesialska2020continual}
M.~Biesialska, K.~Biesialska, and M.~R. Costa-juss{\`a}.
\newblock Continual lifelong learning in natural language processing: A survey.
\newblock In D.~Scott, N.~Bel, and C.~Zong, editors, {\em Proceedings of the 28th International Conference on Computational Linguistics}, pages 6523--6541, Barcelona, Spain (Online), Dec. 2020. International Committee on Computational Linguistics.

\bibitem{bisk2020piqa}
Y.~Bisk, R.~Zellers, J.~Gao, Y.~Choi, et~al.
\newblock Piqa: Reasoning about physical commonsense in natural language.
\newblock In {\em Proceedings of the AAAI conference on artificial intelligence}, volume~34, pages 7432--7439, 2020.

\bibitem{bojar2014findings}
O.~Bojar, C.~Buck, C.~Federmann, B.~Haddow, P.~Koehn, J.~Leveling, C.~Monz, P.~Pecina, M.~Post, H.~Saint-Amand, et~al.
\newblock Findings of the 2014 workshop on statistical machine translation.
\newblock In {\em Proceedings of the ninth workshop on statistical machine translation}, pages 12--58, 2014.

\bibitem{bornschein2024transformers}
J.~Bornschein, Y.~Li, and A.~Rannen-Triki.
\newblock Transformers for supervised online continual learning, 2024.

\bibitem{bourtoule2020machine}
L.~Bourtoule, V.~Chandrasekaran, C.~A. Choquette-Choo, H.~Jia, A.~Travers, B.~Zhang, D.~Lie, and N.~Papernot.
\newblock Machine unlearning, 2020.

\bibitem{brown2020language}
T.~Brown, B.~Mann, N.~Ryder, M.~Subbiah, J.~D. Kaplan, P.~Dhariwal, A.~Neelakantan, P.~Shyam, G.~Sastry, A.~Askell, et~al.
\newblock Language models are few-shot learners.
\newblock {\em Advances in neural information processing systems}, 33:1877--1901, 2020.

\bibitem{Dbpedia}
M.~Br{\"u}mmer, M.~Dojchinovski, and S.~Hellmann.
\newblock Dbpedia abstracts: A large-scale, open, multilingual nlp training corpus.
\newblock In {\em Proceedings of the Tenth International Conference on Language Resources and Evaluation (LREC'16)}, pages 3339--3343, 2016.

\bibitem{buzzega2020dark}
P.~Buzzega, M.~Boschini, A.~Porrello, D.~Abati, and S.~Calderara.
\newblock Dark experience for general continual learning: a strong, simple baseline.
\newblock {\em Advances in neural information processing systems}, 33:15920--15930, 2020.

\bibitem{Cao2023InstructMol}
H.~Cao, Z.~Liu, X.~Lu, Y.~Yao, and Y.~Li.
\newblock Instructmol: Multi-modal integration for building a versatile and reliable molecular assistant in drug discovery.
\newblock {\em CoRR}, abs/2311.16208, 2023.

\bibitem{cao2024generative}
X.~Cao, H.~Lu, L.~Huang, X.~Liu, and M.-M. Cheng.
\newblock Generative multi-modal models are good class incremental learners.
\newblock {\em IEEE Computer Vision and Pattern Recognition (CVPR)}, 2024.

\bibitem{caselaw2018}
{Caselaw Access Project}.
\newblock Caselaw access project, 2018.

\bibitem{scotus}
I.~Chalkidis, T.~Pasini, S.~Zhang, L.~Tomada, S.~F. Schwemer, and A.~S{\o}gaard.
\newblock Fairlex: A multilingual benchmark for evaluating fairness in legal text processing.
\newblock {\em arXiv preprint arXiv:2203.07228}, 2022.

\bibitem{chaudhry2019efficient}
A.~Chaudhry, M.~Ranzato, M.~Rohrbach, and M.~Elhoseiny.
\newblock Efficient lifelong learning with a-gem.
\newblock In {\em ICLR}, 2019.

\bibitem{chaudhry2019tiny}
A.~Chaudhry, M.~Rohrbach, M.~Elhoseiny, T.~Ajanthan, P.~K. Dokania, P.~H. Torr, and M.~Ranzato.
\newblock On tiny episodic memories in continual learning.
\newblock {\em arXiv preprint arXiv:1902.10486}, 2019.

\bibitem{chelba2014billion}
C.~Chelba, T.~Mikolov, M.~Schuster, Q.~Ge, T.~Brants, P.~Koehn, and T.~Robinson.
\newblock One billion word benchmark for measuring progress in statistical language modeling, 2014.

\bibitem{chen2024coin}
C.~Chen, J.~Zhu, X.~Luo, H.~Shen, L.~Gao, and J.~Song.
\newblock Coin: A benchmark of continual instruction tuning for multimodel large language model, 2024.

\bibitem{Chen2023HuatuoGPTII}
J.~Chen, X.~Wang, A.~Gao, F.~Jiang, S.~Chen, H.~Zhang, D.~Song, W.~Xie, C.~Kong, J.~Li, X.~Wan, H.~Li, and B.~Wang.
\newblock Huatuogpt-ii, one-stage training for medical adaption of llms.
\newblock {\em CoRR}, abs/2311.09774, 2023.

\bibitem{chen2020recall}
S.~Chen, Y.~Hou, Y.~Cui, W.~Che, T.~Liu, and X.~Yu.
\newblock Recall and learn: Fine-tuning deep pretrained language models with less forgetting.
\newblock In B.~Webber, T.~Cohn, Y.~He, and Y.~Liu, editors, {\em Proceedings of the 2020 Conference on Empirical Methods in Natural Language Processing (EMNLP)}, pages 7870--7881, Online, Nov. 2020. Association for Computational Linguistics.

\bibitem{chen2023lifelong}
W.~Chen, Y.~Zhou, N.~Du, Y.~Huang, J.~Laudon, Z.~Chen, and C.~Cui.
\newblock Lifelong language pretraining with distribution-specialized experts.
\newblock In {\em International Conference on Machine Learning}, pages 5383--5395. PMLR, 2023.

\bibitem{chen2024take}
X.~Chen, Z.~Wang, D.~Sow, J.~Yang, T.~Chen, Y.~Liang, M.~Zhou, and Z.~Wang.
\newblock Take the bull by the horns: Hard sample-reweighted continual training improves llm generalization.
\newblock {\em arXiv preprint arXiv:2402.14270}, 2024.

\bibitem{chen2024parameterizing}
Y.~Chen, S.~Zhang, G.~Qi, and X.~Guo.
\newblock Parameterizing context: Unleashing the power of parameter-efficient fine-tuning and in-context tuning for continual table semantic parsing.
\newblock {\em Advances in Neural Information Processing Systems}, 36, 2024.

\bibitem{chen2018lifelong}
Z.~Chen and B.~Liu.
\newblock {\em Lifelong machine learning}, volume~1.
\newblock Springer.

\bibitem{cheng2024adapting}
D.~Cheng, S.~Huang, and F.~Wei.
\newblock Adapting large language models via reading comprehension, 2024.

\bibitem{chowdhery2023palm}
A.~Chowdhery, S.~Narang, J.~Devlin, M.~Bosma, G.~Mishra, A.~Roberts, P.~Barham, H.~W. Chung, C.~Sutton, S.~Gehrmann, et~al.
\newblock Palm: Scaling language modeling with pathways.
\newblock {\em Journal of Machine Learning Research}, 24(240):1--113, 2023.

\bibitem{clark2018think}
P.~Clark, I.~Cowhey, O.~Etzioni, T.~Khot, A.~Sabharwal, C.~Schoenick, and O.~Tafjord.
\newblock Think you have solved question answering? try arc, the ai2 reasoning challenge.
\newblock {\em arXiv preprint arXiv:1803.05457}, 2018.

\bibitem{colombo2024saullm7b}
P.~Colombo, T.~P. Pires, M.~Boudiaf, D.~Culver, R.~Melo, C.~Corro, A.~F.~T. Martins, F.~Esposito, V.~L. Raposo, S.~Morgado, and M.~Desa.
\newblock Saullm-7b: A pioneering large language model for law, 2024.

\bibitem{together2023redpajama}
T.~Computer.
\newblock Redpajama: an open dataset for training large language models, 2023.

\bibitem{cossu2022continual}
A.~Cossu, T.~Tuytelaars, A.~Carta, L.~Passaro, V.~Lomonaco, and D.~Bacciu.
\newblock Continual pre-training mitigates forgetting in language and vision, 2022.

\bibitem{das2024larimar}
P.~Das, S.~Chaudhury, E.~Nelson, I.~Melnyk, S.~Swaminathan, S.~Dai, A.~Lozano, G.~Kollias, V.~Chenthamarakshan, S.~Dan, et~al.
\newblock Larimar: Large language models with episodic memory control.
\newblock {\em arXiv preprint arXiv:2403.11901}, 2024.

\bibitem{dasigi-etal-2019-quoref}
P.~Dasigi, N.~F. Liu, A.~Marasovi{\'c}, N.~A. Smith, and M.~Gardner.
\newblock {Q}uoref: A reading comprehension dataset with questions requiring coreferential reasoning.
\newblock In K.~Inui, J.~Jiang, V.~Ng, and X.~Wan, editors, {\em Proceedings of the 2019 Conference on Empirical Methods in Natural Language Processing and the 9th International Joint Conference on Natural Language Processing (EMNLP-IJCNLP)}, pages 5925--5932, Hong Kong, China, Nov. 2019. Association for Computational Linguistics.

\bibitem{de2021editing}
N.~De~Cao, W.~Aziz, and I.~Titov.
\newblock Editing factual knowledge in language models.
\newblock {\em arXiv preprint arXiv:2104.08164}, 2021.

\bibitem{deng2023learning}
C.~Deng, T.~Zhang, Z.~He, Y.~Xu, Q.~Chen, Y.~Shi, L.~Fu, W.~Zhang, X.~Wang, C.~Zhou, Z.~Lin, and J.~He.
\newblock K2: A foundation language model for geoscience knowledge understanding and utilization, 2023.

\bibitem{imagenet_cvpr09}
J.~Deng, W.~Dong, R.~Socher, L.-J. Li, K.~Li, and L.~Fei-Fei.
\newblock {ImageNet: A Large-Scale Hierarchical Image Database}.
\newblock In {\em CVPR09}, 2009.

\bibitem{dettmers2023qlora}
T.~Dettmers, A.~Pagnoni, A.~Holtzman, and L.~Zettlemoyer.
\newblock Qlora: Efficient finetuning of quantized llms.
\newblock {\em arXiv preprint arXiv:2305.14314}, 2023.

\bibitem{devlin2018bert}
J.~Devlin, M.-W. Chang, K.~Lee, and K.~Toutanova.
\newblock Bert: Pre-training of deep bidirectional transformers for language understanding.
\newblock {\em arXiv preprint arXiv:1810.04805}, 2018.

\bibitem{dhingra2022time}
B.~Dhingra, J.~R. Cole, J.~M. Eisenschlos, D.~Gillick, J.~Eisenstein, and W.~W. Cohen.
\newblock Time-aware language models as temporal knowledge bases.
\newblock {\em Transactions of the Association for Computational Linguistics}, 10:257--273, 2022.

\bibitem{dong2022calibrating}
Q.~Dong, D.~Dai, Y.~Song, J.~Xu, Z.~Sui, and L.~Li.
\newblock Calibrating factual knowledge in pretrained language models.
\newblock {\em arXiv preprint arXiv:2210.03329}, 2022.

\bibitem{dosovitskiy2020image}
A.~Dosovitskiy, L.~Beyer, A.~Kolesnikov, D.~Weissenborn, X.~Zhai, T.~Unterthiner, M.~Dehghani, M.~Minderer, G.~Heigold, S.~Gelly, et~al.
\newblock An image is worth 16x16 words: Transformers for image recognition at scale.
\newblock {\em arXiv preprint arXiv:2010.11929}, 2020.

\bibitem{dou2024sailor}
L.~Dou, Q.~Liu, G.~Zeng, J.~Guo, J.~Zhou, W.~Lu, and M.~Lin.
\newblock Sailor: Open language models for south-east asia.
\newblock {\em arXiv preprint arXiv:2404.03608}, 2024.

\bibitem{dua2019drop}
D.~Dua, Y.~Wang, P.~Dasigi, G.~Stanovsky, S.~Singh, and M.~Gardner.
\newblock Drop: A reading comprehension benchmark requiring discrete reasoning over paragraphs.
\newblock {\em arXiv preprint arXiv:1903.00161}, 2019.

\bibitem{ebrahimi2019uncertainty}
S.~Ebrahimi, M.~Elhoseiny, T.~Darrell, and M.~Rohrbach.
\newblock Uncertainty-guided continual learning with bayesian neural networks.
\newblock {\em arXiv preprint arXiv:1906.02425}, 2019.

\bibitem{ebrahimi2020adversarial}
S.~Ebrahimi, F.~Meier, R.~Calandra, T.~Darrell, and M.~Rohrbach.
\newblock Adversarial continual learning.
\newblock In {\em Computer Vision--ECCV 2020: 16th European Conference, Glasgow, UK, August 23--28, 2020, Proceedings, Part XI 16}, pages 386--402. Springer, 2020.

\bibitem{T-rex}
H.~Elsahar, P.~Vougiouklis, A.~Remaci, C.~Gravier, J.~Hare, F.~Laforest, and E.~Simperl.
\newblock T-rex: A large scale alignment of natural language with knowledge base triples.
\newblock In {\em Proceedings of the Eleventh International Conference on Language Resources and Evaluation (LREC 2018)}, 2018.

\bibitem{fujii2024continual}
K.~Fujii, T.~Nakamura, M.~Loem, H.~Iida, M.~Ohi, K.~Hattori, H.~Shota, S.~Mizuki, R.~Yokota, and N.~Okazaki.
\newblock Continual pre-training for cross-lingual llm adaptation: Enhancing japanese language capabilities.
\newblock {\em arXiv preprint arXiv:2404.17790}, 2024.

\bibitem{ganin2016domain}
Y.~Ganin, E.~Ustinova, H.~Ajakan, P.~Germain, H.~Larochelle, F.~Laviolette, M.~Marchand, and V.~Lempitsky.
\newblock Domain-adversarial training of neural networks.
\newblock {\em The journal of machine learning research}, 17(1):2096--2030, 2016.

\bibitem{garg2024tic}
S.~Garg, M.~Farajtabar, H.~Pouransari, R.~Vemulapalli, S.~Mehta, O.~Tuzel, V.~Shankar, and F.~Faghri.
\newblock Tic-clip: Continual training of clip models.
\newblock In {\em The Twelfth International Conference on Learning Representations (ICLR)}, 2024.

\bibitem{gogoulou2024continual}
E.~Gogoulou, T.~Lesort, M.~Boman, and J.~Nivre.
\newblock Continual learning under language shift, 2024.

\bibitem{gokaslan2019OpenWeb}
A.~Gokaslan and V.~Cohen.
\newblock Openwebtext corpus, 2019.

\bibitem{goyal2017making}
Y.~Goyal, T.~Khot, D.~Summers-Stay, D.~Batra, and D.~Parikh.
\newblock Making the v in vqa matter: Elevating the role of image understanding in visual question answering, 2017.

\bibitem{gu2021domain}
Y.~Gu, R.~Tinn, H.~Cheng, M.~Lucas, N.~Usuyama, X.~Liu, T.~Naumann, J.~Gao, and H.~Poon.
\newblock Domain-specific language model pretraining for biomedical natural language processing.
\newblock {\em ACM Transactions on Computing for Healthcare}, 3(1):1–23, Oct. 2021.

\bibitem{guo2024deepseekcoder}
D.~Guo, Q.~Zhu, D.~Yang, Z.~Xie, K.~Dong, W.~Zhang, G.~Chen, X.~Bi, Y.~Wu, Y.~K. Li, F.~Luo, Y.~Xiong, and W.~Liang.
\newblock Deepseek-coder: When the large language model meets programming -- the rise of code intelligence, 2024.

\bibitem{guo2023continuous}
Z.~Guo and Y.~Hua.
\newblock Continuous training and fine-tuning for domain-specific language models in medical question answering, 2023.

\bibitem{gupta2023continual}
K.~Gupta, B.~Thérien, A.~Ibrahim, M.~L. Richter, Q.~Anthony, E.~Belilovsky, I.~Rish, and T.~Lesort.
\newblock Continual pre-training of large language models: How to (re)warm your model?, 2023.

\bibitem{gurari2018vizwiz}
D.~Gurari, Q.~Li, A.~J. Stangl, A.~Guo, C.~Lin, K.~Grauman, J.~Luo, and J.~P. Bigham.
\newblock Vizwiz grand challenge: Answering visual questions from blind people, 2018.

\bibitem{gururangan2022demix}
S.~Gururangan, M.~Lewis, A.~Holtzman, N.~A. Smith, and L.~Zettlemoyer.
\newblock {DEM}ix layers: Disentangling domains for modular language modeling.
\newblock In M.~Carpuat, M.-C. de~Marneffe, and I.~V. Meza~Ruiz, editors, {\em Proceedings of the 2022 Conference of the North American Chapter of the Association for Computational Linguistics: Human Language Technologies}, pages 5557--5576, Seattle, United States, July 2022. Association for Computational Linguistics.

\bibitem{gururangan2020dont}
S.~Gururangan, A.~Marasovi{\'c}, S.~Swayamdipta, K.~Lo, I.~Beltagy, D.~Downey, and N.~A. Smith.
\newblock Don{'}t stop pretraining: Adapt language models to domains and tasks.
\newblock In D.~Jurafsky, J.~Chai, N.~Schluter, and J.~Tetreault, editors, {\em Proceedings of the 58th Annual Meeting of the Association for Computational Linguistics}, pages 8342--8360, Online, July 2020. Association for Computational Linguistics.

\bibitem{han2021econet}
R.~Han, X.~Ren, and N.~Peng.
\newblock {ECONET}: Effective continual pretraining of language models for event temporal reasoning.
\newblock In M.-F. Moens, X.~Huang, L.~Specia, and S.~W.-t. Yih, editors, {\em Proceedings of the 2021 Conference on Empirical Methods in Natural Language Processing}, pages 5367--5380, Online and Punta Cana, Dominican Republic, Nov. 2021. Association for Computational Linguistics.

\bibitem{hartvigsen2023aging}
T.~Hartvigsen, S.~Sankaranarayanan, H.~Palangi, Y.~Kim, and M.~Ghassemi.
\newblock Aging with grace: Lifelong model editing with discrete key-value adaptors.
\newblock In {\em Advances in Neural Information Processing Systems}, 2023.

\bibitem{hase2023does}
P.~Hase, M.~Bansal, B.~Kim, and A.~Ghandeharioun.
\newblock Does localization inform editing? surprising differences in causality-based localization vs.
\newblock {\em Knowledge Editing in Language Models}, 2023.

\bibitem{hase2021language}
P.~Hase, M.~Diab, A.~Celikyilmaz, X.~Li, Z.~Kozareva, V.~Stoyanov, M.~Bansal, and S.~Iyer.
\newblock Do language models have beliefs? methods for detecting, updating, and visualizing model beliefs.
\newblock {\em arXiv preprint arXiv:2111.13654}, 2021.

\bibitem{he2023continual}
J.~He, H.~Guo, M.~Tang, and J.~Wang.
\newblock Continual instruction tuning for large multimodal models, 2023.

\bibitem{he2016ups}
R.~He and J.~McAuley.
\newblock Ups and downs: Modeling the visual evolution of fashion trends with one-class collaborative filtering.
\newblock In {\em Proceedings of the 25th International Conference on World Wide Web}, WWW '16, page 507–517, Republic and Canton of Geneva, CHE, 2016. International World Wide Web Conferences Steering Committee.

\bibitem{he2021analyzing}
T.~He, J.~Liu, K.~Cho, M.~Ott, B.~Liu, J.~Glass, and F.~Peng.
\newblock Analyzing the forgetting problem in pretrain-finetuning of open-domain dialogue response models.
\newblock In P.~Merlo, J.~Tiedemann, and R.~Tsarfaty, editors, {\em Proceedings of the 16th Conference of the European Chapter of the Association for Computational Linguistics: Main Volume}, pages 1121--1133, Online, Apr. 2021. Association for Computational Linguistics.

\bibitem{he2024dont}
Y.~He, X.~Huang, M.~Tang, L.~Meng, X.~Li, W.~Lin, W.~Zhang, and Y.~Gao.
\newblock Don't half-listen: Capturing key-part information in continual instruction tuning, 2024.

\bibitem{hendrycks2023aligning}
D.~Hendrycks, C.~Burns, S.~Basart, A.~Critch, J.~Li, D.~Song, and J.~Steinhardt.
\newblock Aligning ai with shared human values, 2023.

\bibitem{hendryckstest2021}
D.~Hendrycks, C.~Burns, S.~Basart, A.~Zou, M.~Mazeika, D.~Song, and J.~Steinhardt.
\newblock Measuring massive multitask language understanding.
\newblock {\em Proceedings of the International Conference on Learning Representations (ICLR)}, 2021.

\bibitem{Wikireading}
D.~Hewlett, A.~Lacoste, L.~Jones, I.~Polosukhin, A.~Fandrianto, J.~Han, M.~Kelcey, and D.~Berthelot.
\newblock Wikireading: A novel large-scale language understanding task over wikipedia.
\newblock {\em arXiv preprint arXiv:1608.03542}, 2016.

\bibitem{hoffmann2022training}
J.~Hoffmann, S.~Borgeaud, A.~Mensch, E.~Buchatskaya, T.~Cai, E.~Rutherford, D.~d.~L. Casas, L.~A. Hendricks, J.~Welbl, A.~Clark, et~al.
\newblock Training compute-optimal large language models.
\newblock {\em arXiv preprint arXiv:2203.15556}, 2022.

\bibitem{hu2024wilke}
C.~Hu, P.~Cao, Y.~Chen, K.~Liu, and J.~Zhao.
\newblock Wilke: Wise-layer knowledge editor for lifelong knowledge editing, 2024.

\bibitem{hu2021lora}
E.~J. Hu, Y.~Shen, P.~Wallis, Z.~Allen-Zhu, Y.~Li, S.~Wang, L.~Wang, and W.~Chen.
\newblock Lora: Low-rank adaptation of large language models.
\newblock {\em arXiv preprint arXiv:2106.09685}, 2021.

\bibitem{hu2022lora}
E.~J. Hu, Y.~Shen, P.~Wallis, Z.~Allen-Zhu, Y.~Li, S.~Wang, L.~Wang, and W.~Chen.
\newblock Lo{RA}: Low-rank adaptation of large language models.
\newblock In {\em International Conference on Learning Representations}, 2022.

\bibitem{hu2023meetingbank}
Y.~Hu, T.~Ganter, H.~Deilamsalehy, F.~Dernoncourt, H.~Foroosh, and F.~Liu.
\newblock Meetingbank: A benchmark dataset for meeting summarization, 2023.

\bibitem{huang2024mitigating}
J.~Huang, L.~Cui, A.~Wang, C.~Yang, X.~Liao, L.~Song, J.~Yao, and J.~Su.
\newblock Mitigating catastrophic forgetting in large language models with self-synthesized rehearsal, 2024.

\bibitem{huang2019cosmos}
L.~Huang, R.~L. Bras, C.~Bhagavatula, and Y.~Choi.
\newblock Cosmos qa: Machine reading comprehension with contextual commonsense reasoning, 2019.

\bibitem{huang2023lawyer}
Q.~Huang, M.~Tao, Z.~An, C.~Zhang, C.~Jiang, Z.~Chen, Z.~Wu, and Y.~Feng.
\newblock Lawyer llama technical report.
\newblock {\em arXiv preprint arXiv:2305.15062}, 2023.

\bibitem{huang2023transformer}
Z.~Huang, Y.~Shen, X.~Zhang, J.~Zhou, W.~Rong, and Z.~Xiong.
\newblock Transformer-patcher: One mistake worth one neuron.
\newblock {\em arXiv preprint arXiv:2301.09785}, 2023.

\bibitem{hudson2019gqa}
D.~A. Hudson and C.~D. Manning.
\newblock Gqa: A new dataset for real-world visual reasoning and compositional question answering, 2019.

\bibitem{ibrahim2024simple}
A.~Ibrahim, B.~Th{\'e}rien, K.~Gupta, M.~L. Richter, Q.~Anthony, T.~Lesort, E.~Belilovsky, and I.~Rish.
\newblock Simple and scalable strategies to continually pre-train large language models.
\newblock {\em arXiv preprint arXiv:2403.08763}, 2024.

\bibitem{jang2022temporalwiki}
J.~Jang, S.~Ye, C.~Lee, S.~Yang, J.~Shin, J.~Han, G.~Kim, and M.~Seo.
\newblock Temporalwiki: A lifelong benchmark for training and evaluating ever-evolving language models.
\newblock 2022.

\bibitem{jang2022towards}
J.~Jang, S.~Ye, S.~Yang, J.~Shin, J.~Han, G.~Kim, S.~J. Choi, and M.~Seo.
\newblock Towards continual knowledge learning of language models.
\newblock In {\em ICLR}, 2022.

\bibitem{ji2024ai}
J.~Ji, T.~Qiu, B.~Chen, B.~Zhang, H.~Lou, K.~Wang, Y.~Duan, Z.~He, J.~Zhou, Z.~Zhang, F.~Zeng, K.~Y. Ng, J.~Dai, X.~Pan, A.~O'Gara, Y.~Lei, H.~Xu, B.~Tse, J.~Fu, S.~McAleer, Y.~Yang, Y.~Wang, S.-C. Zhu, Y.~Guo, and W.~Gao.
\newblock Ai alignment: A comprehensive survey, 2024.

\bibitem{jiang2024instructiontuned}
Z.~Jiang, Z.~Sun, W.~Shi, P.~Rodriguez, C.~Zhou, G.~Neubig, X.~V. Lin, W.~tau Yih, and S.~Iyer.
\newblock Instruction-tuned language models are better knowledge learners, 2024.

\bibitem{jin2024model}
X.~Jin and X.~Ren.
\newblock What will my model forget? forecasting forgotten examples in language model refinement, 2024.

\bibitem{jin2022lifelong}
X.~Jin, D.~Zhang, H.~Zhu, W.~Xiao, S.-W. Li, X.~Wei, A.~Arnold, and X.~Ren.
\newblock Lifelong pretraining: Continually adapting language models to emerging corpora.
\newblock In A.~Fan, S.~Ilic, T.~Wolf, and M.~Gall{\'e}, editors, {\em Proceedings of BigScience Episode {\#}5 -- Workshop on Challenges {\&} Perspectives in Creating Large Language Models}, pages 1--16, virtual+Dublin, May 2022. Association for Computational Linguistics.

\bibitem{kandel2000principles}
E.~R. Kandel, J.~H. Schwartz, T.~M. Jessell, S.~Siegelbaum, A.~J. Hudspeth, S.~Mack, et~al.
\newblock {\em Principles of neural science}, volume~4.
\newblock McGraw-hill New York, 2000.

\bibitem{kaplan2020scaling}
J.~Kaplan, S.~McCandlish, T.~Henighan, T.~B. Brown, B.~Chess, R.~Child, S.~Gray, A.~Radford, J.~Wu, and D.~Amodei.
\newblock Scaling laws for neural language models.
\newblock {\em arXiv preprint arXiv:2001.08361}, 2020.

\bibitem{kazemzadeh-etal-2014-referitgame}
S.~Kazemzadeh, V.~Ordonez, M.~Matten, and T.~Berg.
\newblock {R}efer{I}t{G}ame: Referring to objects in photographs of natural scenes.
\newblock In A.~Moschitti, B.~Pang, and W.~Daelemans, editors, {\em Proceedings of the 2014 Conference on Empirical Methods in Natural Language Processing ({EMNLP})}, pages 787--798, Doha, Qatar, Oct. 2014. Association for Computational Linguistics.

\bibitem{ke2022continual-train}
Z.~Ke, H.~Lin, Y.~Shao, H.~Xu, L.~Shu, and B.~Liu.
\newblock Continual training of language models for few-shot learning.
\newblock In Y.~Goldberg, Z.~Kozareva, and Y.~Zhang, editors, {\em Proceedings of the 2022 Conference on Empirical Methods in Natural Language Processing}, pages 10205--10216, Abu Dhabi, United Arab Emirates, Dec. 2022. Association for Computational Linguistics.

\bibitem{ke2023continual}
Z.~Ke and B.~Liu.
\newblock Continual learning of natural language processing tasks: A survey, 2023.

\bibitem{ke2021achieve}
Z.~Ke, B.~Liu, N.~Ma, H.~Xu, and S.~Lei.
\newblock Achieving forgetting prevention and knowledge transfer in continual learning.
\newblock In {\em NeurIPS}, 2021.

\bibitem{ke2022continual-pre}
Z.~Ke, Y.~Shao, H.~Lin, T.~Konishi, G.~Kim, and B.~Liu.
\newblock Continual pre-training of language models.
\newblock In {\em The Eleventh International Conference on Learning Representations}, 2022.

\bibitem{kew-etal-2023-20}
T.~Kew, M.~Kostrzewa, and S.~Ebling.
\newblock 20 minuten: A multi-task news summarisation dataset for {G}erman.
\newblock In H.~Ghorbel, M.~Sokhn, M.~Cieliebak, M.~H{\"u}rlimann, E.~de~Salis, and J.~Guerne, editors, {\em Proceedings of the 8th edition of the Swiss Text Analytics Conference}, pages 1--13, Neuchatel, Switzerland, June 2023. Association for Computational Linguistics.

\bibitem{khashabi-etal-2018-looking}
D.~Khashabi, S.~Chaturvedi, M.~Roth, S.~Upadhyay, and D.~Roth.
\newblock Looking beyond the surface: A challenge set for reading comprehension over multiple sentences.
\newblock In {\em Proceedings of the 2018 Conference of the North American Chapter of the Association for Computational Linguistics: Human Language Technologies, Volume 1 (Long Papers)}, pages 252--262, 2018.

\bibitem{khashabi-etal-2017-learning}
D.~Khashabi, T.~Khot, A.~Sabharwal, and D.~Roth.
\newblock Learning what is essential in questions.
\newblock In R.~Levy and L.~Specia, editors, {\em Proceedings of the 21st Conference on Computational Natural Language Learning ({C}o{NLL} 2017)}, pages 80--89, Vancouver, Canada, Aug. 2017. Association for Computational Linguistics.

\bibitem{khot2020qasc}
T.~Khot, P.~Clark, M.~Guerquin, P.~Jansen, and A.~Sabharwal.
\newblock Qasc: A dataset for question answering via sentence composition, 2020.

\bibitem{kim2022theoretical}
G.~Kim, C.~Xiao, T.~Konishi, Z.~Ke, and B.~Liu.
\newblock A theoretical study on solving continual learning.
\newblock In S.~Koyejo, S.~Mohamed, A.~Agarwal, D.~Belgrave, K.~Cho, and A.~Oh, editors, {\em Advances in Neural Information Processing Systems}, volume~35, pages 5065--5079. Curran Associates, Inc., 2022.

\bibitem{kirkpatrick2017overcoming}
J.~Kirkpatrick, R.~Pascanu, N.~Rabinowitz, J.~Veness, G.~Desjardins, A.~A. Rusu, K.~Milan, J.~Quan, T.~Ramalho, A.~Grabska-Barwinska, et~al.
\newblock Overcoming catastrophic forgetting in neural networks.
\newblock {\em Proceedings of the national academy of sciences}, 114(13):3521--3526, 2017.

\bibitem{nq}
T.~Kwiatkowski, J.~Palomaki, O.~Redfield, M.~Collins, A.~Parikh, C.~Alberti, D.~Epstein, I.~Polosukhin, J.~Devlin, K.~Lee, et~al.
\newblock Natural questions: a benchmark for question answering research.
\newblock {\em Transactions of the Association for Computational Linguistics}, 7:453--466, 2019.

\bibitem{lai2017race}
G.~Lai, Q.~Xie, H.~Liu, Y.~Yang, and E.~Hovy.
\newblock Race: Large-scale reading comprehension dataset from examinations.
\newblock {\em arXiv preprint arXiv:1704.04683}, 2017.

\bibitem{lazaridou2021mind}
A.~Lazaridou, A.~Kuncoro, E.~Gribovskaya, D.~Agrawal, A.~Liska, T.~Terzi, M.~Gimenez, C.~de~Masson~d'Autume, T.~Kocisky, S.~Ruder, et~al.
\newblock Mind the gap: Assessing temporal generalization in neural language models.
\newblock {\em Advances in Neural Information Processing Systems}, 34:29348--29363, 2021.

\bibitem{zsRE}
O.~Levy, M.~Seo, E.~Choi, and L.~Zettlemoyer.
\newblock Zero-shot relation extraction via reading comprehension.
\newblock {\em arXiv preprint arXiv:1706.04115}, 2017.

\bibitem{li2024examining}
C.-A. Li and H.-Y. Lee.
\newblock Examining forgetting in continual pre-training of aligned large language models, 2024.

\bibitem{li2022large}
D.~Li, A.~S. Rawat, M.~Zaheer, X.~Wang, M.~Lukasik, A.~Veit, F.~Yu, and S.~Kumar.
\newblock Large language models with controllable working memory.
\newblock {\em arXiv preprint arXiv:2211.05110}, 2022.

\bibitem{li2024blade}
H.~Li, Q.~Ai, J.~Chen, Q.~Dong, Z.~Wu, Y.~Liu, C.~Chen, and Q.~Tian.
\newblock Blade: Enhancing black-box large language models with small domain-specific models.
\newblock {\em arXiv preprint arXiv:2403.18365}, 2024.

\bibitem{li2023cfgpt}
J.~Li, Y.~Bian, G.~Wang, Y.~Lei, D.~Cheng, Z.~Ding, and C.~Jiang.
\newblock Cfgpt: Chinese financial assistant with large language model, 2023.

\bibitem{li2023blip2}
J.~Li, D.~Li, S.~Savarese, and S.~Hoi.
\newblock Blip-2: Bootstrapping language-image pre-training with frozen image encoders and large language models, 2023.

\bibitem{li2023continual}
L.~Li and X.~Qiu.
\newblock {CONTINUAL} {MODEL} {EVOLVEMENT} {WITH} {INNER}-{PRODUCT} {RESTRICTION}, 2023.

\bibitem{li2017learning}
Z.~Li and D.~Hoiem.
\newblock Learning without forgetting.
\newblock {\em IEEE transactions on pattern analysis and machine intelligence}, 40(12):2935--2947, 2017.

\bibitem{lin2022continual}
B.~Y. Lin, S.~Wang, X.~Lin, R.~Jia, L.~Xiao, X.~Ren, and S.~Yih.
\newblock On continual model refinement in out-of-distribution data streams.
\newblock In S.~Muresan, P.~Nakov, and A.~Villavicencio, editors, {\em Proceedings of the 60th Annual Meeting of the Association for Computational Linguistics (Volume 1: Long Papers)}, pages 3128--3139, Dublin, Ireland, May 2022. Association for Computational Linguistics.

\bibitem{lin2004rouge}
C.-Y. Lin.
\newblock Rouge: A package for automatic evaluation of summaries.
\newblock In {\em Text summarization branches out}, pages 74--81, 2004.

\bibitem{lin2019reasoning}
K.~Lin, O.~Tafjord, P.~Clark, and M.~Gardner.
\newblock Reasoning over paragraph effects in situations, 2019.

\bibitem{lin2024mitigating}
Y.~Lin, H.~Lin, W.~Xiong, S.~Diao, J.~Liu, J.~Zhang, R.~Pan, H.~Wang, W.~Hu, H.~Zhang, H.~Dong, R.~Pi, H.~Zhao, N.~Jiang, H.~Ji, Y.~Yao, and T.~Zhang.
\newblock Mitigating the alignment tax of rlhf, 2024.

\bibitem{lin2023geogalactica}
Z.~Lin, C.~Deng, L.~Zhou, T.~Zhang, Y.~Xu, Y.~Xu, Z.~He, Y.~Shi, B.~Dai, Y.~Song, B.~Zeng, Q.~Chen, T.~Shi, T.~Huang, Y.~Xu, S.~Wang, L.~Fu, W.~Zhang, J.~He, C.~Ma, Y.~Zhu, X.~Wang, and C.~Zhou.
\newblock Geogalactica: A scientific large language model in geoscience, 2023.

\bibitem{lin2024rho}
Z.~Lin, Z.~Gou, Y.~Gong, X.~Liu, Y.~Shen, R.~Xu, C.~Lin, Y.~Yang, J.~Jiao, N.~Duan, et~al.
\newblock Rho-1: Not all tokens are what you need.
\newblock {\em arXiv preprint arXiv:2404.07965}, 2024.

\bibitem{liu2023visual}
H.~Liu, C.~Li, Q.~Wu, and Y.~J. Lee.
\newblock Visual instruction tuning, 2023.

\bibitem{liu2022few}
H.~Liu, D.~Tam, M.~Muqeeth, J.~Mohta, T.~Huang, M.~Bansal, and C.~A. Raffel.
\newblock Few-shot parameter-efficient fine-tuning is better and cheaper than in-context learning.
\newblock {\em Advances in Neural Information Processing Systems}, 35:1950--1965, 2022.

\bibitem{liu2019roberta}
Y.~Liu, M.~Ott, N.~Goyal, J.~Du, M.~Joshi, D.~Chen, O.~Levy, M.~Lewis, L.~Zettlemoyer, and V.~Stoyanov.
\newblock Roberta: A robustly optimized bert pretraining approach.
\newblock {\em arXiv preprint arXiv:1907.11692}, 2019.

\bibitem{lo2020s2orc}
K.~Lo, L.~L. Wang, M.~Neumann, R.~Kinney, and D.~Weld.
\newblock {S}2{ORC}: The semantic scholar open research corpus.
\newblock In D.~Jurafsky, J.~Chai, N.~Schluter, and J.~Tetreault, editors, {\em Proceedings of the 58th Annual Meeting of the Association for Computational Linguistics}, pages 4969--4983, Online, July 2020. Association for Computational Linguistics.

\bibitem{lomonaco2020rehearsalfree}
V.~Lomonaco, D.~Maltoni, and L.~Pellegrini.
\newblock Rehearsal-free continual learning over small non-i.i.d. batches, 2020.

\bibitem{lopez2017gradient}
D.~Lopez-Paz and M.~Ranzato.
\newblock Gradient episodic memory for continual learning.
\newblock {\em Advances in neural information processing systems}, 30, 2017.

\bibitem{loureiro2022timelms}
D.~Loureiro, F.~Barbieri, L.~Neves, L.~Espinosa~Anke, and J.~Camacho-collados.
\newblock {T}ime{LM}s: Diachronic language models from {T}witter.
\newblock In V.~Basile, Z.~Kozareva, and S.~Stajner, editors, {\em Proceedings of the 60th Annual Meeting of the Association for Computational Linguistics: System Demonstrations}, pages 251--260, Dublin, Ireland, May 2022. Association for Computational Linguistics.

\bibitem{Lu2023BBTFin}
D.~Lu, H.~Wu, J.~Liang, Y.~Xu, Q.~He, Y.~Geng, M.~Han, Y.~Xin, and Y.~Xiao.
\newblock Bbt-fin: Comprehensive construction of chinese financial domain pre-trained language model, corpus and benchmark.
\newblock {\em CoRR}, abs/2302.09432, 2023.

\bibitem{lu2023ibcl}
P.~Lu, M.~Caprio, E.~Eaton, and I.~Lee.
\newblock Ibcl: Zero-shot model generation for task trade-offs in continual learning, 2023.

\bibitem{lu2022learn}
P.~Lu, S.~Mishra, T.~Xia, L.~Qiu, K.-W. Chang, S.-C. Zhu, O.~Tafjord, P.~Clark, and A.~Kalyan.
\newblock Learn to explain: Multimodal reasoning via thought chains for science question answering, 2022.

\bibitem{lu2021codexglue}
S.~Lu, D.~Guo, S.~Ren, J.~Huang, A.~Svyatkovskiy, A.~Blanco, C.~Clement, D.~Drain, D.~Jiang, D.~Tang, G.~Li, L.~Zhou, L.~Shou, L.~Zhou, M.~Tufano, M.~Gong, M.~Zhou, N.~Duan, N.~Sundaresan, S.~K. Deng, S.~Fu, and S.~Liu.
\newblock Codexglue: A machine learning benchmark dataset for code understanding and generation, 2021.

\bibitem{luo2023wizardmath}
H.~Luo, Q.~Sun, C.~Xu, P.~Zhao, J.~Lou, C.~Tao, X.~Geng, Q.~Lin, S.~Chen, and D.~Zhang.
\newblock Wizardmath: Empowering mathematical reasoning for large language models via reinforced evol-instruct.
\newblock {\em arXiv preprint arXiv:2308.09583}, 2023.

\bibitem{luo2022biogpt}
R.~Luo, L.~Sun, Y.~Xia, T.~Qin, S.~Zhang, H.~Poon, and T.-Y. Liu.
\newblock Biogpt: generative pre-trained transformer for biomedical text generation and mining.
\newblock {\em Briefings in Bioinformatics}, 23(6), Sept. 2022.

\bibitem{luo2023investigating}
Y.~Luo, Z.~Yang, X.~Bai, F.~Meng, J.~Zhou, and Y.~Zhang.
\newblock Investigating forgetting in pre-trained representations through continual learning, 2023.

\bibitem{luo2023empirical}
Y.~Luo, Z.~Yang, F.~Meng, Y.~Li, J.~Zhou, and Y.~Zhang.
\newblock An empirical study of catastrophic forgetting in large language models during continual fine-tuning, 2023.

\bibitem{luo2023biomedgpt}
Y.~Luo, J.~Zhang, S.~Fan, K.~Yang, Y.~Wu, M.~Qiao, and Z.~Nie.
\newblock Biomedgpt: Open multimodal generative pre-trained transformer for biomedicine.
\newblock {\em arXiv preprint arXiv:2308.09442}, 2023.

\bibitem{luo2023wizardcoder}
Z.~Luo, C.~Xu, P.~Zhao, Q.~Sun, X.~Geng, W.~Hu, C.~Tao, J.~Ma, Q.~Lin, and D.~Jiang.
\newblock Wizardcoder: Empowering code large language models with evol-instruct, 2023.

\bibitem{ma2023ecomgptct}
S.~Ma, S.~Huang, S.~Huang, X.~Wang, Y.~Li, H.-T. Zheng, P.~Xie, F.~Huang, and Y.~Jiang.
\newblock Ecomgpt-ct: Continual pre-training of e-commerce large language models with semi-structured data, 2023.

\bibitem{maas2011learning}
A.~Maas, R.~E. Daly, P.~T. Pham, D.~Huang, A.~Y. Ng, and C.~Potts.
\newblock Learning word vectors for sentiment analysis.
\newblock In {\em Proceedings of the 49th annual meeting of the association for computational linguistics: Human language technologies}, pages 142--150, 2011.

\bibitem{mai2022online}
Z.~Mai, R.~Li, J.~Jeong, D.~Quispe, H.~Kim, and S.~Sanner.
\newblock Online continual learning in image classification: An empirical survey.
\newblock {\em Neurocomputing}, 469:28--51, 2022.

\bibitem{mao2016generation}
J.~Mao, J.~Huang, A.~Toshev, O.~Camburu, A.~Yuille, and K.~Murphy.
\newblock Generation and comprehension of unambiguous object descriptions, 2016.

\bibitem{mazzia2023survey}
V.~Mazzia, A.~Pedrani, A.~Caciolai, K.~Rottmann, and D.~Bernardi.
\newblock A survey on knowledge editing of neural networks.
\newblock {\em arXiv preprint arXiv:2310.19704}, 2023.

\bibitem{mccaffary2021towards}
D.~McCaffary.
\newblock Towards continual task learning in artificial neural networks: current approaches and insights from neuroscience.
\newblock {\em arXiv preprint arXiv:2112.14146}, 2021.

\bibitem{mcclelland1995there}
J.~L. McClelland, B.~L. McNaughton, and R.~C. O'Reilly.
\newblock Why there are complementary learning systems in the hippocampus and neocortex: insights from the successes and failures of connectionist models of learning and memory.
\newblock {\em Psychological review}, 102(3):419, 1995.

\bibitem{mccloskey1989catastrophic}
M.~McCloskey and N.~J. Cohen.
\newblock Catastrophic interference in connectionist networks: The sequential learning problem.
\newblock volume~24 of {\em Psychology of Learning and Motivation}, pages 109--165. Academic Press, 1989.

\bibitem{mehta2023empirical}
S.~V. Mehta, D.~Patil, S.~Chandar, and E.~Strubell.
\newblock An empirical investigation of the role of pre-training in lifelong learning.
\newblock {\em Journal of Machine Learning Research}, 24(214):1--50, 2023.

\bibitem{meng2022locating}
K.~Meng, D.~Bau, A.~Andonian, and Y.~Belinkov.
\newblock Locating and editing factual associations in gpt.
\newblock {\em Advances in Neural Information Processing Systems}, 35:17359--17372, 2022.

\bibitem{meng2022mass}
K.~Meng, A.~S. Sharma, A.~Andonian, Y.~Belinkov, and D.~Bau.
\newblock Mass-editing memory in a transformer.
\newblock {\em arXiv preprint arXiv:2210.07229}, 2022.

\bibitem{min2022rethinking}
S.~Min, X.~Lyu, A.~Holtzman, M.~Artetxe, M.~Lewis, H.~Hajishirzi, and L.~Zettlemoyer.
\newblock Rethinking the role of demonstrations: What makes in-context learning work?
\newblock {\em arXiv preprint arXiv:2202.12837}, 2022.

\bibitem{mishraICDAR19}
A.~Mishra, S.~Shekhar, A.~K. Singh, and A.~Chakraborty.
\newblock Ocr-vqa: Visual question answering by reading text in images.
\newblock In {\em ICDAR}, 2019.

\bibitem{mishra2021natural}
S.~Mishra, D.~Khashabi, C.~Baral, and H.~Hajishirzi.
\newblock Natural instructions: Benchmarking generalization to new tasks from natural language instructions.
\newblock {\em arXiv preprint arXiv:2104.08773}, 2021.

\bibitem{mishra2022numglue}
S.~Mishra, A.~Mitra, N.~Varshney, B.~Sachdeva, P.~Clark, C.~Baral, and A.~Kalyan.
\newblock Numglue: A suite of fundamental yet challenging mathematical reasoning tasks, 2022.

\bibitem{fast_edit}
E.~Mitchell, C.~Lin, A.~Bosselut, C.~Finn, and C.~D. Manning.
\newblock Fast model editing at scale.
\newblock {\em arXiv preprint arXiv:2110.11309}, 2021.

\bibitem{mitchell2022memory}
E.~Mitchell, C.~Lin, A.~Bosselut, C.~D. Manning, and C.~Finn.
\newblock Memory-based model editing at scale.
\newblock In {\em International Conference on Machine Learning}, pages 15817--15831. PMLR, 2022.

\bibitem{mok2023large}
J.~Mok, J.~Do, S.~Lee, T.~Taghavi, S.~Yu, and S.~Yoon.
\newblock Large-scale lifelong learning of in-context instructions and how to tackle it.
\newblock In A.~Rogers, J.~Boyd-Graber, and N.~Okazaki, editors, {\em Proceedings of the 61st Annual Meeting of the Association for Computational Linguistics (Volume 1: Long Papers)}, pages 12573--12589, Toronto, Canada, July 2023. Association for Computational Linguistics.

\bibitem{moradidakhel2023github}
A.~{Moradi Dakhel}, V.~Majdinasab, A.~Nikanjam, F.~Khomh, M.~C. Desmarais, and Z.~M.~J. Jiang.
\newblock Github copilot ai pair programmer: Asset or liability?
\newblock {\em Journal of Systems and Software}, 203:111734, 2023.

\bibitem{nakamura2024aurora}
T.~Nakamura, M.~Mishra, S.~Tedeschi, Y.~Chai, J.~T. Stillerman, F.~Friedrich, P.~Yadav, T.~Laud, V.~M. Chien, T.~Y. Zhuo, et~al.
\newblock Aurora-m: The first open source multilingual language model red-teamed according to the us executive order.
\newblock {\em arXiv preprint arXiv:2404.00399}, 2024.

\bibitem{Nguyen2023AstroLLaMA}
T.~D. Nguyen, Y.~Ting, I.~Ciuca, C.~O'Neill, Z.~Sun, M.~Jablonska, S.~Kruk, E.~Perkowski, J.~W. Miller, J.~Li, J.~Peek, K.~Iyer, T.~R{\'{o}}zanski, P.~Khetarpal, S.~Zaman, D.~Brodrick, S.~J.~R. M{\'{e}}ndez, T.~Bui, A.~Goodman, A.~Accomazzi, J.~P. Naiman, J.~Cranney, K.~Schawinski, and UniverseTBD.
\newblock Astrollama: Towards specialized foundation models in astronomy.
\newblock {\em CoRR}, abs/2309.06126, 2023.

\bibitem{ni2019justifying}
J.~Ni, J.~Li, and J.~McAuley.
\newblock Justifying recommendations using distantly-labeled reviews and fine-grained aspects.
\newblock In K.~Inui, J.~Jiang, V.~Ng, and X.~Wan, editors, {\em Proceedings of the 2019 Conference on Empirical Methods in Natural Language Processing and the 9th International Joint Conference on Natural Language Processing (EMNLP-IJCNLP)}, pages 188--197, Hong Kong, China, Nov. 2019. Association for Computational Linguistics.

\bibitem{ni2021revisiting}
Z.~Ni, H.~Shi, S.~Tang, L.~Wei, Q.~Tian, and Y.~Zhuang.
\newblock Revisiting catastrophic forgetting in class incremental learning.
\newblock {\em arXiv preprint arXiv:2107.12308}, 2021.

\bibitem{ni2023continual}
Z.~Ni, L.~Wei, S.~Tang, Y.~Zhuang, and Q.~Tian.
\newblock Continual vision-language representation learning with off-diagonal information.
\newblock In {\em Proceedings of the 40th International Conference on Machine Learning}, pages 26129--26149, 2023.

\bibitem{nijkamp2022codegen}
E.~Nijkamp, B.~Pang, H.~Hayashi, L.~Tu, H.~Wang, Y.~Zhou, S.~Savarese, and C.~Xiong.
\newblock Codegen: An open large language model for code with multi-turn program synthesis.
\newblock {\em ICLR}, 2023.

\bibitem{achiam2022chatgpt}
OpenAI.
\newblock Introducing chatgpt. [online]. available: \url{https://openai.com/blog/chatgpt}.
\newblock 2022.

\bibitem{ouyang2022rlhf}
L.~Ouyang, J.~Wu, X.~Jiang, D.~Almeida, C.~L. Wainwright, P.~Mishkin, C.~Zhang, S.~Agarwal, K.~Slama, A.~Ray, J.~Schulman, J.~Hilton, F.~Kelton, L.~Miller, M.~Simens, A.~Askell, P.~Welinder, P.~Christiano, J.~Leike, and R.~Lowe.
\newblock Training language models to follow instructions with human feedback, 2022.

\bibitem{pallier2003brain}
C.~Pallier, S.~Dehaene, J.-B. Poline, D.~LeBihan, A.-M. Argenti, E.~Dupoux, and J.~Mehler.
\newblock Brain imaging of language plasticity in adopted adults: Can a second language replace the first?
\newblock {\em Cerebral cortex}, 13(2):155--161, 2003.

\bibitem{papineni2002bleu}
K.~Papineni, S.~Roukos, T.~Ward, and W.-J. Zhu.
\newblock Bleu: a method for automatic evaluation of machine translation.
\newblock In {\em Proceedings of the 40th annual meeting of the Association for Computational Linguistics}, pages 311--318, 2002.

\bibitem{paul2024ircoder}
I.~Paul, J.~Luo, G.~Glavaš, and I.~Gurevych.
\newblock Ircoder: Intermediate representations make language models robust multilingual code generators, 2024.

\bibitem{pentina2016theoretical}
A.~Pentina.
\newblock {\em Theoretical foundations of multi-task lifelong learning}.
\newblock PhD thesis, 2016.

\bibitem{petroni2019language}
F.~Petroni, T.~Rockt{\"a}schel, S.~Riedel, P.~Lewis, A.~Bakhtin, Y.~Wu, and A.~Miller.
\newblock Language models as knowledge bases?
\newblock In K.~Inui, J.~Jiang, V.~Ng, and X.~Wan, editors, {\em Proceedings of the 2019 Conference on Empirical Methods in Natural Language Processing and the 9th International Joint Conference on Natural Language Processing (EMNLP-IJCNLP)}, pages 2463--2473, Hong Kong, China, Nov. 2019. Association for Computational Linguistics.

\bibitem{prabhu2023computationally}
A.~Prabhu, H.~A. Al~Kader~Hammoud, P.~K. Dokania, P.~H. Torr, S.-N. Lim, B.~Ghanem, and A.~Bibi.
\newblock Computationally budgeted continual learning: What does matter?
\newblock In {\em Proceedings of the IEEE/CVF Conference on Computer Vision and Pattern Recognition}, pages 3698--3707, 2023.

\bibitem{prabhu2023online}
A.~Prabhu, Z.~Cai, P.~Dokania, P.~Torr, V.~Koltun, and O.~Sener.
\newblock Online continual learning without the storage constraint, 2023.

\bibitem{qin2021lfpt5}
C.~Qin and S.~Joty.
\newblock Lfpt5: A unified framework for lifelong few-shot language learning based on prompt tuning of t5.
\newblock In {\em International Conference on Learning Representations}, 2021.

\bibitem{qin2023recyclable}
Y.~Qin, C.~Qian, X.~Han, Y.~Lin, H.~Wang, R.~Xie, Z.~Liu, M.~Sun, and J.~Zhou.
\newblock Recyclable tuning for continual pre-training.
\newblock {\em arXiv preprint arXiv:2305.08702}, 2023.

\bibitem{qin2022elle}
Y.~Qin, J.~Zhang, Y.~Lin, Z.~Liu, P.~Li, M.~Sun, and J.~Zhou.
\newblock {ELLE}: Efficient lifelong pre-training for emerging data.
\newblock In S.~Muresan, P.~Nakov, and A.~Villavicencio, editors, {\em Findings of the Association for Computational Linguistics: ACL 2022}, pages 2789--2810, Dublin, Ireland, May 2022. Association for Computational Linguistics.

\bibitem{radford2021learning}
A.~Radford, J.~W. Kim, C.~Hallacy, A.~Ramesh, G.~Goh, S.~Agarwal, G.~Sastry, A.~Askell, P.~Mishkin, J.~Clark, et~al.
\newblock Learning transferable visual models from natural language supervision.
\newblock In {\em International conference on machine learning}, pages 8748--8763. PMLR, 2021.

\bibitem{radford2019language}
A.~Radford, J.~Wu, R.~Child, D.~Luan, D.~Amodei, I.~Sutskever, et~al.
\newblock Language models are unsupervised multitask learners.
\newblock {\em OpenAI blog}, 1(8):9, 2019.

\bibitem{rafailov2024dpo}
R.~Rafailov, A.~Sharma, E.~Mitchell, C.~D. Manning, S.~Ermon, and C.~Finn.
\newblock Direct preference optimization: Your language model is secretly a reward model.
\newblock {\em Advances in Neural Information Processing Systems}, 36, 2024.

\bibitem{rafailov2024direct}
R.~Rafailov, A.~Sharma, E.~Mitchell, C.~D. Manning, S.~Ermon, and C.~Finn.
\newblock Direct preference optimization: Your language model is secretly a reward model.
\newblock {\em Advances in Neural Information Processing Systems}, 36, 2024.

\bibitem{raffel2020exploring}
C.~Raffel, N.~Shazeer, A.~Roberts, K.~Lee, S.~Narang, M.~Matena, Y.~Zhou, W.~Li, and P.~J. Liu.
\newblock Exploring the limits of transfer learning with a unified text-to-text transformer.
\newblock {\em Journal of machine learning research}, 21(140):1--67, 2020.

\bibitem{rajpurkar2018know}
P.~Rajpurkar, R.~Jia, and P.~Liang.
\newblock Know what you don't know: Unanswerable questions for squad.
\newblock {\em arXiv preprint arXiv:1806.03822}, 2018.

\bibitem{ramesh2021model}
R.~Ramesh and P.~Chaudhari.
\newblock Model zoo: A growing" brain" that learns continually.
\newblock {\em arXiv preprint arXiv:2106.03027}, 2021.

\bibitem{ramsauer2021hopfield}
H.~Ramsauer, B.~Schäfl, J.~Lehner, P.~Seidl, M.~Widrich, T.~Adler, L.~Gruber, M.~Holzleitner, M.~Pavlović, G.~K. Sandve, V.~Greiff, D.~Kreil, M.~Kopp, G.~Klambauer, J.~Brandstetter, and S.~Hochreiter.
\newblock Hopfield networks is all you need, 2021.

\bibitem{rebuffi2017icarl}
S.-A. Rebuffi, A.~Kolesnikov, G.~Sperl, and C.~H. Lampert.
\newblock icarl: Incremental classifier and representation learning.
\newblock In {\em Proceedings of the IEEE conference on Computer Vision and Pattern Recognition}, pages 2001--2010, 2017.

\bibitem{reid2024gemini}
M.~Reid, N.~Savinov, D.~Teplyashin, D.~Lepikhin, T.~Lillicrap, J.-b. Alayrac, R.~Soricut, A.~Lazaridou, O.~Firat, J.~Schrittwieser, et~al.
\newblock Gemini 1.5: Unlocking multimodal understanding across millions of tokens of context.
\newblock {\em arXiv preprint arXiv:2403.05530}, 2024.

\bibitem{riemer2018learning}
M.~Riemer, I.~Cases, R.~Ajemian, M.~Liu, I.~Rish, Y.~Tu, and G.~Tesauro.
\newblock Learning to learn without forgetting by maximizing transfer and minimizing interference.
\newblock {\em arXiv preprint arXiv:1810.11910}, 2018.

\bibitem{ritter2018online}
H.~Ritter, A.~Botev, and D.~Barber.
\newblock Online structured laplace approximations for overcoming catastrophic forgetting.
\newblock {\em Advances in Neural Information Processing Systems}, 31, 2018.

\bibitem{rongali2021continual}
S.~Rongali, A.~Jagannatha, B.~P.~S. Rawat, and H.~Yu.
\newblock Continual domain-tuning for pretrained language models, 2021.

\bibitem{rosin2022time}
G.~D. Rosin, I.~Guy, and K.~Radinsky.
\newblock Time masking for temporal language models.
\newblock In {\em Proceedings of the Fifteenth ACM International Conference on Web Search and Data Mining}, WSDM '22, page 833–841, New York, NY, USA, 2022. Association for Computing Machinery.

\bibitem{rozière2024code}
B.~Rozière, J.~Gehring, F.~Gloeckle, S.~Sootla, I.~Gat, X.~E. Tan, Y.~Adi, J.~Liu, R.~Sauvestre, T.~Remez, J.~Rapin, A.~Kozhevnikov, I.~Evtimov, J.~Bitton, M.~Bhatt, C.~C. Ferrer, A.~Grattafiori, W.~Xiong, A.~Défossez, J.~Copet, F.~Azhar, H.~Touvron, L.~Martin, N.~Usunier, T.~Scialom, and G.~Synnaeve.
\newblock Code llama: Open foundation models for code, 2024.

\bibitem{Rubungo2023LLM-Prop}
A.~N. Rubungo, C.~Arnold, B.~P. Rand, and A.~B. Dieng.
\newblock Llm-prop: Predicting physical and electronic properties of crystalline solids from their text descriptions.
\newblock {\em CoRR}, abs/2310.14029, 2023.

\bibitem{rusu2016progressive}
A.~A. Rusu, N.~C. Rabinowitz, G.~Desjardins, H.~Soyer, J.~Kirkpatrick, K.~Kavukcuoglu, R.~Pascanu, and R.~Hadsell.
\newblock Progressive neural networks.
\newblock {\em arXiv preprint arXiv:1606.04671}, 2016.

\bibitem{sakaguchi2019winogrande}
K.~Sakaguchi, R.~L. Bras, C.~Bhagavatula, and Y.~Choi.
\newblock Winogrande: An adversarial winograd schema challenge at scale, 2019.

\bibitem{sanh2022multitask}
V.~Sanh, A.~Webson, C.~Raffel, S.~H. Bach, L.~Sutawika, Z.~Alyafeai, A.~Chaffin, A.~Stiegler, T.~L. Scao, A.~Raja, M.~Dey, M.~S. Bari, C.~Xu, U.~Thakker, S.~S. Sharma, E.~Szczechla, T.~Kim, G.~Chhablani, N.~Nayak, D.~Datta, J.~Chang, M.~T.-J. Jiang, H.~Wang, M.~Manica, S.~Shen, Z.~X. Yong, H.~Pandey, R.~Bawden, T.~Wang, T.~Neeraj, J.~Rozen, A.~Sharma, A.~Santilli, T.~Fevry, J.~A. Fries, R.~Teehan, T.~Bers, S.~Biderman, L.~Gao, T.~Wolf, and A.~M. Rush.
\newblock Multitask prompted training enables zero-shot task generalization, 2022.

\bibitem{sarfraz2023error}
F.~Sarfraz, E.~Arani, and B.~Zonooz.
\newblock Error sensitivity modulation based experience replay: Mitigating abrupt representation drift in continual learning.
\newblock {\em arXiv preprint arXiv:2302.11344}, 2023.

\bibitem{schulman2017proximal}
J.~Schulman, F.~Wolski, P.~Dhariwal, A.~Radford, and O.~Klimov.
\newblock Proximal policy optimization algorithms, 2017.

\bibitem{vitaminC}
T.~Schuster, A.~Fisch, and R.~Barzilay.
\newblock Get your vitamin c! robust fact verification with contrastive evidence.
\newblock {\em arXiv preprint arXiv:2103.08541}, 2021.

\bibitem{schwarz2018progress}
J.~Schwarz, W.~Czarnecki, J.~Luketina, A.~Grabska-Barwinska, Y.~W. Teh, R.~Pascanu, and R.~Hadsell.
\newblock Progress \& compress: A scalable framework for continual learning.
\newblock In {\em International conference on machine learning}, pages 4528--4537. PMLR, 2018.

\bibitem{scialom2022fine}
T.~Scialom, T.~Chakrabarty, and S.~Muresan.
\newblock Fine-tuned language models are continual learners.
\newblock In Y.~Goldberg, Z.~Kozareva, and Y.~Zhang, editors, {\em Proceedings of the 2022 Conference on Empirical Methods in Natural Language Processing}, pages 6107--6122, Abu Dhabi, United Arab Emirates, Dec. 2022. Association for Computational Linguistics.

\bibitem{shah2023trillion}
A.~Shah, S.~Paturi, and S.~Chava.
\newblock Trillion dollar words: A new financial dataset, task \& market analysis, 2023.

\bibitem{shao2023class}
Y.~Shao, Y.~Guo, D.~Zhao, and B.~Liu.
\newblock Class-incremental learning based on label generation.
\newblock In A.~Rogers, J.~Boyd-Graber, and N.~Okazaki, editors, {\em Proceedings of the 61st Annual Meeting of the Association for Computational Linguistics (Volume 2: Short Papers)}, pages 1263--1276, Toronto, Canada, July 2023. Association for Computational Linguistics.

\bibitem{shazeer2017outrageously}
N.~Shazeer, A.~Mirhoseini, K.~Maziarz, A.~Davis, Q.~Le, G.~Hinton, and J.~Dean.
\newblock Outrageously large neural networks: The sparsely-gated mixture-of-experts layer.
\newblock {\em arXiv preprint arXiv:1701.06538}, 2017.

\bibitem{shen2024tag}
J.~Shen, N.~Tenenholtz, J.~B. Hall, D.~Alvarez-Melis, and N.~Fusi.
\newblock Tag-llm: Repurposing general-purpose llms for specialized domains.
\newblock {\em arXiv preprint arXiv:2402.05140}, 2024.

\bibitem{shi2024unified}
H.~Shi and H.~Wang.
\newblock A unified approach to domain incremental learning with memory: Theory and algorithm.
\newblock {\em Advances in Neural Information Processing Systems}, 36, 2024.

\bibitem{singh2019vqa}
A.~Singh, V.~Natarajan, M.~Shah, Y.~Jiang, X.~Chen, D.~Batra, D.~Parikh, and M.~Rohrbach.
\newblock Towards vqa models that can read, 2019.

\bibitem{sinitsin2020editable}
A.~Sinitsin, V.~Plokhotnyuk, D.~Pyrkin, S.~Popov, and A.~Babenko.
\newblock Editable neural networks.
\newblock {\em arXiv preprint arXiv:2004.00345}, 2020.

\bibitem{soldaini2024dolma}
L.~Soldaini, R.~Kinney, A.~Bhagia, D.~Schwenk, D.~Atkinson, R.~Authur, B.~Bogin, K.~Chandu, J.~Dumas, Y.~Elazar, V.~Hofmann, A.~H. Jha, S.~Kumar, L.~Lucy, X.~Lyu, N.~Lambert, I.~Magnusson, J.~Morrison, N.~Muennighoff, A.~Naik, C.~Nam, M.~E. Peters, A.~Ravichander, K.~Richardson, Z.~Shen, E.~Strubell, N.~Subramani, O.~Tafjord, P.~Walsh, L.~Zettlemoyer, N.~A. Smith, H.~Hajishirzi, I.~Beltagy, D.~Groeneveld, J.~Dodge, and K.~Lo.
\newblock Dolma: an open corpus of three trillion tokens for language model pretraining research, 2024.

\bibitem{song2023conpet}
C.~Song, X.~Han, Z.~Zeng, K.~Li, C.~Chen, Z.~Liu, M.~Sun, and T.~Yang.
\newblock Conpet: Continual parameter-efficient tuning for large language models, 2023.

\bibitem{song2024code}
D.~Song, H.~Guo, Y.~Zhou, S.~Xing, Y.~Wang, Z.~Song, W.~Zhang, Q.~Guo, H.~Yan, X.~Qiu, and D.~Lin.
\newblock Code needs comments: Enhancing code llms with comment augmentation, 2024.

\bibitem{su2023efficient}
Z.~Su, J.~Li, Z.~Zhang, Z.~Zhou, and M.~Zhang.
\newblock Efficient continue training of temporal language model with structural information.
\newblock In H.~Bouamor, J.~Pino, and K.~Bali, editors, {\em Findings of the Association for Computational Linguistics: EMNLP 2023}, pages 6315--6329, Singapore, Dec. 2023. Association for Computational Linguistics.

\bibitem{sun2024survey}
Q.~Sun, Z.~Chen, F.~Xu, K.~Cheng, C.~Ma, Z.~Yin, J.~Wang, C.~Han, R.~Zhu, S.~Yuan, Q.~Guo, X.~Qiu, P.~Yin, X.~Li, F.~Yuan, L.~Kong, X.~Li, and Z.~Wu.
\newblock A survey of neural code intelligence: Paradigms, advances and beyond, 2024.

\bibitem{sun2020ernie}
Y.~Sun, S.~Wang, Y.~Li, S.~Feng, H.~Tian, H.~Wu, and H.~Wang.
\newblock Ernie 2.0: A continual pre-training framework for language understanding.
\newblock {\em Proceedings of the AAAI Conference on Artificial Intelligence}, 34(05):8968--8975, Apr. 2020.

\bibitem{takahashi2024pretraining}
K.~Takahashi, T.~Omi, K.~Arima, and T.~Ishigaki.
\newblock Pretraining and updating language-and domain-specific large language model: A case study in japanese business domain.
\newblock {\em arXiv preprint arXiv:2404.08262}, 2024.

\bibitem{tao2022can}
M.~Tao, Y.~Feng, and D.~Zhao.
\newblock Can bert refrain from forgetting on sequential tasks? a probing study.
\newblock In {\em The Eleventh International Conference on Learning Representations}, 2022.

\bibitem{deepseekai2024deepseek}
D.-A. Team.
\newblock Deepseek llm: Scaling open-source language models with longtermism, 2024.

\bibitem{team2023gemini}
G.~Team, R.~Anil, S.~Borgeaud, Y.~Wu, J.-B. Alayrac, J.~Yu, R.~Soricut, J.~Schalkwyk, A.~M. Dai, A.~Hauth, et~al.
\newblock Gemini: a family of highly capable multimodal models.
\newblock {\em arXiv preprint arXiv:2312.11805}, 2023.

\bibitem{li2023starcoder}
S.~Team.
\newblock Starcoder: may the source be with you!, 2023.

\bibitem{lozhkov2024starcoder}
S.~Team.
\newblock Starcoder 2 and the stack v2: The next generation, 2024.

\bibitem{fever}
J.~Thorne, A.~Vlachos, C.~Christodoulopoulos, and A.~Mittal.
\newblock Fever: a large-scale dataset for fact extraction and verification.
\newblock {\em arXiv preprint arXiv:1803.05355}, 2018.

\bibitem{thulke2024climategpt}
D.~Thulke, Y.~Gao, P.~Pelser, R.~Brune, R.~Jalota, F.~Fok, M.~Ramos, I.~van Wyk, A.~Nasir, H.~Goldstein, et~al.
\newblock Climategpt: Towards ai synthesizing interdisciplinary research on climate change.
\newblock {\em arXiv preprint arXiv:2401.09646}, 2024.

\bibitem{touvron2023llama}
H.~Touvron, T.~Lavril, G.~Izacard, X.~Martinet, M.-A. Lachaux, T.~Lacroix, B.~Rozi{\`e}re, N.~Goyal, E.~Hambro, F.~Azhar, et~al.
\newblock Llama: Open and efficient foundation language models.
\newblock {\em arXiv preprint arXiv:2302.13971}, 2023.

\bibitem{touvron2023llama2}
H.~Touvron, L.~Martin, K.~Stone, P.~Albert, A.~Almahairi, Y.~Babaei, N.~Bashlykov, S.~Batra, P.~Bhargava, S.~Bhosale, et~al.
\newblock Llama 2: Open foundation and fine-tuned chat models.
\newblock {\em arXiv preprint arXiv:2307.09288}, 2023.

\bibitem{van2022three}
G.~M. Van~de Ven, T.~Tuytelaars, and A.~S. Tolias.
\newblock Three types of incremental learning.
\newblock {\em Nature Machine Intelligence}, 4(12):1185--1197, 2022.

\bibitem{verwimp2024continual}
E.~Verwimp, R.~Aljundi, S.~Ben-David, M.~Bethge, A.~Cossu, A.~Gepperth, T.~L. Hayes, E.~Hüllermeier, C.~Kanan, D.~Kudithipudi, C.~H. Lampert, M.~Mundt, R.~Pascanu, A.~Popescu, A.~S. Tolias, J.~van~de Weijer, B.~Liu, V.~Lomonaco, T.~Tuytelaars, and G.~M. van~de Ven.
\newblock Continual learning: Applications and the road forward, 2024.

\bibitem{volske2017tl}
M.~V{\"o}lske, M.~Potthast, S.~Syed, and B.~Stein.
\newblock Tl; dr: Mining reddit to learn automatic summarization.
\newblock In {\em Proceedings of the Workshop on New Frontiers in Summarization}, pages 59--63, 2017.

\bibitem{gpt-j}
B.~Wang and A.~Komatsuzaki.
\newblock {GPT-J-6B: A 6 Billion Parameter Autoregressive Language Model}.
\newblock \url{https://github.com/kingoflolz/mesh-transformer-jax}, May 2021.

\bibitem{wang2022coscl}
L.~Wang, X.~Zhang, Q.~Li, J.~Zhu, and Y.~Zhong.
\newblock Coscl: Cooperation of small continual learners is stronger than a big one.
\newblock In {\em Computer Vision--ECCV 2022: 17th European Conference, Tel Aviv, Israel, October 23--27, 2022, Proceedings, Part XXVI}, pages 254--271. Springer, 2022.

\bibitem{wang2024comprehensive}
L.~Wang, X.~Zhang, H.~Su, and J.~Zhu.
\newblock A comprehensive survey of continual learning: Theory, method and application.
\newblock {\em IEEE Transactions on Pattern Analysis and Machine Intelligence}, pages 1--20, 2024.

\bibitem{wang2024wise}
P.~Wang, Z.~Li, N.~Zhang, Z.~Xu, Y.~Yao, Y.~Jiang, P.~Xie, F.~Huang, and H.~Chen.
\newblock Wise: Rethinking the knowledge memory for lifelong model editing of large language models.
\newblock {\em arXiv preprint arXiv:2405.14768}, 2024.

\bibitem{wang2021kadapter}
R.~Wang, D.~Tang, N.~Duan, Z.~Wei, X.~Huang, J.~Ji, G.~Cao, D.~Jiang, and M.~Zhou.
\newblock {K-Adapter}: {I}nfusing {K}nowledge into {P}re-{T}rained {M}odels with {A}dapters.
\newblock In C.~Zong, F.~Xia, W.~Li, and R.~Navigli, editors, {\em Findings of the Association for Computational Linguistics: ACL-IJCNLP 2021}, pages 1405--1418, Online, Aug. 2021. Association for Computational Linguistics.

\bibitem{wang2023orthogonal}
X.~Wang, T.~Chen, Q.~Ge, H.~Xia, R.~Bao, R.~Zheng, Q.~Zhang, T.~Gui, and X.~Huang.
\newblock Orthogonal subspace learning for language model continual learning.
\newblock In H.~Bouamor, J.~Pino, and K.~Bali, editors, {\em Findings of the Association for Computational Linguistics: EMNLP 2023}, pages 10658--10671, Singapore, Dec. 2023. Association for Computational Linguistics.

\bibitem{wang2023trace}
X.~Wang, Y.~Zhang, T.~Chen, S.~Gao, S.~Jin, X.~Yang, Z.~Xi, R.~Zheng, Y.~Zou, T.~Gui, Q.~Zhang, and X.~Huang.
\newblock Trace: A comprehensive benchmark for continual learning in large language models, 2023.

\bibitem{wang2023codet5plus}
Y.~Wang, H.~Le, A.~D. Gotmare, N.~D.~Q. Bui, J.~Li, and S.~C.~H. Hoi.
\newblock Codet5+: Open code large language models for code understanding and generation, 2023.

\bibitem{wang2024inscl}
Y.~Wang, Y.~Liu, C.~Shi, H.~Li, C.~Chen, H.~Lu, and Y.~Yang.
\newblock Inscl: A data-efficient continual learning paradigm for fine-tuning large language models with instructions, 2024.

\bibitem{wang2022supernaturalinstructions}
Y.~Wang, S.~Mishra, P.~Alipoormolabashi, Y.~Kordi, A.~Mirzaei, A.~Arunkumar, A.~Ashok, A.~S. Dhanasekaran, A.~Naik, D.~Stap, E.~Pathak, G.~Karamanolakis, H.~G. Lai, I.~Purohit, I.~Mondal, J.~Anderson, K.~Kuznia, K.~Doshi, M.~Patel, K.~K. Pal, M.~Moradshahi, M.~Parmar, M.~Purohit, N.~Varshney, P.~R. Kaza, P.~Verma, R.~S. Puri, R.~Karia, S.~K. Sampat, S.~Doshi, S.~Mishra, S.~Reddy, S.~Patro, T.~Dixit, X.~Shen, C.~Baral, Y.~Choi, N.~A. Smith, H.~Hajishirzi, and D.~Khashabi.
\newblock Super-naturalinstructions: Generalization via declarative instructions on 1600+ nlp tasks, 2022.

\bibitem{wang2021codet5}
Y.~Wang, W.~Wang, S.~Joty, and S.~C. Hoi.
\newblock Codet5: Identifier-aware unified pre-trained encoder-decoder models for code understanding and generation.
\newblock In {\em EMNLP}, 2021.

\bibitem{wang2024codeclm}
Z.~Wang, C.-L. Li, V.~Perot, L.~T. Le, J.~Miao, Z.~Zhang, C.-Y. Lee, and T.~Pfister.
\newblock Codeclm: Aligning language models with tailored synthetic data.
\newblock {\em arXiv preprint arXiv:2404.05875}, 2024.

\bibitem{wang2022sparcl}
Z.~Wang, Z.~Zhan, Y.~Gong, G.~Yuan, W.~Niu, T.~Jian, B.~Ren, S.~Ioannidis, Y.~Wang, and J.~Dy.
\newblock Sparcl: Sparse continual learning on the edge.
\newblock {\em Advances in Neural Information Processing Systems}, 35:20366--20380, 2022.

\bibitem{wang2022dualprompt}
Z.~Wang, Z.~Zhang, S.~Ebrahimi, R.~Sun, H.~Zhang, C.-Y. Lee, X.~Ren, G.~Su, V.~Perot, J.~Dy, et~al.
\newblock Dualprompt: Complementary prompting for rehearsal-free continual learning.
\newblock {\em European Conference on Computer Vision}, 2022.

\bibitem{wang2022learning}
Z.~Wang, Z.~Zhang, C.-Y. Lee, H.~Zhang, R.~Sun, X.~Ren, G.~Su, V.~Perot, J.~Dy, and T.~Pfister.
\newblock Learning to prompt for continual learning.
\newblock In {\em Proceedings of the IEEE/CVF Conference on Computer Vision and Pattern Recognition}, pages 139--149, 2022.

\bibitem{wei2021finetuned}
J.~Wei, M.~Bosma, V.~Y. Zhao, K.~Guu, A.~W. Yu, B.~Lester, N.~Du, A.~M. Dai, and Q.~V. Le.
\newblock Finetuned language models are zero-shot learners.
\newblock {\em arXiv preprint arXiv:2109.01652}, 2021.

\bibitem{wei2022finetuned}
J.~Wei, M.~Bosma, V.~Y. Zhao, K.~Guu, A.~W. Yu, B.~Lester, N.~Du, A.~M. Dai, and Q.~V. Le.
\newblock Finetuned language models are zero-shot learners, 2022.

\bibitem{wei2022emergent}
J.~Wei, Y.~Tay, R.~Bommasani, C.~Raffel, B.~Zoph, S.~Borgeaud, D.~Yogatama, M.~Bosma, D.~Zhou, D.~Metzler, et~al.
\newblock Emergent abilities of large language models.
\newblock {\em arXiv preprint arXiv:2206.07682}, 2022.

\bibitem{wei2022chain}
J.~Wei, X.~Wang, D.~Schuurmans, M.~Bosma, F.~Xia, E.~Chi, Q.~V. Le, D.~Zhou, et~al.
\newblock Chain-of-thought prompting elicits reasoning in large language models.
\newblock {\em Advances in neural information processing systems}, 35:24824--24837, 2022.

\bibitem{weyssow2023usage}
M.~Weyssow, X.~Zhou, K.~Kim, D.~Lo, and H.~Sahraoui.
\newblock On the usage of continual learning for out-of-distribution generalization in pre-trained language models of code.
\newblock In {\em Proceedings of the 31st ACM Joint European Software Engineering Conference and Symposium on the Foundations of Software Engineering}, ESEC/FSE 2023, page 1470–1482, New York, NY, USA, 2023. Association for Computing Machinery.

\bibitem{winata2023overcoming}
G.~Winata, L.~Xie, K.~Radhakrishnan, S.~Wu, X.~Jin, P.~Cheng, M.~Kulkarni, and D.~Preotiuc-Pietro.
\newblock Overcoming catastrophic forgetting in massively multilingual continual learning.
\newblock In A.~Rogers, J.~Boyd-Graber, and N.~Okazaki, editors, {\em Findings of the Association for Computational Linguistics: ACL 2023}, pages 768--777, Toronto, Canada, July 2023. Association for Computational Linguistics.

\bibitem{wistuba2023}
M.~Wistuba, P.~T. Sivaprasad, L.~Balles, and G.~Zappella.
\newblock Continual learning with low rank adaptation.
\newblock In {\em NeurIPS 2023 Workshop on Distribution Shifts (DistShifts)}, 2023.

\bibitem{wu2024llama}
C.~Wu, Y.~Gan, Y.~Ge, Z.~Lu, J.~Wang, Y.~Feng, P.~Luo, and Y.~Shan.
\newblock Llama pro: Progressive llama with block expansion, 2024.

\bibitem{wu2023pmc}
C.~Wu, W.~Lin, X.~Zhang, Y.~Zhang, Y.~Wang, and W.~Xie.
\newblock Pmc-llama: Towards building open-source language models for medicine.
\newblock {\em arXiv preprint arXiv:2305.10415}, 6, 2023.

\bibitem{Wu2023BloombergGPT}
S.~Wu, O.~Irsoy, S.~Lu, V.~Dabravolski, M.~Dredze, S.~Gehrmann, P.~Kambadur, D.~S. Rosenberg, and G.~Mann.
\newblock Bloomberggpt: {A} large language model for finance.
\newblock {\em CoRR}, abs/2303.17564, 2023.

\bibitem{wu2021pretrained}
T.~Wu, M.~Caccia, Z.~Li, Y.-F. Li, G.~Qi, and G.~Haffari.
\newblock Pretrained language model in continual learning: A comparative study.
\newblock In {\em International conference on learning representations}, 2021.

\bibitem{wu2024continual}
T.~Wu, L.~Luo, Y.-F. Li, S.~Pan, T.-T. Vu, and G.~Haffari.
\newblock Continual learning for large language models: A survey, 2024.

\bibitem{wu2019large}
Y.~Wu, Y.~Chen, L.~Wang, Y.~Ye, Z.~Liu, Y.~Guo, and Y.~Fu.
\newblock Large scale incremental learning.
\newblock In {\em Proceedings of the IEEE/CVF Conference on Computer Vision and Pattern Recognition}, pages 374--382, 2019.

\bibitem{wu2018kanerva}
Y.~Wu, G.~Wayne, A.~Graves, and T.~Lillicrap.
\newblock The kanerva machine: A generative distributed memory.
\newblock {\em arXiv preprint arXiv:1804.01756}, 2018.

\bibitem{xie2023quert}
J.~Xie, Y.~Liang, J.~Liu, Y.~Xiao, B.~Wu, and S.~Ni.
\newblock Quert: Continual pre-training of language model for query understanding in travel domain search.
\newblock In {\em Proceedings of the 29th ACM SIGKDD Conference on Knowledge Discovery and Data Mining}, KDD '23, page 5282–5291, New York, NY, USA, 2023. Association for Computing Machinery.

\bibitem{xie2024me}
Q.~Xie, Q.~Chen, A.~Chen, C.~Peng, Y.~Hu, F.~Lin, X.~Peng, J.~Huang, J.~Zhang, V.~Keloth, et~al.
\newblock Me llama: Foundation large language models for medical applications.
\newblock {\em arXiv preprint arXiv:2402.12749}, 2024.

\bibitem{Xie2023PIXIU}
Q.~Xie, W.~Han, X.~Zhang, Y.~Lai, M.~Peng, A.~Lopez{-}Lira, and J.~Huang.
\newblock {PIXIU:} {A} large language model, instruction data and evaluation benchmark for finance.
\newblock {\em CoRR}, abs/2306.05443, 2023.

\bibitem{xie2024data}
S.~M. Xie, S.~Santurkar, T.~Ma, and P.~S. Liang.
\newblock Data selection for language models via importance resampling.
\newblock {\em Advances in Neural Information Processing Systems}, 36, 2024.

\bibitem{xie2023efficient}
Y.~Xie, K.~Aggarwal, and A.~Ahmad.
\newblock Efficient continual pre-training for building domain specific large language models, 2023.

\bibitem{xu2023wizardlm}
C.~Xu, Q.~Sun, K.~Zheng, X.~Geng, P.~Zhao, J.~Feng, C.~Tao, and D.~Jiang.
\newblock Wizardlm: Empowering large language models to follow complex instructions.
\newblock {\em arXiv preprint arXiv:2304.12244}, 2023.

\bibitem{xu2019bert}
H.~Xu, B.~Liu, L.~Shu, and P.~S. Yu.
\newblock Bert post-training for review reading comprehension and aspect-based sentiment analysis, 2019.

\bibitem{Xue2023WeaverBird}
S.~Xue, F.~Zhou, Y.~Xu, H.~Zhao, S.~Xie, Q.~Dai, C.~Jiang, J.~Zhang, J.~Zhou, D.~Xiu, and H.~Mei.
\newblock Weaverbird: Empowering financial decision-making with large language model, knowledge base, and search engine.
\newblock {\em CoRR}, abs/2308.05361, 2023.

\bibitem{yan2023af}
Y.~Yan, K.~Xue, X.~Shi, Q.~Ye, J.~Liu, and T.~Ruan.
\newblock Af adapter: Continual pretraining for building chinese biomedical language model.
\newblock In {\em 2023 IEEE International Conference on Bioinformatics and Biomedicine (BIBM)}, pages 953--957, Los Alamitos, CA, USA, dec 2023. IEEE Computer Society.

\bibitem{yang2024moral}
S.~Yang, M.~A. Ali, C.-L. Wang, L.~Hu, and D.~Wang.
\newblock Moral: Moe augmented lora for llms' lifelong learning, 2024.

\bibitem{Yang2023PLLaMa}
X.~Yang, J.~Gao, W.~Xue, and E.~Alexandersson.
\newblock Pllama: An open-source large language model for plant science.
\newblock {\em CoRR}, abs/2401.01600, 2024.

\bibitem{yang2024reawakening}
Y.~Yang, M.~Jones, M.~C. Mozer, and M.~Ren.
\newblock Reawakening knowledge: Anticipatory recovery from catastrophic interference via structured training, 2024.

\bibitem{yang2024recent}
Y.~Yang, J.~Zhou, X.~Ding, T.~Huai, S.~Liu, Q.~Chen, L.~He, and Y.~Xie.
\newblock Recent advances of foundation language models-based continual learning: A survey.
\newblock {\em arXiv preprint arXiv:2405.18653}, 2024.

\bibitem{yao2024tree}
S.~Yao, D.~Yu, J.~Zhao, I.~Shafran, T.~Griffiths, Y.~Cao, and K.~Narasimhan.
\newblock Tree of thoughts: Deliberate problem solving with large language models.
\newblock {\em Advances in Neural Information Processing Systems}, 36, 2024.

\bibitem{yildiz2024investigating}
{\c{C}}.~Y{\i}ld{\i}z, N.~K. Ravichandran, P.~Punia, M.~Bethge, and B.~Ermis.
\newblock Investigating continual pretraining in large language models: Insights and implications.
\newblock {\em arXiv preprint arXiv:2402.17400}, 2024.

\bibitem{yin2022contintin}
W.~Yin, J.~Li, and C.~Xiong.
\newblock {C}on{T}in{T}in: Continual learning from task instructions.
\newblock In S.~Muresan, P.~Nakov, and A.~Villavicencio, editors, {\em Proceedings of the 60th Annual Meeting of the Association for Computational Linguistics (Volume 1: Long Papers)}, pages 3062--3072, Dublin, Ireland, May 2022. Association for Computational Linguistics.

\bibitem{yu2023melo}
L.~Yu, Q.~Chen, J.~Zhou, and L.~He.
\newblock Melo: Enhancing model editing with neuron-indexed dynamic lora.
\newblock {\em arXiv preprint arXiv:2312.11795}, 2023.

\bibitem{wei2022circle}
W.~Yuan, Q.~Zhang, T.~He, C.~Fang, N.~Q.~V. Hung, X.~Hao, and H.~Yin.
\newblock Circle: continual repair across programming languages.
\newblock In {\em Proceedings of the 31st ACM SIGSOFT International Symposium on Software Testing and Analysis}, ISSTA 2022, page 678–690, New York, NY, USA, 2022. Association for Computing Machinery.

\bibitem{yue2023mammoth}
X.~Yue, X.~Qu, G.~Zhang, Y.~Fu, W.~Huang, H.~Sun, Y.~Su, and W.~Chen.
\newblock Mammoth: Building math generalist models through hybrid instruction tuning.
\newblock {\em arXiv preprint arXiv:2309.05653}, 2023.

\bibitem{zellers2019defending}
R.~Zellers, A.~Holtzman, H.~Rashkin, Y.~Bisk, A.~Farhadi, F.~Roesner, and Y.~Choi.
\newblock Defending against neural fake news.
\newblock {\em Advances in neural information processing systems}, 32, 2019.

\bibitem{zhai2023investigating}
Y.~Zhai, S.~Tong, X.~Li, M.~Cai, Q.~Qu, Y.~J. Lee, and Y.~Ma.
\newblock Investigating the catastrophic forgetting in multimodal large language models, 2023.

\bibitem{Zhang2024SciGLM}
D.~Zhang, Z.~Hu, S.~Zhoubian, Z.~Du, K.~Yang, Z.~Wang, Y.~Yue, Y.~Dong, and J.~Tang.
\newblock Sciglm: Training scientific language models with self-reflective instruction annotation and tuning.
\newblock {\em CoRR}, abs/2401.07950, 2024.

\bibitem{zhang2023copf}
H.~Zhang, L.~Gui, Y.~Zhai, H.~Wang, Y.~Lei, and R.~Xu.
\newblock Copf: Continual learning human preference through optimal policy fitting.
\newblock {\em arXiv preprint arXiv:2310.15694}, 2023.

\bibitem{zhangcppo}
H.~Zhang, Y.~Lei, L.~Gui, M.~Yang, Y.~He, H.~Wang, and R.~Xu.
\newblock Cppo: Continual learning for reinforcement learning with human feedback.

\bibitem{zhang2024instruction}
S.~Zhang, L.~Dong, X.~Li, S.~Zhang, X.~Sun, S.~Wang, J.~Li, R.~Hu, T.~Zhang, F.~Wu, and G.~Wang.
\newblock Instruction tuning for large language models: A survey, 2024.

\bibitem{Zhang2023xuanyuan}
X.~Zhang and Q.~Yang.
\newblock Xuanyuan 2.0: A large chinese financial chat model with hundreds of billions parameters.
\newblock In {\em Proceedings of the 32nd ACM International Conference on Information and Knowledge Management}, CIKM '23, page 4435–4439, New York, NY, USA, 2023. Association for Computing Machinery.

\bibitem{zhang2015character}
X.~Zhang, J.~Zhao, and Y.~LeCun.
\newblock Character-level convolutional networks for text classification.
\newblock In C.~Cortes, N.~Lawrence, D.~Lee, M.~Sugiyama, and R.~Garnett, editors, {\em Advances in Neural Information Processing Systems}, volume~28. Curran Associates, Inc., 2015.

\bibitem{zhang2022continual}
Y.~Zhang, X.~Wang, and D.~Yang.
\newblock Continual sequence generation with adaptive compositional modules.
\newblock In S.~Muresan, P.~Nakov, and A.~Villavicencio, editors, {\em Proceedings of the 60th Annual Meeting of the Association for Computational Linguistics (Volume 1: Long Papers)}, pages 3653--3667, Dublin, Ireland, May 2022. Association for Computational Linguistics.

\bibitem{zhang2023citb}
Z.~Zhang, M.~Fang, L.~Chen, and M.-R. Namazi-Rad.
\newblock {CITB}: A benchmark for continual instruction tuning.
\newblock In H.~Bouamor, J.~Pino, and K.~Bali, editors, {\em Findings of the Association for Computational Linguistics: EMNLP 2023}, pages 9443--9455, Singapore, Dec. 2023. Association for Computational Linguistics.

\bibitem{zhao-etal-2023-c}
C.~Zhao, Y.~Li, and C.~Caragea.
\newblock {C}-{STANCE}: A large dataset for {C}hinese zero-shot stance detection.
\newblock In A.~Rogers, J.~Boyd-Graber, and N.~Okazaki, editors, {\em Proceedings of the 61st Annual Meeting of the Association for Computational Linguistics (Volume 1: Long Papers)}, pages 13369--13385, Toronto, Canada, July 2023. Association for Computational Linguistics.

\bibitem{zhao2024large}
H.~Zhao, H.~Han, J.~Shi, C.~Du, J.~Liang, and Y.~Xiao.
\newblock Large language model can continue evolving from mistakes.
\newblock {\em arXiv preprint arXiv:2404.08707}, 2024.

\bibitem{zhao2022memory}
H.~Zhao, H.~Wang, Y.~Fu, F.~Wu, and X.~Li.
\newblock Memory-efficient class-incremental learning for image classification.
\newblock {\em IEEE Transactions on Neural Networks and Learning Systems}, 33(10):5966--5977, 2022.

\bibitem{zhao2024reconstruct}
S.~Zhao, X.~Zou, T.~Yu, and H.~Xu.
\newblock Reconstruct before query: Continual missing modality learning with decomposed prompt collaboration, 2024.

\bibitem{zhao2024sapt}
W.~Zhao, S.~Wang, Y.~Hu, Y.~Zhao, B.~Qin, X.~Zhang, Q.~Yang, D.~Xu, and W.~Che.
\newblock Sapt: A shared attention framework for parameter-efficient continual learning of large language models, 2024.

\bibitem{zheng2024antiforgetting}
J.~Zheng, Q.~Ma, Z.~Liu, B.~Wu, and H.~Feng.
\newblock Beyond anti-forgetting: Multimodal continual instruction tuning with positive forward transfer, 2024.

\bibitem{zheng2023learn}
J.~Zheng, S.~Qiu, and Q.~Ma.
\newblock Learn or recall? revisiting incremental learning with pre-trained language models, 2023.

\bibitem{zheng2023preventing}
Z.~Zheng, M.~Ma, K.~Wang, Z.~Qin, X.~Yue, and Y.~You.
\newblock Preventing zero-shot transfer degradation in continual learning of vision-language models.
\newblock In {\em Proceedings of the IEEE/CVF International Conference on Computer Vision}, pages 19125--19136, 2023.

\bibitem{Zheng2023MarineGPT}
Z.~Zheng, J.~Zhang, T.~Vu, S.~Diao, Y.~H.~W. Tim, and S.~Yeung.
\newblock Marinegpt: Unlocking secrets of ocean to the public.
\newblock {\em CoRR}, abs/2310.13596, 2023.

\bibitem{zhou2019goingvacationtakeslonger}
B.~Zhou, D.~Khashabi, Q.~Ning, and D.~Roth.
\newblock "going on a vacation" takes longer than "going for a walk": A study of temporal commonsense understanding, 2019.

\bibitem{zhou2020pre}
W.~Zhou, D.-H. Lee, R.~K. Selvam, S.~Lee, B.~Y. Lin, and X.~Ren.
\newblock Pre-training text-to-text transformers for concept-centric common sense.
\newblock 2021.

\bibitem{zhu2024model}
D.~Zhu, Z.~Sun, Z.~Li, T.~Shen, K.~Yan, S.~Ding, K.~Kuang, and C.~Wu.
\newblock Model tailor: Mitigating catastrophic forgetting in multi-modal large language models, 2024.

\end{thebibliography}
}

\appendix
\clearpage
\section*{\LARGE Supplementary Material}

\section{Preliminaries}
\label{app:preliminary}
In this section, we provide an overview of the fundamental concepts of large language models (LLMs) and continual learning~(CL)
We begin by introducing the notation used in this paper. Subsequently, we discuss the pre-training and downstream adaptation of LLMs, as well as mainstream LLM families (\appref{app:preliminary-llm}), followed by an introduction to basic continual learning techniques studied by the community (\appref{app:preliminary-cl}).

\textbf{Notation.}\quad
We denote scalars with lowercase letters, vectors with lowercase boldface letters, and matrices with uppercase boldface letters.
The $l_2$-norm of vectors and the Frobenius norm of a matrix are represented by $\|\cdot\|_2$. For a vector $\vv = [v_1, v_2, \cdots, v_n]^\top$, $\|\vv\|_2 = (\sum_{i=1}^{n} v_i^2)^{\nicefrac{1}{2}}$; for a matrix $\mA\in \mathbb{R}^{m\times n}$, $\|\mA\|_2 = (\sum_{ij} A_{ij}^2)^{\nicefrac{1}{2}}$.
We use $\epsilon_{\gD}$, $\gL_{\gD}$ to denote the error function, and loss function that is deployed for training, respectively, where the subscript is used to denote the error/loss measured by taking the expectation on the data distribution $\gD$. 
We further use $\hat{\gL}_{S}$ to represent the empirical evaluation of the loss function $\gL$ over the set of examples $S$.
Probability and expectation are denoted by $P$ and $\E$, respectively. 
We use $[m]$ to denote the set of positive integers up to $m$, $\{1, \cdots, m\}$.

\subsection{Large Language Models}
\label{app:preliminary-llm}
In the past two decades, neural language modeling has emerged as the dominant field of deep learning, marked by significant and rapid advancements. Primarily built on the transformer architecture, pre-trained language models~(PLMs) like BERT have established a universal hidden embedding space through extensive pre-training on large-scale unlabeled text corpora. 
Following the pre-training and fine-tuning paradigms, PLMs exhibit promising performance across various natural language processing tasks after being fine-tuned upon small amounts of task-specific data~\cite{devlin2018bert, liu2019roberta, raffel2020exploring}.
Research on scaling laws indicates that increasing model size enhances the capacity of language modelss~\cite{kaplan2020scaling, hoffmann2022training}.
By scaling parameters to billions or even hundreds of billions and training on massive text datasets, PLMs not only demonstrate superior language understanding and generation capabilities but also manifest emergent abilities such as in-context learning, instruction following, and multi-step reasoning, which are absent in small-scale language models like BERT~\cite{wei2022chain,wei2022emergent,yao2024tree,wei2021finetuned,min2022rethinking}. 
These larger models are commonly referred to as Large Language Models~(LLMs).

\subsubsection{Pre-Training of LLMs}
\label{app:preliminary-lmm-pt}
Pre-training is essential for language models to acquire broad language representations. Decoder-only models typically employ probability language modeling (LM) tasks during pre-training. LM, in this context, specifically refers to auto-regressive LM.
Given a sequence of tokens $\vx = [x_1, x_2, \cdots, x_N ]$, LM predicts the next token $x_t$ autoregressively based on all preceding tokens $\vx_{<t} = [x_1, x_2, \cdots, x_{t-1}]$, and trains the entire network by minimizing the negative log-likelihood:
\begin{align}\label{lm}
    \mathcal{L}_{{\rm LM}}(\vx) &\triangleq -\sum^N_{t=1} \log P( x_t | \vx_{<t} ), 
\end{align}
where $P(x_1|\vx_{<1})\triangleq P(x_1)$ is the unconditional probability estimation of the first token.
The three most popular families of decoder-only models are GPT, PaLM, and LLaMA. The GPT family, developed by OpenAI, includes models such as GPT-2~\cite{radford2019language}, GPT-3 ~\cite{brown2020language}, ChatGPT~\cite{achiam2022chatgpt}, and GPT-4 ~\cite{achiam2023gpt}. Notably, GPT-3 was the first LLM to exhibit emergent abilities not found in smaller PLMs. Another notable family, Gemini, developed by Google, is comparable to the GPT family~\cite{team2023gemini, reid2024gemini}. While both GPT and Gemini families are closed-source, LLaMA, released by Meta, is currently the most popular open-source family of LLMs~\cite{touvron2023llama, touvron2023llama2}. The weights of these models are made available to the research community under non-commercial licenses.

Masked language modeling (MLM) task serves as a common pre-training objective for encoder-only models like BERT~\cite{devlin2018bert,liu2019roberta}. 
In MLM, for the input sequence $\vx$, a subset of input tokens $m(\vx)$ are masked and replaced with the special [MASK] token. The pre-training goal is to utilize the unmasked parts $\vx_{\backslash m(\vx)}$ to predict the masked portions $m(\vx)$. 
In summary, the overarching goal of MLM is to minimize the negative log-likelihood:
\begin{align}\label{mlm}
    \mathcal{L}_{{\rm MLM}}(\vx) &\triangleq -\sum_{\hat{x} \in m(\vx)}{\rm log} \, P( \hat{x}|\vx_{\backslash m(\vx)} ).
\end{align}

Some encoder-decoder architecture models, such as T5~\cite{raffel2020exploring}, also utilize Sequence-to-Sequence MLM task as the pre-training objective. They take masked sentences as encoder inputs and utilize the decoder to sequentially predict the masked tokens.

\subsubsection{Adaptation of LLMs}
\label{app:preliminary-llm-adaptation}
LLMs are primarily trained to generate linguistically coherent text. However, this training may not align with human values, preferences, or practical needs. Furthermore, the pre-training data can be outdated, leading to knowledge cutoffs or inaccuracies. To address these issues, various computational paradigms such as Instruction Tuning~(IT)~\cite{zhang2024instruction}, Model Refinement~(MR)~\cite{de2021editing}, and Model Alignment~(MA)~\cite{ouyang2022rlhf,rafailov2024dpo} have been proposed. These approaches adapt LLMs to better meet diverse downstream tasks and user requirements.

\begin{definition}[\textbf{Instruction Tuning, IT}]\label{def:it}
    Let $h(\vx)$ be a language model that takes as input data $\vx$, typically consisting of natural language instructions or queries. Instruction Tuning (IT) is a specialized training approach designed to enhance the model's ability to accurately and effectively respond to specific instructions. The objective of IT is to refine $h$ by adjusting its parameters using a designated set of training examples $\gI = \{(\vx_i, \hat{\vy}_i)\}_{i=1}^N$ drawn from the IT data distribution $\gD_\gI$, where $\hat{\vy}_i$ represents the desired output for $\vx$. This set is curated to target specific tasks or functionalities that require improved performance. 
    Formally, IT seeks to find an optimal refined hypothesis $h^*$ that satisfies:
    \begin{align}
    \label{eq:it}
        h^* &\triangleq \arg\min_{h^\prime} \E_{(\vx, \vy) \sim \gD_\gI} \left[ -\log P(\hat{\vy}|\vx, h^\prime) \right] 
        \approx \arg\min_{h^\prime} \sum_{i=1}^N -\log P(\hat{\vy}_i|\vx_i, h^\prime).
    \end{align}
\end{definition}

\begin{remark}
    The task of Model Alignment~(MA) is usually formulated in the same problem definition as IT, with an alignment dataset of size $M$ as $\gA = \{(\vx_a, \vy_a, \hat{\vy}_a)\}_{a=1}^M$, where $\vy_a$ represents the model's original decision for input $\vx_a$, and $\hat{\vy}_a$ denotes the aligned decision that adheres to specified ethical guidelines or desired outcomes.
\end{remark}
    
\begin{definition}[\textbf{Model Refinement, MR}]\label{def:mr}
    Suppose we have a model $h(\vx)$ taking data $\vx$ (e.g., natural language queries) as inputs. Consider a size-$N$ editing set $\gE = \{(\vx_e, \vy_e, \hat{\vy}_e)\}_{e=1}^N$, where $\hat{\vy}_e$ denotes the true label of $\vx_e$, but the model incorrectly outputs $\vy_e$ for $\vx_e$. Model Refinement~(MR) aims to efficiently update the model from $h$ to $h^\prime$ such that it correctly predicts the editing set $\gE$, while preserving the original outputs outside $\gE$. Formally, we aims to find $h^\prime$ satisfying
    \begin{align}
    \label{eq:mr}
        h^\prime(\vx_0) = 
        \begin{cases}
            \hat{\vy}_0 & \text{if } (\vx_0, \hat{\vy}_0) \in \gE, \\
            h(\vx_0) & o.w.
        \end{cases}
    \end{align}
\end{definition}

\subsection{Continual Learning}
\label{app:preliminary-cl}
Humans gradually accumulate knowledge and skills across tasks without significant performance decline on previous tasks~\cite{mcclelland1995there,kandel2000principles,pallier2003brain,mccaffary2021towards}. In contrast, machine learning models are usually data-centric, minimizing the training loss on the subsequent tasks will cause the model fail on the old ones, which phenomenon is phrased as \emph{``catastrophic forgetting''}. Addressing this challenge is a focal point in continual learning research. The problem of efficiently adapting models to a sequence of tasks without forgetting is extensively studied in the continual learning community~\cite{pentina2016theoretical,chen2018lifelong,van2022three,wang2024comprehensive}. These studies are typically conducted under the following memory constraint of CL.

\begin{definition}[\textbf{Memory Constraint of Continual Learning}]\label{def:memory}
Suppose $T$ sets of observations $\{S_t\sim \gT_t\}_{t=1}^T$ come in as a sequence, where $\{\gT_t\}_{t=1}^T$ denotes the $T$ task distributions . At the learning stage $t>1$, the sets of observations $\{S_i\}_{i=1}^{t-1}$ are not accessible (\textbf{strong}) or only partially accessible (\textbf{relaxed}).
\end{definition}

\begin{remark}
    In early stages of CL, works mostly focused on the strong memory constraint~\cite{kirkpatrick2017overcoming,li2017learning,aljundi2018memory,lomonaco2020rehearsalfree}; as the research field progresses, more focus was put on relaxing the memory constraint to a small buffer for replay~\cite{rebuffi2017icarl,chaudhry2019tiny,buzzega2020dark,shi2024unified}; some modern CL works completely discard the memory constraint but put focus on the computational budget~\cite{prabhu2023online,verwimp2024continual}.
\end{remark}

\subsubsection{Three Types of Continual Learning}
There are three outstanding types of continual learning scenarios: task-incremental learning~(TIL), domain-incremental learning~(DIL), and class-incremental learning~(CIL).
To establish a groundwork for subsequent discussions (as illustrated in \Tabref{tab:cft} and \Secref{sec:discussion-xil}), we adhere to the conceptual framework proposed by \cite{van2022three,kim2022theoretical,wang2024comprehensive} and offer formal definitions for these three continual learning scenarios.

\begin{definition}[\textbf{Task-Incremental Learning, TIL}]\label{def:til}
Suppose $T$ task distributions $\{\gT_t\}_{t=1}^T$ come in as a sequence, where $\gT_t$ denotes the joint distribution over the $t$-th task's input space and the label space $(\gX_t, \gY_t)$. Denote $\gX \triangleq \bigcup_{t=1}^T \gX_t$ and $\gY \triangleq \bigcup_{t=1}^T \gY_t$ as the union of the input and label spaces, respectively.
Under the memory constraint defined in \Defref{def:memory}, Task-Incremental Learning~(TIL) aims to find the optimal hypothesis $h^*: \gX \times [T] \rightarrow \gY$ that satisfies:
\begin{align}
    h^* &= \arg\min_{h} \sum_{t=1}^{T} \E_{(\vx, y)\sim \gT_t} \left[ \mathbbm{1}_{h(\vx, t)\neq y} \right].
\end{align}
\end{definition}

\begin{definition}[\textbf{Domain-Incremental Learning, DIL}]\label{def:dil}
Suppose $T$ domain distributions $\{\gD_t\}_{t=1}^T$ come in as a sequence, where $\gD_t$ denotes the $t$-th joint distribution over the shared input space and label space $(\gX, \gY)$. 
Under the memory constraint defined in \Defref{def:memory}, Domain-Incremental Learning~(DIL) aims to find the optimal hypothesis $h^*: \gX \rightarrow \gY$ that satisfies:
\begin{align}
    h^* &= \arg\min_{h} \sum_{t=1}^{T} \E_{(\vx, y)\sim \gD_t} \left[ \mathbbm{1}_{h(\vx)\neq y} \right].
\end{align}
\end{definition}

\begin{definition}[\textbf{Class-Incremental Learning, CIL}]\label{def:cil}
Suppose $T$ task distributions $\{\gT_t\}_{t=1}^T$ come in as a sequence, where $\gT_t$ denotes the joint distribution over the $t$-th task's input space and the label space $(\gX_t, \gY_t)$. Denote $\gX \triangleq \bigcup_{t=1}^T \gX_t$ and $\gY \triangleq \bigcup_{t=1}^T \gY_t$ as the union of the input and label spaces, respectively.
Under the memory constraint defined in \Defref{def:memory}, Class-Incremental Learning~(CIL) aims to find the optimal hypothesis $h^*: \gX \rightarrow [T] \times \gY$ that satisfies:
\begin{align}
    h^* &= \arg\min_{h} \sum_{t=1}^{T} \E_{(\vx, y)\sim \gT_t} \left[ \mathbbm{1}_{h(\vx) \neq (t,y)} \right].
\end{align}
\end{definition}

\begin{remark}
    In TIL, it is common to have a shared input space $\gX = \gX_t, \forall t \in [T]$, but the space of the label distribution $\gY_t$ can be distinct~($\gY_i \cap \gY_j = \emptyset, \forall i\neq j$), partially shared~($\gY_i \cap \gY_j \neq \emptyset, \exists i\neq j$), or shared across different tasks~($\gY = \gY_t, \forall t \in [T]$). 
    In DIL, the tasks are defined in the same format, i.e., same input space $\gX$ and same output space $\gY$. During the inference, no task IDs are provided for the hypothesis, which means the continual learning model needs to capture the pattern between the domain-invariant features and the labels. DIL is commonly perceived as more difficult than TIL.
    CIL is commonly viewed as the most challenging continual learning scenario, as the model needs to infer the label and the task ID at the same time. Another possible formulation of CIL is to represent it as DIL but the output label spaces are disjoint, $\gY_i \cap \gY_j = \emptyset, \forall i\neq j$.
\end{remark}

\subsubsection{Techniques of Continual Learning}

The objective of CL is to find a hypothesis that minimizes risk across all tasks/domains. Consider DIL as an example~\cite{shi2024unified}, at $t$-th learning stage, the ideal training objective $\gL(h)$ is defined as
\begin{align}
    \gL(h) &\triangleq \underbrace{\sum_{i=1}^{t-1} \gL_{\gD_i}(h)}_{\text{past domains}} + \underbrace{\gL_{\gD_t}(h) \vphantom{\sum_{i=1}^{t-1} \gL_{\gD_i}(h)}}_{\text{current domain}}.
\end{align}
The objectives for past domains are often challenging to measure or optimize due to the memory constraints~(\Defref{def:memory}). Therefore, the core of designing CL algorithms lies in identifying a proxy learning objective for the first term without violating the memory constraint.
Existing CL techniques can be roughly categorized into 5 groups: (i)~replay-based, (ii)~regularization-based, (iii)~architecture-based, (iv)~optimization-based, and (v)~representation-based~\cite{wang2024comprehensive}.
Here, we provide a concise yet comprehensive introduction to the first three categories of continual learning techniques, as they find extensive application in continually learning large language models.

\textbf{Replay-Based Methods.}\quad 
Replay-based methods adopt the relaxed memory constraint by keeping a small buffer of observed data $\{M_i\}_{i=1}^{t-1}$ for each task $\gT_i$. Formally, they seek to optimize the following empirical training objective:
\begin{align}
    \hat{\gL}_{\text{replay}}(h) &\triangleq \underbrace{\sum_{i=1}^{t-1} \hat{\gL}_{M_i}(h)}_{\substack{\text{proxy for past domains}}} + \underbrace{\hat{\gL}_{S_t}(h) \vphantom{\sum_{i=1}^{t-1} \hat{\gL}_{M_i}(h)}}_{\text{current domain}},
\end{align}
where $\hat{\gL}_{S}$ denotes the empirical loss term evaluated on the set of examples $S$.
Often regarded as a simplistic solution to CL, replay-based methods may theoretically lead to loose generalization bounds~\cite{shi2024unified}. Despite this, they are valued for their simplicity, stability, and high performance, even with a small episodic memory~\cite{chaudhry2019tiny,riemer2018learning}. For instance, DER++~\cite{buzzega2020dark} demonstrates consistent performance enhancement by replaying a small set of past examples along with their logits~(known as dark experience replay). ESM-ER~\cite{sarfraz2023error} introduces error sensitivity modulation~(ESM) to mitigate abrupt representational drift caused by high-error new examples.
A significant focus in replay-based CL is enhancing sample efficiency for buffer maintenance. For instance, ~\cite{rebuffi2017icarl} prioritizes exemplar selection based on herding to accurately model class mean throughout class-incremental learning. 
\cite{zhao2022memory} propose storing low-fidelity examples to achieve memory-efficient exemplar set maintenance. 

\textbf{Regularization-Based Methods.}\quad
Suppose $h_{\vtheta_{t-1}}$ is the hypothesis yielded after the $t-1$-th stage of training, parameterized by $\vtheta_{t-1}$. Regularization-based methods utilize a regularization term as a proxy for past domain losses, determined by the distance in the parameter space.
\begin{align}
    \hat{\gL}_{\text{reg}}(h_\vtheta) &\triangleq \underbrace{\lambda \cdot \left\| \vtheta - \vtheta_{t-1}\right\|_\mSigma}_{\substack{\text{proxy for past domains}}} + \underbrace{\hat{\gL}_{S_t}(h_\vtheta) \vphantom{}}_{\text{current domain}},
\end{align}
where $\|\vv\|_\mSigma = \vv^\top \mSigma \vv$ is the vector norm evaluated on a positive-semi-definite matrix $\mSigma$, and $\lambda$ is the regularization coefficient, a hyper-parameter introduced to balance the past knowledge retention and current knowledge learning. 
The matrix $\mSigma$ introduced is to measure the different level of importance of each parameters and their correlations in retaining the past knowledge. 
In practice, to reduce computational overhead, diagonal matrices are often designed to encode only the importance of each parameter.
For example, Elastic Weight Consolidation (EWC)~\cite{kirkpatrick2017overcoming} adopts a Bayesian perspective, using diagonal values from the Fisher Information Matrix (FIM) as an approximation for the Hessian matrix of parameters. This forms a sequential Maximize A Posteriori (MAP) optimization for continual learning. Memory Aware Synapses (MAS)~\cite{aljundi2018memory} computes parameter importance in an online and unsupervised manner, defining importance by accumulated absolute gradient during training.
It is also worth noting that when $\mSigma=\mI$ degenerates to an identity matrix, the regularization term simplifies to a basic $l_2$-penalty term, equally penalizing each parameter, which can be surprisingly effective in some cases of continual LLMs~\cite{rongali2021continual}.

\textbf{Architecture-Based Methods.}\quad 
Expanding the network architecture dynamically to assimilate new knowledge is deemed the most efficient form of CL~\cite{wang2022learning,wang2022dualprompt}. This method primarily tackles adaptation challenges and can achieve zero-forgetting when task IDs are available during inference or can be correctly inferred~\cite{gururangan2022demix,wistuba2023}. 
However, due to the difficulty of task ID inference, architecture expansion is predominantly utilized in TIL but is scarcely explored in DIL or CIL.
Progressive Neural Networks~(PNN)~\cite{rusu2016progressive} proposes learning laterally connected neurons as new tasks arise, ensuring non-forgetting and enabling transfer of previously learned neurons for future tasks. In conjunction with pre-trained backbone large models like ViT~\cite{dosovitskiy2020image}, CoLoR~\cite{wistuba2023} trains various low-rank adaptation (LoRA)~\cite{hu2021lora} modules for different tasks. It estimates and stores prototypes for each task and utilizes the natural clustering ability of the pre-trained model during testing to infer task IDs, selecting the corresponding LoRA component for prediction generation.
In the domain of continual LLMs, architecture expansion has resurged in popularity following the rise of parameter-efficient fine-tuning (PEFT)~\cite{shazeer2017outrageously,hu2021lora,dettmers2023qlora}, a topic we will delve into shortly~\cite{yang2024moral,wang2023orthogonal,li2024examining,jang2022towards,jin2022lifelong,paul2024ircoder,yan2023af,wu2024llama}.


\section{Evaluation Protocols and Datasets}
\label{app:eval-and-data}

In \appref{app:eval}, we review common continual learning evaluation metrics and provide formal definitions.
In \appref{app:eval-llm}, we introduce metrics designed specifically for continual LLMs. 
Finally, in \appref{app:data}, we outline the datasets available for each discussed topic.

\subsection{Evaluation Metrics of Continual Learning}
\label{app:eval}
In the realm of conventional continual learning, where task streams take the form of classification, many metrics rely on the concept of Accuracy Matrix~\cite{lopez2017gradient,shi2024unified}.
Extending this notion to the context of continually learning LLMs, we introduce the \textbf{Performance Matrix} $\mP\in\mathbb{R}^{T\times T}$, where $T$ represents the total number of training stages.
Each entry of $\mP$ corresponds to a performance metric evaluated on the models, such as perplexity on pre-training data~\cite{jin2022lifelong,chen2023lifelong,gupta2023continual}, zero-shot/few-shot evaluation metrics on downstream data without fine-tuning~\cite{colombo2024saullm7b,wu2023pmc,Azerbayev2023LLEMMA,deng2023learning,nijkamp2022codegen,rozière2024code}, fine-tuned accuracies on downstream tasks~\cite{amba2021dynamic,qin2023recyclable,chen2023lifelong,jang2022towards}, and probing accuracies from fine-tuning add-on components evaluated on downstream tasks~\cite{tao2022can,luo2023investigating,zheng2023learn}.
In $\mP$, $P_{i,j}$ denotes the model's performance after training on task $i$ and evaluating on task $j$. With this Performance Matrix definition, we introduce the primary evaluation protocols widely adopted.

\textbf{Overall Performance~(OP)}.\quad 
The Overall Performance~(OP)~\cite{ke2021achieve,zhang2022continual,zhang2023copf} is a natural extension of the concept of Average Accuracy~\cite{lopez2017gradient,shi2024unified}. The OP measured up until training stage $t$ is the average performance of the model trained right after the stage $t$. Denote it as $\operatorname{OP}_t$ and we have: 
\begin{align}
    \operatorname{OP}_t &\triangleq \frac{1}{t} \sum_{i=1}^{t} P_{t,i}.
\end{align}
As noted in \cite{shi2024unified}, the OP corresponds to the primary optimization objective defined in \Defref{def:til}, \ref{def:dil}, and \ref{def:cil}. In much of the continual learning literature, once all $T$ tasks are completed, the final $\operatorname{OP}$~($\operatorname{OP}_T$) is reported, with the subscript $_T$ often omitted for brevity. 
In some works, OP is weighted by the importance of tasks $\tilde{\operatorname{OP}} \triangleq \frac{1}{T}\sum_{i=1}^T w_i P_{t,i}$, where $w_i = N_i / \sum_{j=1}^T N_j$ represents the ratio of data. In some literature, $\tilde{\operatorname{OP}}$ is referred to as ``example accuracy''~\cite{chen2024parameterizing}, ``whole accuracy''~\cite{song2023conpet}, or ``edit success rate'' in  CMR~\cite{hartvigsen2023aging}. 

\textbf{Forgetting (F).}\quad
Define $F_t$ as the forgetting up to task $t$, which represents the largest performance drop observed throughout the training process, averaged over $t$ training stages:
\begin{align}
    F_t &\triangleq \frac{1}{t-1}\sum_{j=1}^{t-1} \left[ \max_{l\in [t-1]} \{P_{l,j} - P_{t,j}\} \right].
\end{align}
Typically, researchers report the average forgetting $F=F_T$ at the end of the entire training process.
Forgetting quantifies the impact of learning new tasks on previously acquired knowledge. Ideally, a robust continual learning framework should achieve \textbf{Backward Transfer~(BWT)}, where learning new tasks enhances performance on prior tasks. This enhancement is typically measured by negating the forgetting, thus indicating an improvement in performance on earlier tasks. The concepts of Forgetting and Backward Transfer underpin various evaluation metrics, such as knowledge retention~\cite{jin2022lifelong}, performance on unchanged knowledge~\cite{jang2022temporalwiki}, average increased perplexity~(AP$^+$)~\cite{qin2022elle}, and test and edit retention rate in CMR~\cite{hartvigsen2023aging}.

\textbf{Forward Transfer (FWT).}\quad 
Forward Transfer measures the generalization ability of the continual learning algorithms. Formally, forward transfer $\operatorname{FWT}_t$ up to training stage $t$ is defined as
\begin{align}
    \operatorname{FWT}_t &\triangleq \frac{1}{t-1} \sum_{i=2}^{t}P_{i-1, i} - b_i,
\end{align}
where $b_i$ is the baseline performance of the model evaluated on task~$i$ before undergoing continual learning. Strictly speaking, the definition of $b_i$ is not the same as defined in the previous work~\cite{lopez2017gradient,shi2024unified}, where it is used to denote the performance of a random initialization of the model.
Additionally, we extend the notation of forward transfer in the vertical direction to represent the performance improvement on downstream tasks resulting from domain-adaptive pre-training~(see \Tabref{tab:dap}). 
Forward Transfer is alternatively referred to as temporal generalization~\cite{jin2022lifelong} or knowledge transfer~\cite{lazaridou2021mind} in some literature.
In this section, we introduce the evaluation protocols and datasets for continul LLMs.

\subsection{Continual LLMs' Evaluation Protocols}
\label{app:eval-llm}
\textbf{LAnguage Model Analysis~(LAMA).}\quad 
LAnguage Model Analysis (LAMA) is an evaluation framework designed to \emph{probe the world knowledge} embedded in language models~\cite{petroni2019language}.
It converts each world fact into a cloze statement, which is then inputted into the language models to predict the correct answer.
LAMA has been extended for continual pre-training, particularly for those under the temporal shifts~\cite{jang2022temporalwiki,jang2022towards}. In CKL, three LAMA benchmarks are constructed for different dimensions: InvariantLAMA assesses knowledge retention on time-invariant facts, UpdatedLAMA focuses on knowledge update, and NewLAMA evaluates knowledge acquisition~\cite{jang2022towards}. 

\textbf{Forgotten / (Updated + Acquired) Ratio~(FUAR).}\quad
As the performance of a pre-trained LLM is decomposed into a fine-grained set in CKL~\cite{jang2022towards}, OP becomes a too general metric and cannot accurately reflect the balance and trade-offs of the model's behavior. 
To address this issue, CKL proposes a joint evaluation metric FUAR~(Forgotten / (Updated + Acquired) Ratio) for continual pre-training. 
A FUAR value of 1 represents an equal trade-off between the knowledge forgetting and knowledge learning: for each piece of updated or acquired knowledge, one piece of time-invariant knowledge is forgotten on average. A FUAR less than 1 suggests high learning efficacy, where more than one piece of knowledge is acquired at the expense of forgetting one piece of time-invariant knowledge.

\textbf{X-Delta.}\quad
In TRACE~\cite{wang2023trace}, the authors propose a set of ``X-Delta'' metrics for continual instruction tuning, quantifying the forward transfer on specific abilities of LLMs. Let's denote a set of $M$ datasets $\{X_1, X_2, \cdots, X_M\}$ for task X. The baseline performances of the pre-trained LLM evaluated on these tasks are denoted as $\{b_1^X, \cdots, b_M^X\}$. The model undergoes continuous fine-tuning on a different set of tasks, distinct from those used for evaluation. Throughout the sequential training process, the performance of the model after learning task $t$ on evaluation tasks $X_i$ is $R_{t,i}^X$. The X-Delta $\Delta R_{t}^X$ after learning task $t$ is defined as:
\begin{align}
    \Delta R_t^X &\triangleq \frac{1}{M}\sum_{m=1}^M (R_{t,i}^X - b_i^X).
\end{align}
In the public TRACE benchmark, the authors construct three sets of evaluation tasks to benchmark the ability of LLMs, including \emph{general ability}, \emph{instruction following}, and \emph{safety}~\cite{wang2023trace}.


\textbf{NLG Score.}\quad In continual model alignment, three prominent metrics used to evaluate different aspects of Natural language generation (NLG) are BLEU-4~\cite{papineni2002bleu}, METEOR~\cite{banerjee2005meteor}, and ROUGE-L~\cite{lin2004rouge}. BLEU-4~\cite{papineni2002bleu}, designed for machine translation (MT), evaluates the precision of n-grams between the machine-generated and reference texts, focusing especially on four-word sequences to gauge fluency and adequacy. METEOR~\cite{banerjee2005meteor} also targets MT but aims to improve correlation with human judgment by considering synonyms and stemming, thus providing a more nuanced assessment of translation quality. On the other hand, ROUGE-L~\cite{lin2004rouge} is commonly applied in summarization tasks, assessing the longest common subsequence between the generated summary and a set of reference summaries, effectively measuring the recall of essential content. Each metric has its strengths and is tailored to specific kinds of language processing tasks, reflecting different dimensions of text generation quality. 


\subsection{Datasets}
\label{app:data}
In this section, we provide a comprehensive review of the datasets available for benchmarking continual LLMs, as illustrated in \Tabref{tab:datasets}. We intentionally exclude datasets used for domain-adaptive pre-training LLMs in vertical domains such as legal, medical, and financial, unless they are specifically designed for continual domain-adaptive pre-training. Furthermore, we omit datasets used in general continual fine-tuning, as they have already been extensively studied in existing works~\cite{biesialska2020continual,ke2023continual}.

\textbf{Datasets for Continual Pre-Training~(CPT) and Domain Adaptive Pre-Training~(DAP).}\quad
Current research lacks a widely recognized benchmark for evaluating continual pre-training LLMs under temporal shifts.
TimeLMs utilizes a series of Twitter corpora collected until 2022, sequentially pre-training RoBERTa models quarterly~\cite{loureiro2022timelms}.
CC-RecentNews, adopted as unlabeled pre-training data for LMs in CKL~\cite{jang2022towards}, consists of recent news and serves as a single-stage dataset. 
Additionally, CKL introduces InvariantLAMA, NewLAMA, and UpdatedLAMA to assess the principles of continual knowledge learning.
TWiki, a dataset derived from the articles of Wikipedia between August and December 2021, is curated and cleaned in TemporalWiki~\cite{jang2022temporalwiki}. This dataset facilitates the exploration of incremental learning by providing the Diffsets between neighboring snapshots.
For works that study the content-level distributional shifts in CPT and DAP, researchers often resort to a similar set of publicly available datasets~\cite{lo2020s2orc,xu2019bert,ni2019justifying} to construct their own test beds for continual learning algorithms. 
The $^*$DAPT dataset, developed by \cite{gururangan2020dont}, comprises four domains: BioMed and Computer Science from S2ORC \cite{lo2020s2orc}, News from \cite{zellers2019defending}, and Reviews from \cite{he2016ups}. In $^*$DAPT's original study, each domain undergoes individual domain adaptive pre-training stages to demonstrate the universality of DAP's effectiveness. Subsequent works, such as ELLE~\cite{qin2022elle} and Recyclable Tuning~\cite{qin2023recyclable}, follow suit by employing these domains for multi-stage CPT.
DEMix~\cite{gururangan2022demix} presents another large-scale dataset, featuring eight semantic domains with over 73.8 billion tokens. Alongside the training set, it includes eight additional datasets for validating the generalization ability of LLMs. 
On a smaller scale, $^*$CPT~\cite{ke2022continual-train} and $^*$DAS~\cite{ke2022continual-pre} datasets consist of four and eight domains, respectively, with approximately 3.12 million examples and a size of 4.16GB each. These datasets are constructed similarly to the aforementioned ones.

\textbf{Datasets for Continual Instruction Tuning.}\quad
Measuring the effectiveness of CIT is crucial, particularly because traditional evaluation metrics may not be suitable for LLMs: many of them are overly simplistic and fail to comprehensively assess the model's ability to learn continually. New benchmarks and metrics are required to evaluate both the retention of old knowledge and the integration of new instructions. TRACE~\cite{wang2023trace} stands as a continual learning benchmark designed specifically for LLMs, encompassing diverse tasks such as multilingual capabilities, code generation, and mathematical reasoning. CITB~\cite{zhang2023citb} represents another benchmark for CIT, incorporating both learning and evaluation protocols. It in addition demonstrates that replay generally yields the best performance across all methods. CoIN~\cite{chen2024coin} extends the benchmark to MLLMs, incorporating a balanced and diverse set of instructions from vision-language datasets. 


\textbf{Datasets for Continual Model Refinement.}\quad
Most datasets for continual model refinement can be categorized into two types~\cite{mazzia2023survey}: fact checking and question answering. For fact checking, models are asked to verify the truthfulness of certain claims, typically modeled as a classification task. Key datasets include FEVER~\cite{fever} (used by~\cite{de2021editing, hase2021language}) and VitaminC~\cite{vitaminC} (used by~\cite{mitchell2022memory}), both sourced from Wikipedia. For question answering, models are tasked with providing specific answers instead of choices. Zero-shot Relation Extraction (zsRE)~\cite{zsRE} is the most widely employed dataset for this purpose ~\cite{hase2021language, meng2022locating, meng2022mass, hase2023does, hartvigsen2023aging, das2024larimar}, alongside Natural Questions (NQ) ~\cite{nq} and T-rex ~\cite{T-rex}. \cite{meng2022locating} adapted zsRE with additional counterfactuals to create the more challenging CounterFact dataset, used by ~\cite{yu2023melo, hu2024wilke, das2024larimar}. Beyond these two categories, SCOTUS~\cite{scotus} is also utilized~\cite{hartvigsen2023aging} in the assessment of continual model refinement through a document classification task for U.S. Supreme Court cases into 11 topics.





\textbf{Datasets for Continual Model Alignment.}\quad
In the domain of reinforcement learning with human feedback (RLHF), several datasets are commonly employed across different studies to evaluate the adaptation and effectiveness of models under varying scenarios and continuous learning conditions. The IMDB~\cite{maas2011learning} and HH-RLHF~\cite{bai2022training} dataset, as introduced in~\cite{zhang2023copf} within their study on continual learning through optimal policy fitting, leverages data gathered from interactive RL scenarios to model human preferences dynamically. Similarly, the Reddit TL;DR dataset~\cite{volske2017tl} used by~\cite{zhangcppo,zhang2023copf} is focused on text summarization, providing a robust platform for testing the longevity and adaptability of learning algorithms under evolving conditions. Lastly, Common Sense QA~\cite{clark2018think, lai2017race, bisk2020piqa}, Reading Comprehension~\cite{rajpurkar2018know,dua2019drop}, and Translation~\cite{bojar2014findings}, which are utilized in~\cite{lin2024mitigating} are selected to assess the challenges of aligning RL agents with human expectations without incurring significant performance penalties. Each of these datasets is pivotal in advancing the understanding of continual learning and the interplay between human feedback and machine learning adaptation.

\textbf{Datasets for Continual Multimodal Large Language Models.}\quad
Following LLaVA~\cite{liu2023visual}, many MLLMs adopt the pattern of instruction tuning to enable assessing alignment with human intention and knowledge preservation for reasoning. Thus, traditional tasks like image classification can be transformed to VQA tasks to evaluate the ability of MLLMs, which are otherwise challenging to assess using conventional methods. Several benchmarks have been proposed to evaluate the CL method for MLLMs. MCIT~\cite{he2023continual} proposes the first continual instruction tuning benchmarks, Benchmark1 and Benchmark2. The difference between benchmark1 and benchmark2 is that benchmark2 includes Multi-task Joint Instruction Tuning, which aims to explore whether multi-task joint instruction tuning improves the model’s continual learning ability.
\cite{zhai2023investigating} proposes EMT, the first classification evaluation framework to investigate catastrophic forgetting in MLLMs. \cite{chen2024coin} presents a comprehensive benchmark CoIN, spanning 8 task categories and evaluating MLLMs from two perspectives:  Instruction Following and General Knowledge, which assess the alignment with human intention and knowledge preserved for reasoning, respectively. 
\cite{zhao2024reconstruct} constructs two datasets, UPMC-Food101-CMML and MM-IMDb-CMML to benchmark the novel CMML task, which means the data of certain modalities is missing during continual fine-tuning. UPMC-Food101-CMM contains 101 food categories and 61,142 training, 6,846 validation, and 22,716 test image-text pairs. MM-IMDb-CMML is a multi-label classification dataset across 27 distinct movie genres, consisting of 15,552 training, 2,608 validation and 7,799 test image-text pairs.

\end{document}